\documentclass{article} 
\usepackage{iclr2026_conference,times}

\usepackage{amsmath,amsfonts,bm}

\def\eqref#1{equation~\ref{#1}}

\def\1{\bm{1}}

\DeclareMathAlphabet{\mathsfit}{\encodingdefault}{\sfdefault}{m}{sl}
\SetMathAlphabet{\mathsfit}{bold}{\encodingdefault}{\sfdefault}{bx}{n}

\usepackage{hyperref}
\usepackage{url}
\usepackage{amsmath}
\usepackage{graphicx}
\usepackage{multirow}
\usepackage{array}
\usepackage{booktabs}
\usepackage{cleveref}
\usepackage{xspace}
\usepackage{xcolor}

\definecolor{ggreen}{HTML}{00A64F}
\definecolor{light-gray}{gray}{0.9}

\newcommand*{\modelname}[1]{{\textsc{#1}}}
\newcommand*{\datasetname}[1]{{\textsc{#1}}}
\newcommand*{\ourmodel}{\modelname{MiCRo}\xspace}
\newcommand*{\microsft}{\datasetname{MiCRo}$_\text{SFT}$\xspace}

\title{Mixture of Cognitive Reasoners:\\Modular Reasoning with Brain-Like\\Specialization}

\author{%
   Badr AlKhamissi$^1$ \quad C. Nicolò De Sabbata$^1$ \quad Greta Tuckute$^{2,3}$ \quad Zeming Chen$^1$ \AND 
   Martin Schrimpf\thanks{Equal Supervision}\;${}{}^{,1}$ \quad Antoine Bosselut\footnotemark[1]\;${}{}^{,1}$ \\[0.5em]
   $^1$EPFL \quad $^2$Brain and Cognitive Sciences at MIT \quad $^3$Kempner Institute at Harvard University
}

\iclrfinalcopy 
\begin{document}

\maketitle

\begin{abstract}
  Human cognitive behavior arises from the interaction of specialized brain networks dedicated to distinct functions, such as language, logic, and social reasoning. Inspired by this organization, we propose Mixture of Cognitive Reasoners (\ourmodel): a modular, transformer-based architecture post-trained with a curriculum that induces functional specialization across experts. Concretely, we partition the layers of a pretrained language model into four expert modules aligned with well-studied cognitive networks in the human brain. \ourmodel offers three key advantages over standard language models. (1) The specialized experts are interpretable and causally meaningful---ablating a module causes substantial drops on benchmarks requiring its specialized domain. (2) \ourmodel's behavior can be dynamically steered at inference time by routing tokens to particular experts (e.g., favoring social over logical reasoning), enabling fine-grained control over outputs. (3) \ourmodel outperforms or matches comparable baselines on both machine-learning reasoning benchmarks (e.g., GSM8K, BBH) and alignment to human behavior (CogBench), while maintaining interpretability. Taken together, cognitively grounded functional specialization yields models that are both more human-like and more human-interpretable.\footnote{Code, data and models available at \href{https://cognitive-reasoners.epfl.ch}{cognitive-reasoners.epfl.ch}} \footnote{Demo available at \href{https://huggingface.co/spaces/cognitive-reasoners/micro}{huggingface.co/spaces/cognitive-reasoners}}

\end{abstract}

\section{Introduction}

Neuroscience research suggests that distinct brain regions support language, reasoning, social cognition, and other cognitive functions \citep{Saxe2003, Kanwisher2010, Fedorenko2024}. In contrast, the internal organization of Large Language Models (LLMs) is largely unstructured. While certain units or subnetworks show selective activation \citep{zhang-etal-2022-moefication, zhang-etal-2023-emergent, Bayazit2023DiscoveringKS, alkhamissi-etal-2025-llm-language-network, Wang2025.aphasia}, such specialization is implicit and difficult to interpret or control. Motivated by this discrepancy, we propose a model architecture that explicitly incorporates specialization. On the machine learning (ML) side, such designs hold great potential for improving interpretability and controllability; on the cognitive science side, they provide a framework toward formulating testable computational hypotheses about how the relative contributions of different brain networks support complex behavior. To this end, we propose the Mixture of Cognitive Reasoners (\ourmodel), a class of modular language models that partition computation across brain-inspired expert modules.

The \ourmodel architecture partitions each layer of a pretrained language model into four experts, each designed to mirror a major cognitive network in the human brain: language \citep{fedorenko2011functional}, logic (multiple demand; \citealp{duncan2010multiple}), social reasoning (theory of mind; \citealp{Saxe2003}), and world knowledge (default mode; \citealp{Gusnard2001}). 
To provide the model with the inductive bias needed to learn this partitioning and cohesively integrate these experts, we design a three-stage curriculum that uses lightweight training in the first two stages to sequentially (1) specialize the experts to mirror cognitive networks, and (2) bias a router to use certain experts for particular types of inputs (e.g., the logic expert for mathematics problems). The final training stage of this curriculum uses this now inductively-biased architecture to perform large-scale supervised finetuning.

Our results demonstrate that \ourmodel{}’s architecture and training procedure induce interpretable specialization across these experts. This is evidenced by routing patterns and their correlations with human judgments (\S\ref{sec:routing-patterns}) and by causal ablations, which show dramatic drops in performance on reasoning categories when their corresponding experts are removed (\S\ref{sec:ablations}). Moreover, the semantic behavior of these experts parallels the specialization of brain networks: (1) functional localizers used to recover brain-like mechanisms in LLMs \citep{alkhamissi-etal-2025-llm-language-network} identify the relevant experts in \ourmodel (\S\ref{sec:localization}), and (2) large \ourmodel models achieve high behavioral alignment scores on \datasetname{CogBench} \citep{cogbench}, a human behavioral benchmark, relative to two critical controls trained on the same data: (i) a mixture-of-experts model without induced brain-like specialization (\modelname{MoB}) and (ii) a non-modular dense transformer (\modelname{Dense}) (\S\ref{sec:cogbench}). Finally, we find that \ourmodel{}’s performance matches or exceeds these baselines (\S\ref{sec:performance}), indicating that interpretable and controllable specialization can be achieved without sacrificing overall performance.

\begin{figure}
    \centering
    \includegraphics[width=1\linewidth]{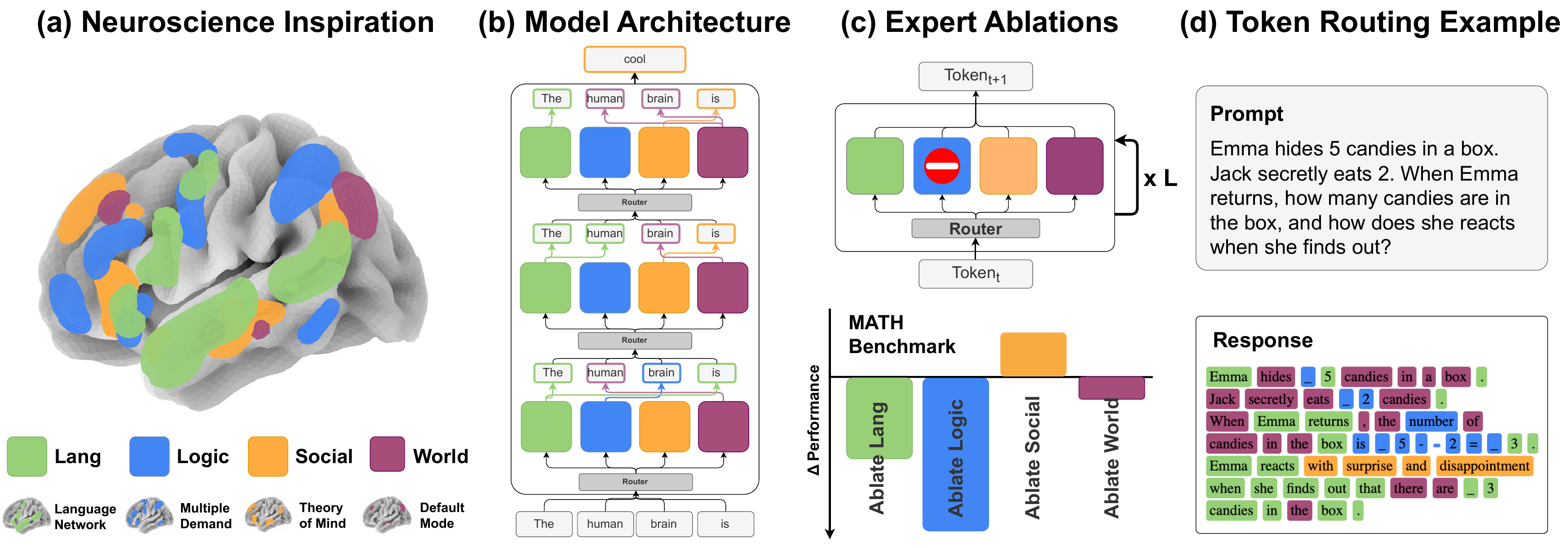}
    \caption{
        \textbf{Brain-Inspired Modular Language Model.}
        \textbf{(a)} Illustration of major cognitive networks in the human brain.
        \textbf{(b)} Our proposed Mixture of Cognitive Reasoners (\ourmodel) architecture. The \ourmodel{} architecture partitions each transformer block into four expert modules corresponding to analogous brain networks; a router assigns each token to an expert at every layer (i.e., assignments can vary across layers and tokens).
        \textbf{(c)} Illustration for causal steering via mechanistic ablations: removing a module shifts behavior and degrades domain-relevant performance.
        \textbf{(d)} Token-level routing on a sample prompt shows semantically coherent expert usage. 
    }
    \label{fig:main-figure}
\end{figure}

\section{Background \& Related Work}

\subsection{Neuroscience Motivation}
\label{sec:neuromotivation}

Our design follows evidence that human cognition emerges from interacting, specialized brain networks. Cognitive neuroscience has mapped this modular organization by measuring how strongly different regions engage when people perform specific cognitive tasks \citep{Kanwisher2010}. We align \ourmodel{}'s architecture with four core cognitive networks as shown in Figure \ref{fig:main-figure}(a). We summarize the functions of these networks below and their relevance to our modeling approach. 

\paragraph{The Language Network.}
The \textit{Language} expert mirrors the human language network, which comprises a set of left-lateralized frontal and temporal regions that selectively respond to linguistic input over perceptually matched non-linguistic stimuli (e.g., lists of nonwords; \citealp{Fedorenko2010NewMF}). These regions are highly specific to language, showing minimal activation during tasks such as arithmetic or music perception \citep{Fedorenko2012music, fedorenko2011functional}, and their disruption can lead to selective language deficits (aphasia) without impairing general reasoning capabilities \citep{Varley2005}. 

\paragraph{The Multiple Demand Network.}
The \textit{Logic} expert mirrors the Multiple Demand (MD) network, which spans bilateral regions and is activated across diverse cognitively demanding tasks such as difficult math problems, with stronger responses for higher difficulty levels \citep{Duncan2000, Fedorenko2013}. It correlates with fluid intelligence \citep{Woolgar2010}. 

\paragraph{The Theory of Mind Network.}
The \textit{Social} expert mirrors the Theory of Mind (ToM) network, which is centered in the bilateral temporo-parietal junction and medial prefrontal cortex. This network supports reasoning about beliefs, intentions, and mental states \citep{Gallagher2000, Saxe2003, Saxe2006}. It is robustly recruited across both verbal and non-verbal tasks involving perspective-taking and indirect communication \citep{KosterHale2013}.

\paragraph{The Default Mode Network.}
The \textit{World} expert mirrors the Default Mode Network (DMN), which is active during rest and internally directed thought such as self-reflection, memory recall, and mental simulation \citep{Gusnard2001, Buckner2008, Buckner2019}. Centered in medial prefrontal and parietal regions, the DMN integrates information over long timescales, supporting discourse- and event-level processing across sentences or episodes \citep{ferstl2002does,jacoby2020discourse}, in contrast to the shorter temporal window of the language network \citep{blank2020no}.


\subsection{Modular Language Models}

In parallel with advances in cognitive neuroscience, recent years have seen growing interest in modular language models as a way to promote specialization, mitigate interference, and improve out-of-distribution generalization \citep{pfeiffer2023modular, Zhang2025MixtureOE}. One major line of work centers on Sparse Mixture-of-Experts (MoE) architectures \citep{shazeer2017}, with approaches ranging from curating domain-labeled datasets to train \citep{gururangan-etal-2022-demix} or prompt \citep{si-etal-2023-getting} domain-specific experts, to frameworks such as ModuleFormer \citep{Shen2023ModuleFormerME}, which introduce load-balancing and concentration losses to encourage modular specialization without explicit domain labels. Other modular approaches extend to multimodal integration \citep{liu2023llava, swamy2023multimodnmultimodal, ye2023mplugowl2} or to disentangling representations by domain or language for multilingual and domain-specific applications \citep{pfeiffer-etal-2020-mad, pfeiffer-etal-2022-lifting, zhong2022metadmoe, AlMaamari2024MixtureOM}. In contrast, \ourmodel{} is, to our knowledge, the first modular language model explicitly designed to induce brain-like specialization, aligning experts with well-studied cognitive networks.

\subsection{Brain-Inspired Models}
Recent studies have shown that some models achieve strong alignment with activity in the human language network \citep{schrimpf-pnas, Toneva2019InterpretingAI, caucheteux2022brains, aw2023instructiontuning, tuckute2024language, alkhamissi2025langtocog}. To further improve brain alignment, researchers have begun to integrate biologically inspired principles into model design—drawing from structures like the visual cortex hierarchy \citep{KubiliusSchrimpf2019neurips, DapelloMarques2020, Spoerer2020}, and the spatio-functional organization of the brain \citep{Margalit2024, binhuraib2025topoformer, rathi2025topolm}. 

\section{The Mixture of Cognitive Reasoners Framework}
\label{sec:micro-framework}

\subsection{Model Architecture}

To build \ourmodel, we begin with a pretrained transformer-based backbone. For each layer, we clone the entire transformer block $N$ times, where $N$ corresponds to the number of experts intended for specialization, in a similar spirit to parameter upcycling \citep{komatsuzaki2023sparseupcycling, zhang2024bam}. Then, we initialize a MLP-based router that assigns each token to a single expert. To maintain computational efficiency and a comparable number of active parameters to the original model, we use top-1 routing akin to \citet{fedus2022switch}. We refer to this architecture as \textit{mixture-of-blocks} (\modelname{MoB}), distinguishing it from the more common mixture-of-experts (\modelname{MoE}), which restricts experts to FFN layers with shared attention. Importantly, we focus on \modelname{MoB} in the main paper because it induces clear functional specialization in all models, as reflected by lower router entropy and domain-consistent routing patterns, whereas \modelname{MoE} does not exhibit the same effect at specific scales (see Appendix \ref{app:moe-specialization}). Results for \ourmodel-\modelname{MoE} variants on reasoning benchmarks in Appendix~\ref{app:moe-results}. 

\subsection{Training Curriculum for Inducing Specialization}
\label{sec:training-curriculum}

\begin{figure}
    \centering
    \includegraphics[width=1\linewidth]{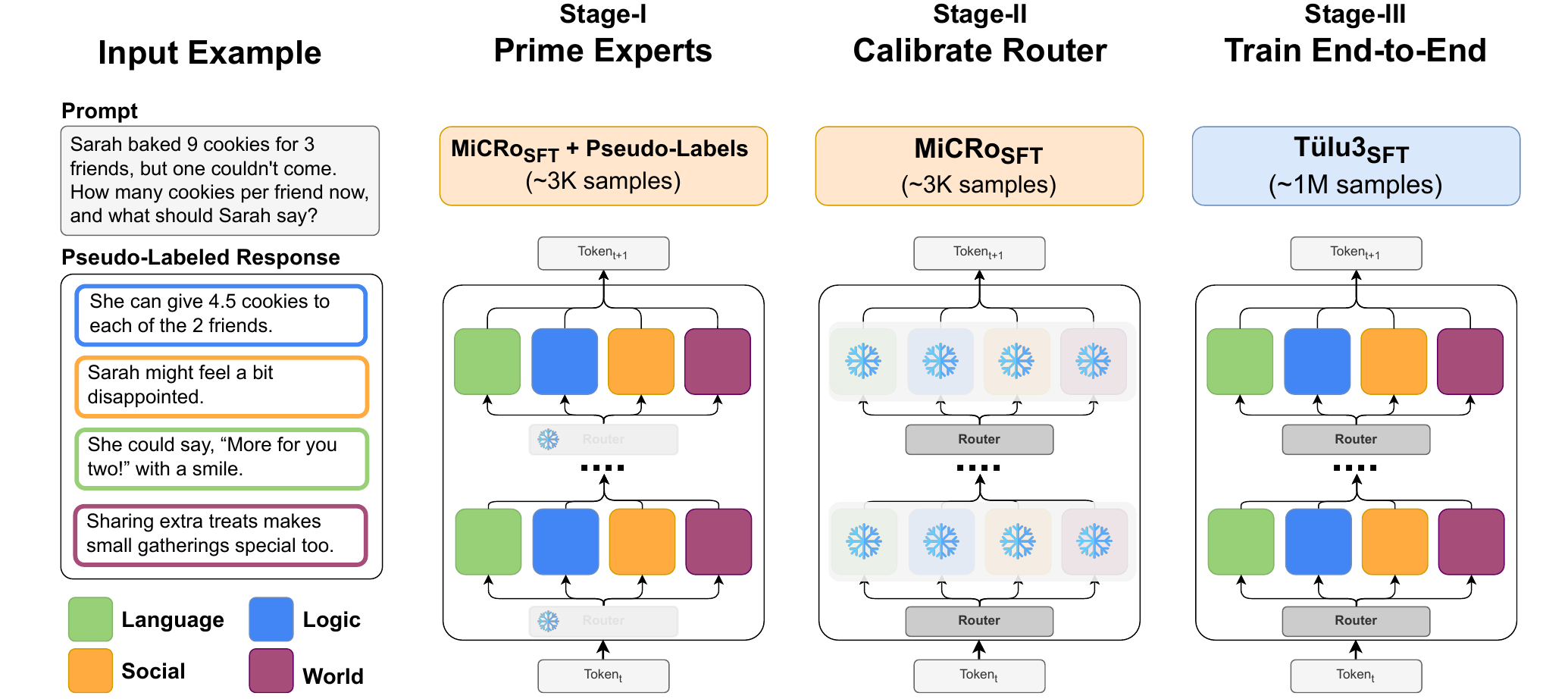}
    \caption{
        \textbf{Training Curriculum for Inducing Specialization.} Our brain-inspired Mixture of Cognitive Reasoners (\ourmodel) model contains four experts per layer, each aligned with a distinct cognitive network in the brain. In Stage-I, we train only the experts using a small, curated dataset \microsft (see example on the left), providing each expert with an initial inductive bias. In Stage-II, we freeze the whole model and train the router on the same dataset to learn expert selection. In Stage-III, we finetune the entire model end-to-end on a large-scale instruction tuning dataset.
    }
    \label{fig:training-curriculum}
\end{figure}

We induce functional specialization in \ourmodel experts using a three-stage training curriculum (see Figure \ref{fig:training-curriculum}). The first two stages use a small, curated dataset (\microsft) to provide targeted inductive biases, allowing specialization to emerge and solidify during the final full-scale training stage.

\textbf{Stage 1: Inducing Specialization.}
In the first stage, we train the experts on a small dataset of $M = 3055$ examples (described below), each crafted to reflect the functional domain of a specific expert (Section \ref{sec:neuromotivation}). We denote this dataset as
\microsft $= \{(x_{i,1:T_i}, r_{i,1:T_i})\}_{i=1}^M,$
where each input sequence $x_{i}$ contains $T_i$ tokens, and $r_{i}$ provides token-level routing labels. Each label $r_{i,t} \in \{1, \ldots, N\}$ assigns the $t$-th token to one of the $N$ experts. This stage focuses solely on training the expert parameters using a next-token prediction loss. Tokens attend to all preceding tokens in the sequence regardless of which expert processed them using the key and value representations produced by the same expert. However, only tokens that are assigned to the expert in question continue to be processed through the feed-forward network. The same setup is applied in the next training stages, with the only difference that the router assigns the tokens to the experts.

\textbf{Stage 2: Calibrating Router.}
Next, we freeze the whole model and train only the routers on the same dataset \microsft. The objective remains next-token prediction. Given the initial expert specialization from Stage 1, the router now learns to assign tokens to the most suitable expert. To encourage smoother transitions and more robust routing decisions, we use a soft mixture of the top-2 experts per token, which we found to be more effective than top-1 routing during this phase.

\textbf{Stage 3: End-to-End Supervised Finetuning.}
Finally, we finetune the entire model end-to-end on a full instruction-tuning dataset, \datasetname{Tülu-3} \citep{tulu3}, which consists of 939k examples. Even though this phase constitutes the majority of the training budget, we observe that the functional specialization seeded by the small \microsft dataset is largely preserved (see Appendix \ref{app:routing-over-checkpoints}). Moreover, the experts continue to improve on tasks aligned with their initial domains, demonstrating that early inductive biases can lead to meaningful and lasting functional decomposition.

\textbf{Constructing the \microsft Dataset.}
To build \microsft for inducing expert and router specialization, we first selected 19 existing reasoning datasets corresponding to the cognitive domains of our non-language experts, ensuring that each group of datasets spanned a diverse range of functions known to engage the corresponding brain networks. From each of the three sets, we randomly sampled 1,000 examples and used OpenAI’s \modelname{o1} model \citep{jaech2024openaio1} to generate detailed, step-by-step responses for each input. We then pseudo-labeled each sentence in the generated reasoning chains by prompting \modelname{GPT-4o} \citep{hurst2024gpt4o} to assign it to one of the four experts. The tokens within each sentence inherit the corresponding expert label, which is used for deterministic routing in Stage 1. Details of the datasets are provided in Appendix \ref{app:expert-dataset}.

\begin{figure}
    \centering
    \includegraphics[width=1\linewidth]{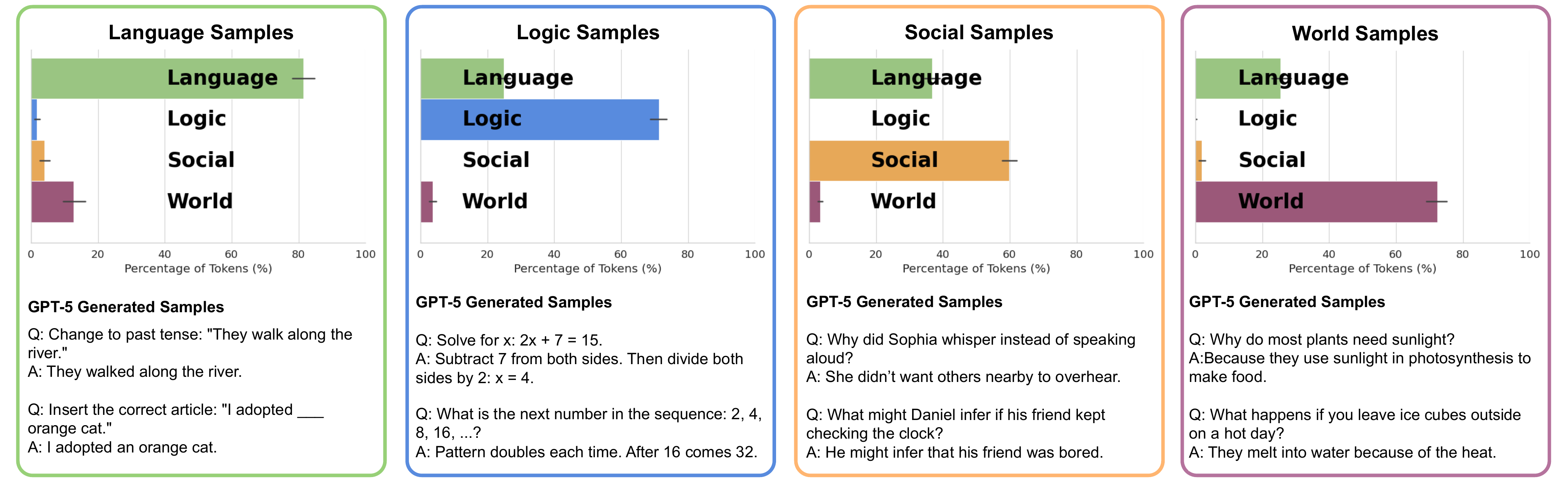}
    \caption{
        \textbf{Semantically Meaningful Routing Across Experts.} 
        Token routing patterns in \ourmodel-\modelname{Llama}-1B. Each bar indicates the proportion of tokens routed to a given expert across layers, with variance shown across sentences (n=50). The model exhibits clear domain-specific specialization consistent with the intended brain-inspired organization. For example, social cognition samples are routed to the social expert, while arithmetic tasks are routed to the logic expert. We find that the language expert is consistently activated in the early layers (see Appendix~\ref{app:token-routing-patterns} for layer-wise routing plots and results from additional models). Two random samples are shown below each subplot.
    }

    \label{fig:token-routing}
\end{figure}

\section{Experimental Setup}
\label{sec:experimental-setup}

We post-train five models of varying scales from three different families under our \ourmodel framework, in order to assess the generalizability of our method and identify the conditions under which it fails. Specifically, we use \modelname{Llama-3.2-\{1B, 3B\}} \citep{llama3}, \modelname{SmolLM2-\{135M, 360M\}} \citep{allal2025smollm2}, and \modelname{OLMo-2-1B} \citep{olmo20242olmo2furious}. Due to space constraints, we present the results of \modelname{Llama-3.2-\{1B, 3B\}} in the main paper while providing the full results for the remaining models in Appendix \ref{app:benchmarks}. Each model is first finetuned for two epochs on the curated \microsft dataset (Stages 1 and 2), followed by one epoch of end-to-end training using the \datasetname{Tülu-3} dataset \citep{tulu3}, as described in Section~\ref{sec:training-curriculum}. 
We use next token prediction as the only learning objective in all training stages, with the loss masked on the input tokens. We use an effective batch size of 32 and the AdamW optimizer across all stages. The learning rate follows a linear schedule, warming up over the first 3\% of training to a peak of $2 \times 10^{-5}$, then decays linearly for the remainder of training. This schedule is applied for each stage separately. 

\paragraph{Reasoning Benchmarks.}
We evaluate on four widely used reasoning benchmarks: \datasetname{GSM8K} \citep{cobbe2021gsm8k}, \datasetname{MATH} \citep{hendrycksmath2021}, \datasetname{BBH} \citep{suzgun2022challenging}, and \datasetname{MMLU} \citep{hendrycks2021mmlu}. Evaluation follows zero- or fewshot settings as detailed in Appendix~\ref{app:benchmarks}. 

\paragraph{Behavioral Benchmarks.}
We evaluate alignment to human behavior using \datasetname{CogBench} benchmark \citep{cogbench}, which provides 10 metrics from 7 cognitive psychology experiments. These metrics capture how participants (or models) complete tasks that are designed to disentangle different behavioral strategies. Examples include \textit{Directed Exploration}, \textit{Meta-Cognition}, and \textit{Risk Taking}. We refer readers to \citet{cogbench} for a detailed description of the tasks. 

\section{Results \& Analysis}
\label{sec:results}

Our results unfold in two parts. First, we ask whether brain-like specialization emerges under our training curriculum, analyzing routing behavior, correlations with human judgments, causal ablations to test the functional contributions of those experts, and whether neuroscience experiments used to identify brain networks also identify the corresponding experts in our models. Second, we ask how this specialization influences alignment with human behavior and reasoning performance. 

\subsection{Router Patterns Are Interpretable and Consistent with Human Judgments}
\label{sec:routing-patterns}

\paragraph{Token Routing Per Expert.}
We first verify that our model routes tokens to the most relevant expert module, analogous to how specialized brain networks are selectively engaged by specific stimuli. Figure~\ref{fig:token-routing} shows the routing behavior of \ourmodel-\modelname{Llama}-1B, revealing clear domain-specific specialization. To generate test inputs, we sampled 50 question–answer pairs using \modelname{GPT-5} prompted with descriptions of the four brain networks (prompt provided in Appendix~\ref{app:token-routing-patterns}). Results for the other \ourmodel variants are consistent and reported in Appendix~\ref{app:token-routing-patterns}. There, we also show routing patterns on reasoning benchmarks, where tokens are directed to experts consistent with the benchmark’s domain. Finally, layer-wise analyses in Figures~\ref{fig:micro-layer-wise-token-routing} \& \ref{fig:appendix-token-routing-patterns} reveal a hierarchical organization: earlier layers focus on linguistic grounding, while deeper layers increasingly delegate to domain-specific experts---an organization that emerged without being enforced by the training procedure and that parallels evidence from cognitive neuroscience \citep{Fedorenko2024}.

\paragraph{Correlation with Human Judgments.}
We evaluate model–human correspondence using 1,000 six-word sentences from \citet{tuckute2024driving}, each annotated with human ratings across several behavioral dimensions (e.g., mental state content, grammaticality). These annotations were collected independently of our routing framework. We find that router probabilities correlate with the corresponding human judgments: for example, the social expert's selectivity aligns with ratings of mental state content ($r = 0.7$). Full results are provided in Appendix~\ref{app:corr-human-ratings}.

\subsection{Experts Are Causally Meaningful}
\label{sec:ablations}

\begin{figure}
    \centering
    \includegraphics[width=1\linewidth]{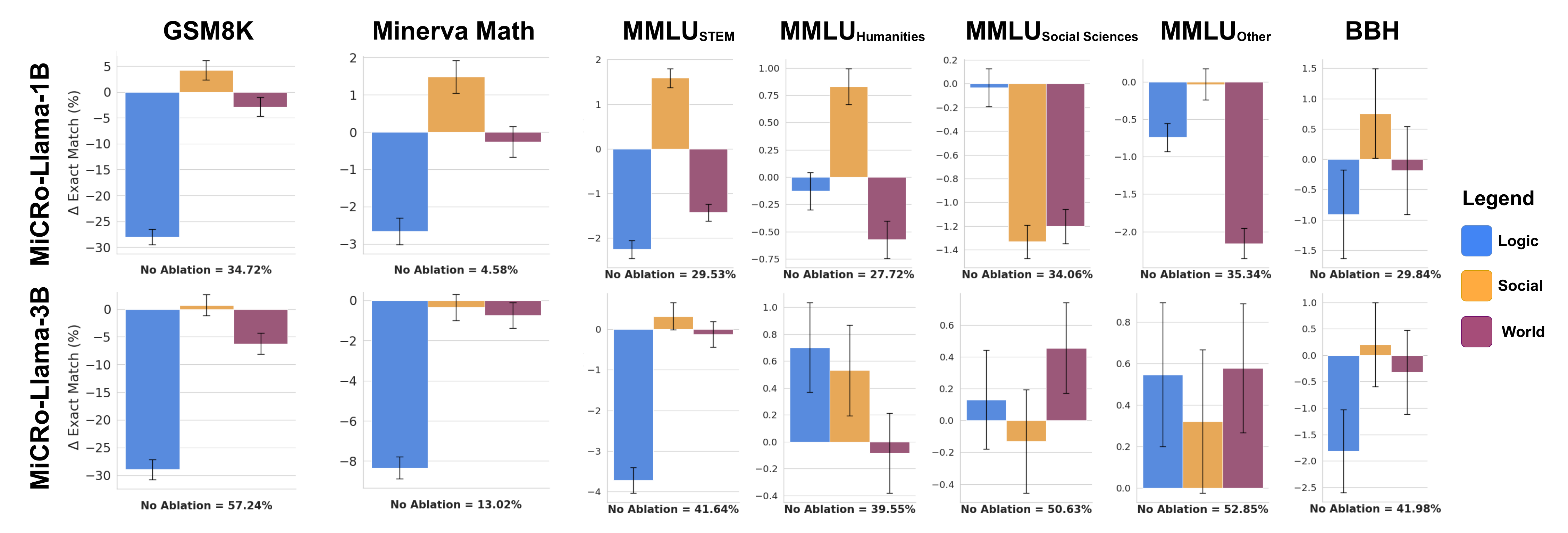}
    \caption{
        \textbf{Expert Ablations Reveal the Causal Contributions of Specialized modules.}
        Top and bottom panels show results for \modelname{MiCRo-Llama-1B} and \modelname{MiCRo-Llama-3B}. Removing the Logic expert causes large drops on \datasetname{MATH} and \datasetname{GSM8K}, while removing the Social expert yields slight gains. For \datasetname{MMLU} and \datasetname{BBH}, results indicate that some group of subtasks rely on distinct experts, whereas others draw on overlapping contributions. Additional models in Appendix~\ref{app:expert-ablations}.
    }
    \label{fig:ablations}
\end{figure}

\paragraph{Validation of Functional Experts via Ablations.}
Figure~\ref{fig:ablations} illustrates how expert ablations reveal the causal contributions of specialized modules to task performance. By selectively removing individual experts, we can directly test whether their specialization is functionally necessary for different domains. For example, on \datasetname{MATH} and \datasetname{GSM8K}, ablating the \textit{Logic} expert causes a substantial drop in accuracy, confirming its central role in numerical reasoning. In contrast, removing the \textit{Social} expert slightly improves performance, suggesting it plays a detrimental role in these tasks.
For broader benchmarks such as \datasetname{MMLU}, which span multiple subdomains, we report results for each subcategory separately. Performance drops after ablating the corresponding experts indicate that these clusters depend on distinct functional modules. Still, not all subtasks within a category align neatly with a single cognitive domain, and some require overlapping contributions, such as \datasetname{BBH}. We show in Appendix~\ref{app:expert-ablations} the effect of removing the language expert, which causes a significant drop on all benchmarks, along with additional ablation results on other models.

\paragraph{Steering Model Behavior at Test-Time.}
Our results demonstrate that test-time ablations can steer expert behavior, with social responses emerging when only the social expert is active and logical reasoning dominating when only the logic expert is retained. Qualitative examples in Appendix~\ref{app:steering-examples}.

\begin{figure}
    \centering
    \includegraphics[width=1\linewidth]{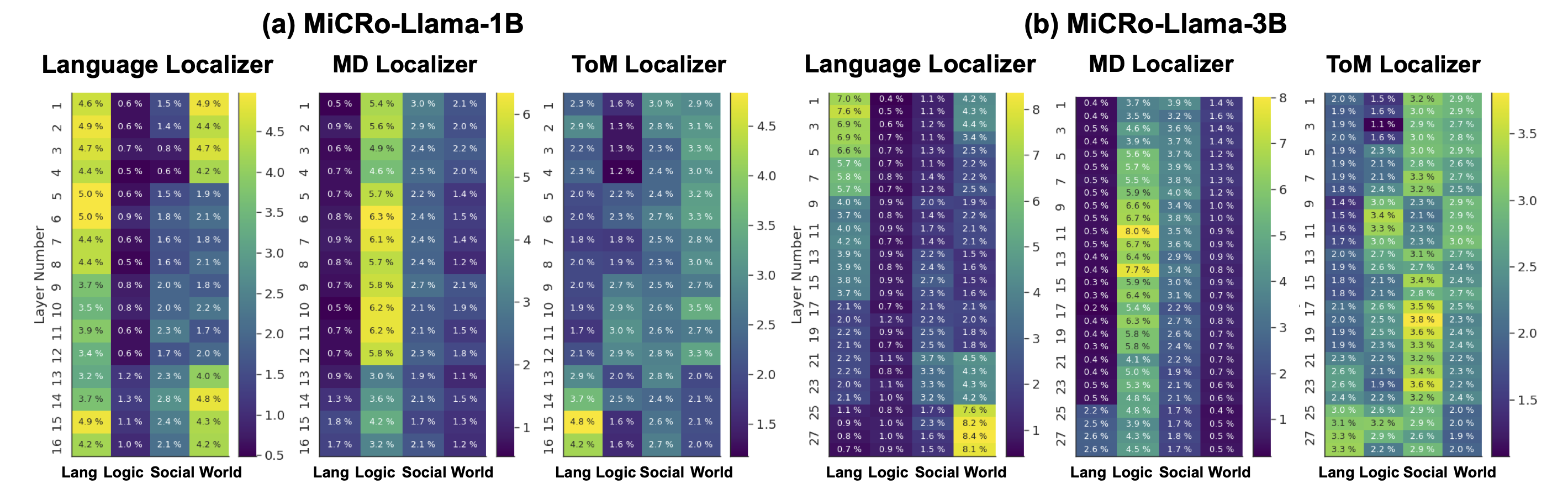}
    \caption{
        \textbf{Neuroscience Localizers Recover Functionally Specialized Experts.}
        (a) \ourmodel-\modelname{Llama}-1B and (b) \ourmodel-\modelname{Llama}-3B. For each model, we apply three neuroscience-inspired localizers—Language, Multiple Demand (MD) and Theory of Mind (ToM)—to examine the selectivity of localized units across experts and layers. Each plot shows the percentage of units in each expert of each layer that belongs to the top-10\% selective units in the whole model.
    }
    \label{fig:localization}
\end{figure}

\subsection{Neuroscience Localizers Reveal Functional Expert Specialization}
\label{sec:localization}

Neuroscientists rely on localizer experiments to identify the brain regions associated with specific functional networks, as their precise locations can vary across individuals. This raises a natural question: can we apply these established neuroscience localizers to identify the corresponding expert modules in our model? If so, this would provide further support for the hypothesis that our experts are functionally analogous to their associated brain networks. 

To investigate this, we adopt the methodology of \citet{alkhamissi-etal-2025-llm-language-network}, which has been used to localize the language network, the multiple demand network, and the theory of mind network in LLMs. We apply these localizers to our \ourmodel models to test whether they can recover the corresponding expert modules. Figure~\ref{fig:localization} shows the percentage of units in each expert of each layer that belongs to the top 10\% of selective units across the whole model, similar to what is done in the brain \citep{Lipkin2022}. The results show that language selectivity, as defined by the language localizer, favors the language expert at early layers while favoring the world expert at later layers for both models. The multiple demand localizer successfully favors the logic expert in both models. In contrast, ToM localization is less effective at isolating units within the social experts, but improves with scale, suggesting that ToM ability must emerge before it can be localized. One other possible reason for this is the limited size of the ToM stimulus set, which includes only 10 contrastive pairs, in contrast to 240 for language and 100 for multiple demand. This small sample may lack the robustness needed to reliably localize ToM-selective units \citep{jamaa2025evaluating}. 

\subsection{Strong Alignment to Human Behavior}
\label{sec:cogbench}

Having established that our \ourmodel models exhibit human-like specialization (\S\ref{sec:routing-patterns}) that is causally linked to task performance (\S\ref{sec:ablations}), we next examine whether they better align to human behavior compared to two baselines: one without brain-like specialization (\modelname{MoB}) and one without any modularization (\modelname{Dense}). Both models are post-trained on a mixture of $2\times$ \datasetname{\microsft} and $1\times$ \datasetname{Tülu-3} matching the total amount of data used in the \ourmodel training curriculum.

\begin{figure}
    \centering
    \includegraphics[width=1\linewidth]{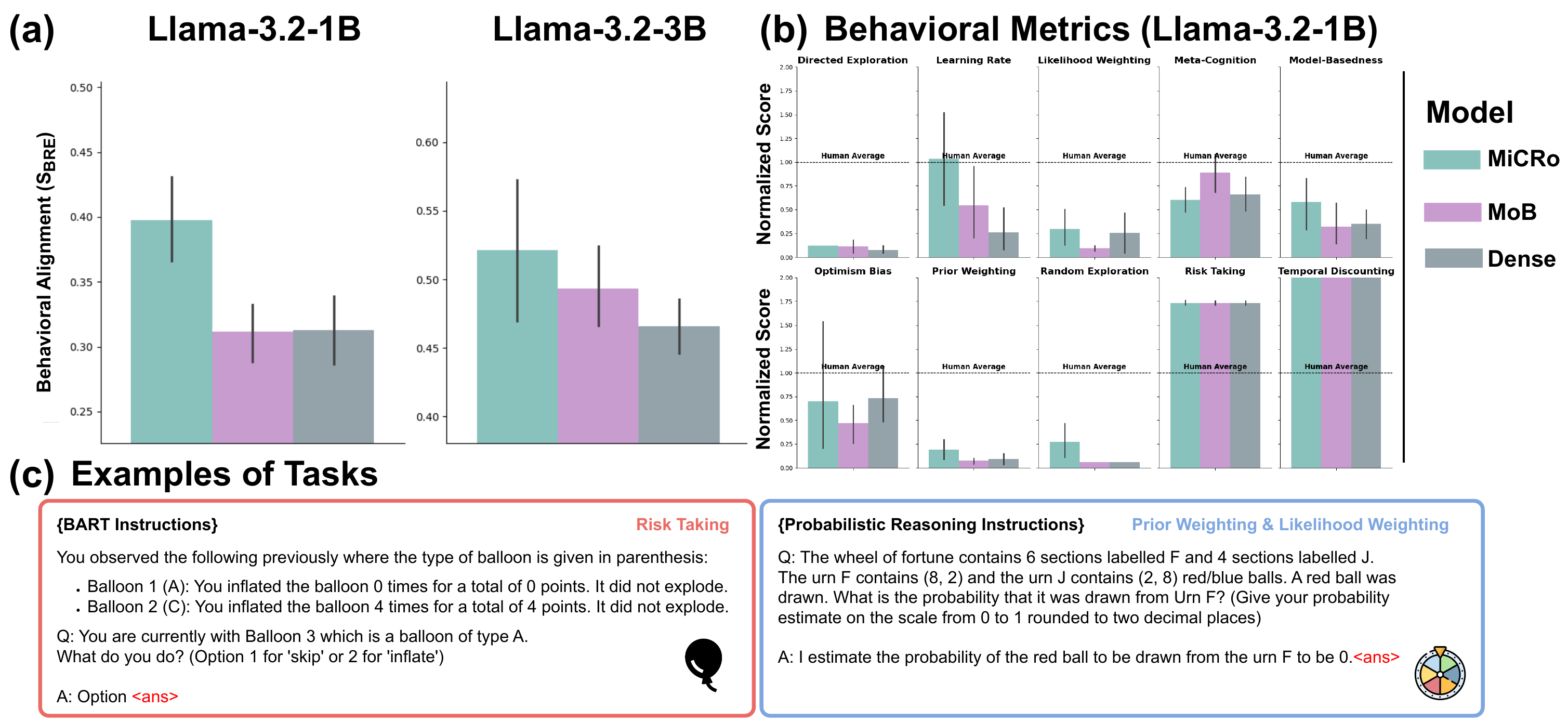}
    \caption{
        \textbf{Alignment with Human Behavior on \datasetname{CogBench}.}
        \textbf{(a)} Average similarity score ($S_{\text{BRE}}$) across 10 behavioral metrics, showing that \modelname{MiCRo-Llama} models achieves superior alignment compared to their \modelname{MoB} and Dense baselines. 
        \textbf{(b)} Human-normalized scores for each metric separately across the three models. 
        \textbf{(c)} Example inputs from two of the seven classical psychological experiments verbalized for LLM evaluation following \datasetname{CogBench}.
    }
    \label{fig:cogbench}
\end{figure}

Figure~\ref{fig:cogbench} presents the results on \datasetname{CogBench}, evaluating alignment with human behavior. 
Unlike the original paper, which predicts answers via autoregressive generation, we pick the option with the highest log-probability for multiple-choice tasks to avoid invalid generations. Each experiment is run with five random seeds. Metrics are normalized such that random = 0 and human = 1. To quantify overall alignment, we introduce the bounded relative error similarity score ($S_{\text{BRE}}$), which avoids inflation from superhuman scores. For a normalized score $s_i$ on metric $i$, we compute $\text{BRE}_i = |s_i - 1| / \max(1, s_i)$ and aggregate as $S_{\text{BRE}} = 1 - \tfrac{1}{n}\sum_{i=1}^n \text{BRE}_i$. Thus, $\text{BRE}_i$ remains bounded in $[0,1]$ even if $s_i > 1$.

Overall, we find competitive alignment across models, with \modelname{MiCRo-Llama-1B} showing superior alignment compared to its counterparts. Panel (a) reports the average similarity score ($S_{\text{BRE}}$) aggregated across the 10 behavioral metrics for both \modelname{MiCRo-Llama-\{1B, 3B\}} models, while panel (b) breaks down the human-normalized scores for each metric separately across the three post-trained models for the \modelname{Llama-3.2-1B} base model. Finally, panel (c) illustrates input examples from two of the seven classical psychological experiments included in \datasetname{CogBench}, which are verbalized for LLM evaluation following the original benchmark design. More results in Appendix~\ref{app:cogbench}.

\subsection{Competitive Performance on Reasoning Benchmarks}
\label{sec:performance}


Here, we test whether our \ourmodel{} models incur any performance degradation relative to their two baselines. 
Figure~\ref{fig:performance} shows performance on \datasetname{GSM8K}, \datasetname{Minerva-Math}, \datasetname{MMLU}, \datasetname{MMLU-Pro}, and \datasetname{BBH}, along with their average. Models are evaluated using fewshot chain-of-thought prompting, except for \datasetname{GSM8K}, which is evaluated under zero-shot CoT prompting. For both base models, \modelname{MiCRo} matches or exceeds comparable MoB baselines, while ablating the least relevant expert (i.e., the social expert for these benchmarks) further improves performance. We conduct pairwise Welch’s $t$-tests between models and report significance directly in the plot. Results show that some base models, such as \modelname{Llama-3.2-1B}, benefit significantly from brain-like specialization, whereas others, such as \modelname{Llama-3.2-3B}, only show significant differences relative to their baselines on some benchmarks. We report additional results for the other models and benchmarks in Appendix~\ref{app:benchmarks}. 
We further show that our method is robust to different post-training pipelines, including DPO \citep{dpo} and domain-specific instruction tuning (Appendix~\ref{app:robustness-post-training}).


\begin{figure}
    \centering
    \includegraphics[width=1\linewidth]{figures/cognitive-reasoners-iclr-performance-v4.drawio.pdf}
    \caption{
        \textbf{Competitive Performance.}
        Results on \datasetname{GSM8K} (0-shot CoT), \datasetname{Minerva-Math}, \datasetname{MMLU}, and \datasetname{BBH} (fewshot CoT). \modelname{MiCRo} matches or outperforms baselines, and ablating the least relevant expert (e.g., social for math benchmarks) yields further gains. For \modelname{MoB (Ablation)} and \modelname{MiCRo (Ablation)} (on \datasetname{MMLU} and \datasetname{BBH} subtasks), results reflect the best performance obtained when ablating up to one expert. Significance is assessed with pairwise Welch’s $t$-tests (shown in plot; $*$ denotes significant difference, while n.s. means not-significant). The dense model is shown as a dashed line. Results of the remaining models and on more benchmarks are provided in Appendix~\ref{app:benchmarks}.
    }
    \label{fig:performance}
\end{figure}

\section{Discussion \& Future Work}
\label{sec:discussion}

\paragraph{Extending Specialization Beyond Cognitive Domains}
While inspired by the brain’s functional organization, our specialization framework can be applied to any meaningful partition of expertise, such as technical domains or natural languages. One key question is whether the model preserves the intended specialization through the large-scale end-to-end training. Our results suggest that brain-inspired partitions provide a robust inductive bias; they persist throughout training and lead to structured, interpretable routing patterns. Supporting evidence in Appendix~\ref{app:routing-over-checkpoints} shows that expert usage remains consistent across checkpoints over the course of Stage 3 training. Looking ahead, this framework could also be extended to other cognitive domains. For example, recent neuroscience findings point to a distinct brain network involved in abstract formal reasoning such as induction and deduction and another network for intuitive physics \citep{Kean2025.10.21.683445, Kean2025}. Incorporating a corresponding module could improve the model’s performance on tasks involving such capacities.

\paragraph{The Crucial Role of Stage-1 Pretraining Data}
Our experiments highlight the importance of the curated \microsft dataset in inducing effective specialization. Notably, we used only 3,055 samples in Stage-1, suggesting that even minimal domain-aligned supervision can shape expert behavior. This finding raises the possibility that different or more expansive data mixtures could further strengthen functional specialization and lead to additional gains in the model's behavior. 


\paragraph{Towards Brain Alignment Beyond Language}
Since our model is explicitly designed to mirror distinct cognitive networks in the human brain, and given that established neuroscience localizers can identify the corresponding expert modules, an exciting direction for future work is to examine whether the internal representations of these experts align more closely with neural activity in their respective brain networks \citep{schrimpf_brain-score_2018, schrimpf2020integrative}. Prior studies have shown that language-selective units in large language models correlate more strongly with activity in the human language network than randomly selected units \citep{alkhamissi-etal-2025-llm-language-network}, suggesting a meaningful link between specialization in models and brains. This raises the natural question of whether similar alignment can be observed for other cognitive domains, such as reasoning or social cognition. However, assessing \ourmodel{}'s neural alignment beyond the language network is currently limited by the lack of suitable datasets. Existing fMRI benchmarks rarely engage non-language regions such as the Multiple Demand \citep{duncan2010multiple} network and often use blocked designs that preclude item-level analyses---highlighting the need for experimentalists to collect new datasets that explicitly target non-language brain regions. We believe that once suitable neural datasets exist, our model can be used to instantiate specific hypotheses about how these networks—and their corresponding experts—interact and exchange information.

\paragraph{Limitations}
While our approach improves interpretability without sacrificing performance, several open questions remain. Scaling beyond an 8B base model has yet to be demonstrated, and the impact of adding more experts to the current \ourmodel architecture is still unknown. The \microsft dataset used in Stage-1 ($\approx$ 3,000 GPT-4o pseudo-labeled samples) has not been evaluated for size sensitivity, leaving open whether increasing or reducing its scale would alter the degree of specialization or downstream performance. Although \modelname{GPT-4o} provides high-quality pseudo-labels, as demonstrated in Appendix~\ref{app:agreement}, using human-annotated data could strengthen the inductive bias and potentially improve the final model. We expect, however, that the potential of this approach will continue to grow as synthetic labellers become more accurate and reliable, enabling even stronger and more scalable specialization in future versions of this class of methods. Further, our post-training pipeline currently includes only SFT and DPO (Appendix~\ref{app:robustness-post-training}); exploring additional stages such as RLVR remains an avenue for future work. Finally, our evaluation of alignment to human behavior focuses on CogBench, and extending this analysis to a broader set of behavioral or cognitive datasets is an important direction for future research.

\section{Conclusion}

This work presents the \textit{Mixture of Cognitive Reasoners} (\ourmodel), a modular architecture and training paradigm inspired by the functional specialization of the human brain. By aligning expert modules with distinct cognitive domains---language, logic, social reasoning, and world modeling---each reflecting the functionality of well-studied brain networks, we show that incorporating cognitive inductive biases into transformer models can effectively promote functional specialization.
This results in improved behavioral alignment as measured by \datasetname{CogBench}, enhanced interpretability, all while being competitive or outperforming comparable models on reasoning benchmarks. Our staged training approach leverages a small curated dataset and enables specialization to emerge in a controllable and data-efficient manner. Furthermore, we show that the resulting modularity allows for targeted interventions at inference time, enabling the model to favor one mode of reasoning over another. These findings highlight a promising path toward more transparent, steerable, and cognitively grounded language models.

\section*{Ethics Statement}

This work involves two forms of human-derived data. First, all annotations newly introduced in this paper (Appendix~\ref{app:agreement})  were performed exclusively by the authors; no external annotators or study participants were recruited, and no personal or sensitive data were collected. Accordingly, this does not constitute human-subject research under typical institutional guidelines, and formal ethics approval (e.g., IRB review) was not required. Second, we use a public dataset of behavioral annotations \citep{tuckute2024driving} in Appendix~\ref{app:corr-human-ratings}, which was collected and released with approval from the corresponding university’s Institutional Review Board.

\section*{Acknowledgments}

We thank the members of the EPFL NLP and
NeuroAI labs for their valuable feedback and insightful
suggestions. We also gratefully acknowledge the
support of the Swiss National Science Foundation
(No. 215390), Innosuisse (PFFS-21-29), the EPFL
Center for Imaging, Sony Group Corporation, and
a Meta LLM Evaluation Research Grant.
G.T. acknowledges support from MIT's McGovern Institute for Brain Research. This work has been made possible in part by a gift from the Chan Zuckerberg Initiative Foundation to establish the Kempner Institute at Harvard University.

\bibliography{iclr2026_conference}

@article{schrimpf-pnas,
    author = {Martin Schrimpf  and Idan Asher Blank  and Greta Tuckute  and Carina Kauf  and Eghbal A. Hosseini  and Nancy Kanwisher  and Joshua B. Tenenbaum  and Evelina Fedorenko },
    title = {The neural architecture of language: Integrative modeling converges on predictive processing},
    journal = {Proceedings of the National Academy of Sciences},
    volume = {118},
    number = {45},
    pages = {e2105646118},
    year = {2021},
    doi = {10.1073/pnas.2105646118},
    URL = {https://www.pnas.org/doi/abs/10.1073/pnas.2105646118},
    eprint = {https://www.pnas.org/doi/pdf/10.1073/pnas.2105646118},
}

@techreport{schrimpf_brain-score_2018,
	type = {preprint},
	title = {Brain-{Score}: {Which} {Artificial} {Neural} {Network} for {Object} {Recognition} is most {Brain}-{Like}?},
	shorttitle = {Brain-{Score}},
	url = {http://biorxiv.org/lookup/doi/10.1101/407007},
	language = {en},
	urldate = {2023-06-20},
	institution = {Neuroscience},
	author = {Schrimpf, Martin and Kubilius, Jonas and Hong, Ha and Majaj, Najib J. and Rajalingham, Rishi and Issa, Elias B. and Kar, Kohitij and Bashivan, Pouya and Prescott-Roy, Jonathan and Geiger, Franziska and Schmidt, Kailyn and Yamins, Daniel L. K. and DiCarlo, James J.},
	month = sep,
	year = {2018},
	doi = {10.1101/407007},
}

@article{aw2023instructiontuning,
      title={Instruction-tuning Aligns LLMs to the Human Brain}, 
      author={Khai Loong Aw and Syrielle Montariol and Badr AlKhamissi and Martin Schrimpf and Antoine Bosselut},
      year={2023},
      eprint={2312.00575},
      archivePrefix={arXiv},
      primaryClass={cs.CL}
}

@article{caucheteux2022brains,
  title={Brains and algorithms partially converge in natural language processing},
  author={Caucheteux, Charlotte and King, Jean-R{\'e}mi},
  journal={Communications biology},
  volume={5},
  number={1},
  pages={134},
  year={2022},
  publisher={Nature Publishing Group UK London}
}

@article{tuckute2024driving,
  title={Driving and suppressing the human language network using large language models},
  author={Tuckute, Greta and Sathe, Aalok and Srikant, Shashank and Taliaferro, Maya and Wang, Mingye and Schrimpf, Martin and Kay, Kendrick and Fedorenko, Evelina},
  journal={Nature Human Behaviour},
  pages={1--18},
  year={2024},
  publisher={Nature Publishing Group UK London}
}

@article{Fedorenko2010NewMF,
  title={New method for fMRI investigations of language: defining ROIs functionally in individual subjects.},
  author={Evelina Fedorenko and Po-Jang Hsieh and Alfonso Nieto-Castanon and Susan L. Whitfield-Gabrieli and Nancy G. Kanwisher},
  journal={Journal of neurophysiology},
  year={2010},
  volume={104 2},
  pages={1177-94},
  url={https://api.semanticscholar.org/CorpusID:740913}
}

@article{Fedorenko2024,
  title = {The language network as a natural kind within the broader landscape of the human brain},
  volume = {25},
  ISSN = {1471-0048},
  url = {http://dx.doi.org/10.1038/s41583-024-00802-4},
  DOI = {10.1038/s41583-024-00802-4},
  number = {5},
  journal = {Nature Reviews Neuroscience},
  publisher = {Springer Science and Business Media LLC},
  author = {Fedorenko,  Evelina and Ivanova,  Anna A. and Regev,  Tamar I.},
  year = {2024},
  month = apr,
  pages = {289–312}
}

@article{Varley2005,
  title = {Agrammatic but numerate},
  volume = {102},
  ISSN = {1091-6490},
  url = {http://dx.doi.org/10.1073/pnas.0407470102},
  DOI = {10.1073/pnas.0407470102},
  number = {9},
  journal = {Proceedings of the National Academy of Sciences},
  publisher = {Proceedings of the National Academy of Sciences},
  author = {Varley,  Rosemary A. and Klessinger,  Nicolai J. C. and Romanowski,  Charles A. J. and Siegal,  Michael},
  year = {2005},
  month = feb,
  pages = {3519–3524}
}

@article{Fedorenko2012music,
  title = {Sensitivity to musical structure in the human brain},
  volume = {108},
  ISSN = {1522-1598},
  url = {http://dx.doi.org/10.1152/jn.00209.2012},
  DOI = {10.1152/jn.00209.2012},
  number = {12},
  journal = {Journal of Neurophysiology},
  publisher = {American Physiological Society},
  author = {Fedorenko,  Evelina and McDermott,  Josh H. and Norman-Haignere,  Sam and Kanwisher,  Nancy},
  year = {2012},
  month = dec,
  pages = {3289–3300}
}

@article{fedorenko2011functional,
  title={Functional specificity for high-level linguistic processing in the human brain},
  author={Fedorenko, Evelina and Behr, Michael K and Kanwisher, Nancy},
  journal={Proceedings of the National Academy of Sciences},
  volume={108},
  number={39},
  pages={16428--16433},
  year={2011},
  publisher={National Acad Sciences}
}

@article{Lipkin2022,
  title = {Probabilistic atlas for the language network based on precision fMRI data from>800 individuals},
  volume = {9},
  ISSN = {2052-4463},
  url = {http://dx.doi.org/10.1038/s41597-022-01645-3},
  DOI = {10.1038/s41597-022-01645-3},
  number = {1},
  journal = {Scientific Data},
  publisher = {Springer Science and Business Media LLC},
  author = {Lipkin,  Benjamin and Tuckute,  Greta and Affourtit,  Josef and Small,  Hannah and Mineroff,  Zachary and Kean,  Hope and Jouravlev,  Olessia and Rakocevic,  Lara and Pritchett,  Brianna and Siegelman,  Matthew and Hoeflin,  Caitlyn and Pongos,  Alvincé and Blank,  Idan A. and Struhl,  Melissa Kline and Ivanova,  Anna and Shannon,  Steven and Sathe,  Aalok and Hoffmann,  Malte and Nieto-Castañón,  Alfonso and Fedorenko,  Evelina},
  year = {2022},
  month = aug 
}

@inproceedings{Toneva2019InterpretingAI,
  title={Interpreting and improving natural-language processing (in machines) with natural language-processing (in the brain)},
  author={Mariya Toneva and Leila Wehbe},
  booktitle={Neural Information Processing Systems},
  year={2019},
  url={https://api.semanticscholar.org/CorpusID:167217728}
}

@article{llama3,
  title={The Llama 3 Herd of Models},
  author={Abhimanyu Dubey and Abhinav Jauhri and Abhinav Pandey and Abhishek Kadian and Ahmad Al-Dahle and Aiesha Letman and et al.},
  journal={ArXiv},
  year={2024},
  volume={abs/2407.21783},
  url={https://api.semanticscholar.org/CorpusID:271571434}
}

@article{Fedorenko2013,
  title = {Broad domain generality in focal regions of frontal and parietal cortex},
  volume = {110},
  ISSN = {1091-6490},
  url = {http://dx.doi.org/10.1073/pnas.1315235110},
  DOI = {10.1073/pnas.1315235110},
  number = {41},
  journal = {Proceedings of the National Academy of Sciences},
  publisher = {Proceedings of the National Academy of Sciences},
  author = {Fedorenko,  Evelina and Duncan,  John and Kanwisher,  Nancy},
  year = {2013},
  month = sep,
  pages = {16616–16621}
}

@article{hendrycksmath2021,
  title={Measuring Mathematical Problem Solving With the MATH Dataset},
  author={Dan Hendrycks and Collin Burns and Saurav Kadavath and Akul Arora and Steven Basart and Eric Tang and Dawn Song and Jacob Steinhardt},
  journal={NeurIPS},
  year={2021}
}

@article{duncan2010multiple,
  title={The multiple-demand (MD) system of the primate brain: mental programs for intelligent behaviour},
  author={Duncan, John},
  journal={Trends in cognitive sciences},
  volume={14},
  number={4},
  pages={172--179},
  year={2010},
  publisher={Elsevier}
}

@article{Duncan2000,
  title = {Common regions of the human frontal lobe recruited by diverse cognitive demands},
  volume = {23},
  ISSN = {0166-2236},
  url = {http://dx.doi.org/10.1016/S0166-2236(00)01633-7},
  DOI = {10.1016/s0166-2236(00)01633-7},
  number = {10},
  journal = {Trends in Neurosciences},
  publisher = {Elsevier BV},
  author = {Duncan,  John and Owen,  Adrian M},
  year = {2000},
  month = oct,
  pages = {475–483}
}

@article{Woolgar2010,
  title = {Fluid intelligence loss linked to restricted regions of damage within frontal and parietal cortex},
  volume = {107},
  ISSN = {1091-6490},
  url = {http://dx.doi.org/10.1073/pnas.1007928107},
  DOI = {10.1073/pnas.1007928107},
  number = {33},
  journal = {Proceedings of the National Academy of Sciences},
  publisher = {Proceedings of the National Academy of Sciences},
  author = {Woolgar,  Alexandra and Parr,  Alice and Cusack,  Rhodri and Thompson,  Russell and Nimmo-Smith,  Ian and Torralva,  Teresa and Roca,  Maria and Antoun,  Nagui and Manes,  Facundo and Duncan,  John},
  year = {2010},
  month = aug,
  pages = {14899–14902}
}

@article{Saxe2003,
  title = {People thinking about thinking peopleThe role of the temporo-parietal junction in “theory of mind”},
  volume = {19},
  ISSN = {1053-8119},
  url = {http://dx.doi.org/10.1016/s1053-8119(03)00230-1},
  DOI = {10.1016/s1053-8119(03)00230-1},
  number = {4},
  journal = {NeuroImage},
  publisher = {Elsevier BV},
  author = {Saxe,  R and Kanwisher,  N},
  year = {2003},
  month = aug,
  pages = {1835–1842}
}

@article{Gallagher2000,
  title = {Reading the mind in cartoons and stories: an fMRI study of ‘theory of mind’ in verbal and nonverbal tasks},
  volume = {38},
  ISSN = {0028-3932},
  url = {http://dx.doi.org/10.1016/s0028-3932(99)00053-6},
  DOI = {10.1016/s0028-3932(99)00053-6},
  number = {1},
  journal = {Neuropsychologia},
  publisher = {Elsevier BV},
  author = {Gallagher,  H.L and Happé,  F and Brunswick,  N and Fletcher,  P.C and Frith,  U and Frith,  C.D},
  year = {2000},
  month = jan,
  pages = {11–21}
}

@article{Saxe2006,
  title = {It’s the Thought That Counts: Specific Brain Regions for One Component of Theory of Mind},
  volume = {17},
  ISSN = {1467-9280},
  url = {http://dx.doi.org/10.1111/j.1467-9280.2006.01768.x},
  DOI = {10.1111/j.1467-9280.2006.01768.x},
  number = {8},
  journal = {Psychological Science},
  publisher = {SAGE Publications},
  author = {Saxe,  Rebecca and Powell,  Lindsey J.},
  year = {2006},
  month = aug,
  pages = {692–699}
}

@article{KosterHale2013,
  title = {Theory of Mind: A Neural Prediction Problem},
  volume = {79},
  ISSN = {0896-6273},
  url = {http://dx.doi.org/10.1016/j.neuron.2013.08.020},
  DOI = {10.1016/j.neuron.2013.08.020},
  number = {5},
  journal = {Neuron},
  publisher = {Elsevier BV},
  author = {Koster-Hale,  Jorie and Saxe,  Rebecca},
  year = {2013},
  month = sep,
  pages = {836–848}
}

@article{Bayazit2023DiscoveringKS,
  title={Discovering Knowledge-Critical Subnetworks in Pretrained Language Models},
  author={Deniz Bayazit and Negar Foroutan and Zeming Chen and Gail Weiss and Antoine Bosselut},
  journal={ArXiv},
  year={2023},
  volume={abs/2310.03084},
  url={https://api.semanticscholar.org/CorpusID:263671765}
}

@article{tuckute2024language,
  title={Language in brains, minds, and machines},
  author={Tuckute, Greta and Kanwisher, Nancy and Fedorenko, Evelina},
  journal={Annual Review of Neuroscience},
  volume={47},
  year={2024},
  publisher={Annual Reviews}
}

@article{schrimpf2020integrative,
  title={Integrative benchmarking to advance neurally mechanistic models of human intelligence},
  author={Schrimpf, Martin and Kubilius, Jonas and Lee, Michael J and Murty, N Apurva Ratan and Ajemian, Robert and DiCarlo, James J},
  journal={Neuron},
  volume={108},
  number={3},
  pages={413--423},
  year={2020},
  publisher={Elsevier}
}

@article{Buckner2008,
  title = {The Brain’s Default Network: Anatomy,  Function,  and Relevance to Disease},
  volume = {1124},
  ISSN = {1749-6632},
  url = {http://dx.doi.org/10.1196/annals.1440.011},
  DOI = {10.1196/annals.1440.011},
  number = {1},
  journal = {Annals of the New York Academy of Sciences},
  publisher = {Wiley},
  author = {Buckner,  Randy L. and Andrews‐Hanna,  Jessica R. and Schacter,  Daniel L.},
  year = {2008},
  month = mar,
  pages = {1–38}
}

@article{Buckner2019,
  title = {The brain’s default network: updated anatomy,  physiology and evolving insights},
  volume = {20},
  ISSN = {1471-0048},
  url = {http://dx.doi.org/10.1038/s41583-019-0212-7},
  DOI = {10.1038/s41583-019-0212-7},
  number = {10},
  journal = {Nature Reviews Neuroscience},
  publisher = {Springer Science and Business Media LLC},
  author = {Buckner,  Randy L. and DiNicola,  Lauren M.},
  year = {2019},
  month = sep,
  pages = {593–608}
}

@article{Gusnard2001,
  title = {Medial prefrontal cortex and self-referential mental activity: Relation to a default mode of brain function},
  volume = {98},
  ISSN = {1091-6490},
  url = {http://dx.doi.org/10.1073/pnas.071043098},
  DOI = {10.1073/pnas.071043098},
  number = {7},
  journal = {Proceedings of the National Academy of Sciences},
  publisher = {Proceedings of the National Academy of Sciences},
  author = {Gusnard,  Debra A. and Akbudak,  Erbil and Shulman,  Gordon L. and Raichle,  Marcus E.},
  year = {2001},
  month = mar,
  pages = {4259–4264}
}

@article{jacoby2020discourse,
  title={Discourse-level comprehension engages medial frontal Theory of Mind brain regions even for expository texts},
  author={Jacoby, Nir and Fedorenko, Evelina},
  journal={Language, Cognition and Neuroscience},
  volume={35},
  number={6},
  pages={780--796},
  year={2020},
  publisher={Taylor \& Francis}
}

@article{binhuraib2025topoformer,
  title={Topoformer: brain-like topographic organization in transformer language models through spatial querying and reweighting},
  author={Binhuraib, Taha and Tuckute, Greta and Blauch, Nicholas},
  journal={arXiv preprint arXiv:2510.18745},
  year={2025}
}

@article{ferstl2002does,
  title={What does the frontomedian cortex contribute to language processing: coherence or theory of mind?},
  author={Ferstl, Evelyn C and von Cramon, D Yves},
  journal={Neuroimage},
  volume={17},
  number={3},
  pages={1599--1612},
  year={2002},
  publisher={Elsevier}
}

@inproceedings{
    komatsuzaki2023sparseupcycling,
    title={Sparse Upcycling: Training Mixture-of-Experts from Dense Checkpoints},
    author={Aran Komatsuzaki and Joan Puigcerver and James Lee-Thorp and Carlos Riquelme Ruiz and Basil Mustafa and Joshua Ainslie and Yi Tay and Mostafa Dehghani and Neil Houlsby},
    booktitle={The Eleventh International Conference on Learning Representations },
    year={2023},
    url={https://openreview.net/forum?id=T5nUQDrM4u}
}

@inproceedings{
    zhang2024bam,
    title={{BAM}! Just Like That: Simple and Efficient Parameter Upcycling for Mixture of Experts},
    author={Qizhen Zhang and Nikolas Gritsch and Dwaraknath Gnaneshwar and Simon Guo and David Cairuz and Bharat Venkitesh and Jakob Nicolaus Foerster and Phil Blunsom and Sebastian Ruder and Ahmet {\"U}st{\"u}n and Acyr Locatelli},
    booktitle={The Thirty-eighth Annual Conference on Neural Information Processing Systems},
    year={2024},
    url={https://openreview.net/forum?id=BDrWQTrfyI}
}

@article{tulu3,
  title={T{\"U}LU 3: Pushing Frontiers in Open Language Model Post-Training},
  author={Nathan Lambert and Jacob Daniel Morrison and Valentina Pyatkin and Shengyi Huang and Hamish Ivison and Faeze Brahman and Lester James Validad Miranda and Alisa Liu and Nouha Dziri and Xinxi Lyu and Yuling Gu and Saumya Malik and Victoria Graf and Jena D. Hwang and Jiangjiang Yang and Ronan Le Bras and Oyvind Tafjord and Chris Wilhelm and Luca Soldaini and Noah A. Smith and Yizhong Wang and Pradeep Dasigi and Hanna Hajishirzi},
  journal={ArXiv},
  year={2024},
  volume={abs/2411.15124},
  url={https://api.semanticscholar.org/CorpusID:274192505}
}

@article{jaech2024openaio1,
  title={Openai o1 system card},
  author={Jaech, Aaron and Kalai, Adam and Lerer, Adam and Richardson, Adam and El-Kishky, Ahmed and Low, Aiden and Helyar, Alec and Madry, Aleksander and Beutel, Alex and Carney, Alex and others},
  journal={arXiv preprint arXiv:2412.16720},
  year={2024}
}

@article{hurst2024gpt4o,
  title={Gpt-4o system card},
  author={Hurst, Aaron and Lerer, Adam and Goucher, Adam P and Perelman, Adam and Ramesh, Aditya and Clark, Aidan and Ostrow, AJ and Welihinda, Akila and Hayes, Alan and Radford, Alec and others},
  journal={arXiv preprint arXiv:2410.21276},
  year={2024}
}

@misc{olmo20242olmo2furious,
      title={{2 OLMo 2 Furious}},
      author={Team OLMo and Pete Walsh and Luca Soldaini and Dirk Groeneveld and Kyle Lo and Shane Arora and Akshita Bhagia and Yuling Gu and Shengyi Huang and Matt Jordan and Nathan Lambert and Dustin Schwenk and Oyvind Tafjord and Taira Anderson and David Atkinson and Faeze Brahman and Christopher Clark and Pradeep Dasigi and Nouha Dziri and Michal Guerquin and Hamish Ivison and Pang Wei Koh and Jiacheng Liu and Saumya Malik and William Merrill and Lester James V. Miranda and Jacob Morrison and Tyler Murray and Crystal Nam and Valentina Pyatkin and Aman Rangapur and Michael Schmitz and Sam Skjonsberg and David Wadden and Christopher Wilhelm and Michael Wilson and Luke Zettlemoyer and Ali Farhadi and Noah A. Smith and Hannaneh Hajishirzi},
      year={2024},
      eprint={2501.00656},
      archivePrefix={arXiv},
      primaryClass={cs.CL},
      url={https://arxiv.org/abs/2501.00656},
}

@inproceedings{alkhamissi-etal-2025-llm-language-network,
    title = "The {LLM} Language Network: A Neuroscientific Approach for Identifying Causally Task-Relevant Units",
    author = "AlKhamissi, Badr  and
      Tuckute, Greta  and
      Bosselut, Antoine  and
      Schrimpf, Martin",
    editor = "Chiruzzo, Luis  and
      Ritter, Alan  and
      Wang, Lu",
    booktitle = "Proceedings of the 2025 Conference of the Nations of the Americas Chapter of the Association for Computational Linguistics: Human Language Technologies (Volume 1: Long Papers)",
    month = apr,
    year = "2025",
    address = "Albuquerque, New Mexico",
    publisher = "Association for Computational Linguistics",
    url = "https://aclanthology.org/2025.naacl-long.544/",
    pages = "10887--10911",
    ISBN = "979-8-89176-189-6"
}

@article{Kanwisher2010,
  title = {Functional specificity in the human brain: A window into the functional architecture of the mind},
  volume = {107},
  ISSN = {1091-6490},
  url = {http://dx.doi.org/10.1073/pnas.1005062107},
  DOI = {10.1073/pnas.1005062107},
  number = {25},
  journal = {Proceedings of the National Academy of Sciences},
  publisher = {Proceedings of the National Academy of Sciences},
  author = {Kanwisher,  Nancy},
  year = {2010},
  month = may,
  pages = {11163–11170}
}

@article{cobbe2021gsm8k,
  title={Training Verifiers to Solve Math Word Problems},
  author={Cobbe, Karl and Kosaraju, Vineet and Bavarian, Mohammad and Chen, Mark and Jun, Heewoo and Kaiser, Lukasz and Plappert, Matthias and Tworek, Jerry and Hilton, Jacob and Nakano, Reiichiro and Hesse, Christopher and Schulman, John},
  journal={arXiv preprint arXiv:2110.14168},
  year={2021}
}

@inproceedings{buechel-etal-2018-modeling-empathy,
    title = "Modeling Empathy and Distress in Reaction to News Stories",
    author = "Buechel, Sven  and
      Buffone, Anneke  and
      Slaff, Barry  and
      Ungar, Lyle  and
      Sedoc, Jo{\~a}o",
    editor = "Riloff, Ellen  and
      Chiang, David  and
      Hockenmaier, Julia  and
      Tsujii, Jun{'}ichi",
    booktitle = "Proceedings of the 2018 Conference on Empirical Methods in Natural Language Processing",
    month = oct # "-" # nov,
    year = "2018",
    address = "Brussels, Belgium",
    publisher = "Association for Computational Linguistics",
    url = "https://aclanthology.org/D18-1507/",
    doi = "10.18653/v1/D18-1507",
    pages = "4758--4765"
}

@inproceedings{hendrycks2021mmlu,
    title={Measuring Massive Multitask Language Understanding},
    author={Dan Hendrycks and Collin Burns and Steven Basart and Andy Zou and Mantas Mazeika and Dawn Song and Jacob Steinhardt},
    booktitle={International Conference on Learning Representations},
    year={2021},
    url={https://openreview.net/forum?id=d7KBjmI3GmQ}
}

@article{suzgun2022challenging,
  title={Challenging BIG-Bench Tasks and Whether Chain-of-Thought Can Solve Them},
  author={Suzgun, Mirac and Scales, Nathan and Sch{\"a}rli, Nathanael and Gehrmann, Sebastian and Tay, Yi and Chung, Hyung Won and Chowdhery, Aakanksha and Le, Quoc V and Chi, Ed H and Zhou, Denny and and Wei, Jason},
  journal={arXiv preprint arXiv:2210.09261},
  year={2022}
}

@article{Kean2025,
  title = {Intuitive physical reasoning is not mediated by linguistic nor exclusively domain-general abstract representations},
  volume = {213},
  ISSN = {0028-3932},
  url = {http://dx.doi.org/10.1016/j.neuropsychologia.2025.109125},
  DOI = {10.1016/j.neuropsychologia.2025.109125},
  journal = {Neuropsychologia},
  publisher = {Elsevier BV},
  author = {Kean,  Hope H. and Fung,  Alexander and Pramod,  R.T. and Chomik-Morales,  Jessica and Kanwisher,  Nancy and Fedorenko,  Evelina},
  year = {2025},
  month = jul,
  pages = {109125}
}

@article{
    pfeiffer2023modular,
    title={Modular Deep Learning},
    author={Jonas Pfeiffer and Sebastian Ruder and Ivan Vuli{\'c} and Edoardo Ponti},
    journal={Transactions on Machine Learning Research},
    issn={2835-8856},
    year={2023},
    url={https://openreview.net/forum?id=z9EkXfvxta},
    note={Survey Certification}
}

@inproceedings{gururangan-etal-2022-demix,
    title = "{DEM}ix Layers: Disentangling Domains for Modular Language Modeling",
    author = "Gururangan, Suchin  and
      Lewis, Mike  and
      Holtzman, Ari  and
      Smith, Noah A.  and
      Zettlemoyer, Luke",
    editor = "Carpuat, Marine  and
      de Marneffe, Marie-Catherine  and
      Meza Ruiz, Ivan Vladimir",
    booktitle = "Proceedings of the 2022 Conference of the North American Chapter of the Association for Computational Linguistics: Human Language Technologies",
    month = jul,
    year = "2022",
    address = "Seattle, United States",
    publisher = "Association for Computational Linguistics",
    url = "https://aclanthology.org/2022.naacl-main.407/",
    doi = "10.18653/v1/2022.naacl-main.407",
    pages = "5557--5576"
}

@inproceedings{
    shazeer2017,
    title={ Outrageously Large Neural Networks: The Sparsely-Gated Mixture-of-Experts Layer},
    author={Noam Shazeer and *Azalia Mirhoseini and *Krzysztof Maziarz and Andy Davis and Quoc Le and Geoffrey Hinton and Jeff Dean},
    booktitle={International Conference on Learning Representations},
    year={2017},
    url={https://openreview.net/forum?id=B1ckMDqlg}
}

@inproceedings{Shen2023ModuleFormerME,
  title={ModuleFormer: Modularity Emerges from Mixture-of-Experts},
  author={Yikang Shen and Zheyu Zhang and Tianyou Cao and Shawn Tan and Zhenfang Chen and Chuang Gan},
  year={2023},
  url={https://api.semanticscholar.org/CorpusID:261697418}
}

@inproceedings{
swamy2023multimodnmultimodal,
title={MultiMo{DN}{\textemdash}Multimodal, Multi-Task, Interpretable Modular Networks},
author={Vinitra Swamy and Malika Satayeva and Jibril Frej and Thierry Bossy and Thijs Vogels and Martin Jaggi and Tanja K{\"a}ser and Mary-Anne Hartley},
booktitle={Thirty-seventh Conference on Neural Information Processing Systems},
year={2023},
url={https://openreview.net/forum?id=iB3Ew6z4WL}
}

@inproceedings{pfeiffer-etal-2020-mad,
    title = "{MAD-X}: {A}n {A}dapter-{B}ased {F}ramework for {M}ulti-{T}ask {C}ross-{L}ingual {T}ransfer",
    author = "Pfeiffer, Jonas  and
      Vuli{\'c}, Ivan  and
      Gurevych, Iryna  and
      Ruder, Sebastian",
    editor = "Webber, Bonnie  and
      Cohn, Trevor  and
      He, Yulan  and
      Liu, Yang",
    booktitle = "Proceedings of the 2020 Conference on Empirical Methods in Natural Language Processing (EMNLP)",
    month = nov,
    year = "2020",
    address = "Online",
    publisher = "Association for Computational Linguistics",
    url = "https://aclanthology.org/2020.emnlp-main.617/",
    doi = "10.18653/v1/2020.emnlp-main.617",
    pages = "7654--7673"
}

@inproceedings{pfeiffer-etal-2022-lifting,
    title = "Lifting the Curse of Multilinguality by Pre-training Modular Transformers",
    author = "Pfeiffer, Jonas  and
      Goyal, Naman  and
      Lin, Xi  and
      Li, Xian  and
      Cross, James  and
      Riedel, Sebastian  and
      Artetxe, Mikel",
    editor = "Carpuat, Marine  and
      de Marneffe, Marie-Catherine  and
      Meza Ruiz, Ivan Vladimir",
    booktitle = "Proceedings of the 2022 Conference of the North American Chapter of the Association for Computational Linguistics: Human Language Technologies",
    month = jul,
    year = "2022",
    address = "Seattle, United States",
    publisher = "Association for Computational Linguistics",
    url = "https://aclanthology.org/2022.naacl-main.255/",
    doi = "10.18653/v1/2022.naacl-main.255",
    pages = "3479--3495"
}

@article{AlMaamari2024MixtureOM,
  title={Mixture of Modular Experts: Distilling Knowledge from a Multilingual Teacher into Specialized Modular Language Models},
  author={Mohammed Al-Maamari and Mehdi Ben Amor and Michael Granitzer},
  journal={ArXiv},
  year={2024},
  volume={abs/2407.19610},
  url={https://api.semanticscholar.org/CorpusID:271533631}
}

@inproceedings{
    zhong2022metadmoe,
    title={Meta-{DM}oE: Adapting to Domain Shift by Meta-Distillation from Mixture-of-Experts},
    author={Tao Zhong and Zhixiang Chi and Li Gu and Yang Wang and Yuanhao Yu and Jin Tang},
    booktitle={Thirty-Sixth Conference on Neural Information Processing Systems (NeurIPS)},
    year={2022}
}

@misc{ye2023mplugowl2,
      title={mPLUG-Owl2: Revolutionizing Multi-modal Large Language Model with Modality Collaboration}, 
      author={Qinghao Ye and Haiyang Xu and Jiabo Ye and Ming Yan and Anwen Hu and Haowei Liu and Qi Qian and Ji Zhang and Fei Huang and Jingren Zhou},
      year={2023},
      eprint={2311.04257},
      archivePrefix={arXiv},
      primaryClass={cs.CL}
}

@misc{liu2023llava,
      title={Visual Instruction Tuning}, 
      author={Liu, Haotian and Li, Chunyuan and Wu, Qingyang and Lee, Yong Jae},
      publisher={NeurIPS},
      year={2023},
}

@inproceedings{si-etal-2023-getting,
    title = "Getting {M}o{RE} out of Mixture of Language Model Reasoning Experts",
    author = "Si, Chenglei  and
      Shi, Weijia  and
      Zhao, Chen  and
      Zettlemoyer, Luke  and
      Boyd-Graber, Jordan",
    editor = "Bouamor, Houda  and
      Pino, Juan  and
      Bali, Kalika",
    booktitle = "Findings of the Association for Computational Linguistics: EMNLP 2023",
    month = dec,
    year = "2023",
    address = "Singapore",
    publisher = "Association for Computational Linguistics",
    url = "https://aclanthology.org/2023.findings-emnlp.552/",
    doi = "10.18653/v1/2023.findings-emnlp.552",
    pages = "8234--8249"
}

@article{alkhamissi2025langtocog,
  title={From language to cognition: How llms outgrow the human language network},
  author={AlKhamissi, Badr and Tuckute, Greta and Tang, Yingtian and Binhuraib, Taha and Bosselut, Antoine and Schrimpf, Martin},
  journal={arXiv preprint arXiv:2503.01830},
  year={2025}
}

@inproceedings{KubiliusSchrimpf2019neurips,
  title={Brain-Like Object Recognition with High-Performing Shallow Recurrent ANNs},
  author={Kubilius, Jonas and Schrimpf, Martin and Kar, Kohitij and Hong, Ha and Majaj, Najib J and Rajalingham, Rishi and Issa, Elias B and Bashivan, Pouya and Prescott-Roy, Jonathan and Schmidt, Kailyn and Nayebi, Aran and Bear, Daniel and Yamins, Daniel L K and DiCarlo, James J},
  booktitle={Advances in Neural Information Processing Systems},
  year={2019}
}

@inproceedings{
    rathi2025topolm,
    title={Topo{LM}: brain-like spatio-functional organization in a topographic language model},
    author={Neil Rathi and Johannes Mehrer and Badr AlKhamissi and Taha Osama A Binhuraib and Nicholas Blauch and Martin Schrimpf},
    booktitle={The Thirteenth International Conference on Learning Representations},
    year={2025},
    url={https://openreview.net/forum?id=aWXnKanInf}
}

@article{Margalit2024,
  title = {A unifying framework for functional organization in early and higher ventral visual cortex},
  volume = {112},
  ISSN = {0896-6273},
  url = {http://dx.doi.org/10.1016/j.neuron.2024.04.018},
  DOI = {10.1016/j.neuron.2024.04.018},
  number = {14},
  journal = {Neuron},
  publisher = {Elsevier BV},
  author = {Margalit,  Eshed and Lee,  Hyodong and Finzi,  Dawn and DiCarlo,  James J. and Grill-Spector,  Kalanit and Yamins,  Daniel L.K.},
  year = {2024},
  month = jul,
  pages = {2435--2451.e7}
}

@article{Spoerer2020,
  title = {Recurrent neural networks can explain flexible trading of speed and accuracy in biological vision},
  volume = {16},
  ISSN = {1553-7358},
  url = {http://dx.doi.org/10.1371/journal.pcbi.1008215},
  DOI = {10.1371/journal.pcbi.1008215},
  number = {10},
  journal = {PLOS Computational Biology},
  publisher = {Public Library of Science (PLoS)},
  author = {Spoerer,  Courtney J. and Kietzmann,  Tim C. and Mehrer,  Johannes and Charest,  Ian and Kriegeskorte,  Nikolaus},
  editor = {Isik,  Leyla},
  year = {2020},
  month = oct,
  pages = {e1008215}
}

@inproceedings{hu-etal-2023-fine,
    title = "A fine-grained comparison of pragmatic language understanding in humans and language models",
    author = "Hu, Jennifer  and
      Floyd, Sammy  and
      Jouravlev, Olessia  and
      Fedorenko, Evelina  and
      Gibson, Edward",
    editor = "Rogers, Anna  and
      Boyd-Graber, Jordan  and
      Okazaki, Naoaki",
    booktitle = "Proceedings of the 61st Annual Meeting of the Association for Computational Linguistics (Volume 1: Long Papers)",
    month = jul,
    year = "2023",
    address = "Toronto, Canada",
    publisher = "Association for Computational Linguistics",
    url = "https://aclanthology.org/2023.acl-long.230/",
    doi = "10.18653/v1/2023.acl-long.230",
    pages = "4194--4213"
}

@inproceedings{Kim:2021:empathy,
  title={Perspective-taking and Pragmatics for Generating Empathetic Responses Focused on Emotion Causes},
  author={Kim, Hyunwoo and Kim, Byeongchang and Kim, Gunhee},
  booktitle={EMNLP},
  year=2021
}

@inproceedings{kim2023fantom,
    title={FANToM: A Benchmark for Stress-testing Machine Theory of Mind in Interactions},
    author={Hyunwoo Kim and Melanie Sclar and Xuhui Zhou and Ronan Le Bras and Gunhee Kim and Yejin Choi and Maarten Sap},
    booktitle={EMNLP},
    year=2023
}

@misc{li2023camel,
      title={CAMEL: Communicative Agents for "Mind" Exploration of Large Scale Language Model Society}, 
      author={Guohao Li and Hasan Abed Al Kader Hammoud and Hani Itani and Dmitrii Khizbullin and Bernard Ghanem},
      year={2023},
      eprint={2303.17760},
      archivePrefix={arXiv},
      primaryClass={cs.AI}
}

@article{liu2020logiqa,
  title={Logiqa: A challenge dataset for machine reading comprehension with logical reasoning},
  author={Liu, Jian and Cui, Leyang and Liu, Hanmeng and Huang, Dandan and Wang, Yile and Zhang, Yue},
  journal={arXiv preprint arXiv:2007.08124},
  year={2020}
}

@article{han2022folio,
  title={FOLIO: Natural Language Reasoning with First-Order Logic},
  author = {Han, Simeng and Schoelkopf, Hailey and Zhao, Yilun and Qi, Zhenting and Riddell, Martin and Benson, Luke and Sun, Lucy and Zubova, Ekaterina and Qiao, Yujie and Burtell, Matthew and Peng, David and Fan, Jonathan and Liu, Yixin and Wong, Brian and Sailor, Malcolm and Ni, Ansong and Nan, Linyong and Kasai, Jungo and Yu, Tao and Zhang, Rui and Joty, Shafiq and Fabbri, Alexander R. and Kryscinski, Wojciech and Lin, Xi Victoria and Xiong, Caiming and Radev, Dragomir},
  journal={arXiv preprint arXiv:2209.00840},
  url = {https://arxiv.org/abs/2209.00840},
  year={2022}
}

@inproceedings{Bisk2020,
  author = {Yonatan Bisk and Rowan Zellers and
            Ronan Le Bras and Jianfeng Gao
            and Yejin Choi},
  title = {PIQA: Reasoning about Physical Commonsense in
           Natural Language},
  booktitle = {Thirty-Fourth AAAI Conference on
               Artificial Intelligence},
  year = {2020},
}

@inproceedings{wang2024spatial,
    title={Is A Picture Worth A Thousand Words? Delving Into Spatial Reasoning for Vision Language Models},
    author={Wang, Jiayu and Ming, Yifei and Shi, Zhenmei and Vineet, Vibhav and Wang, Xin and Li, Yixuan and Joshi, Neel},
    booktitle={The Thirty-Eighth Annual Conference on Neural Information Processing Systems},
    year={2024}
}

@inproceedings{
    li2025temporal,
    title={Temporal Reasoning Transfer from Text to Video},
    author={Lei Li and Yuanxin Liu and Linli Yao and Peiyuan Zhang and Chenxin An and Lean Wang and Xu Sun and Lingpeng Kong and Qi Liu},
    booktitle={The Thirteenth International Conference on Learning Representations},
    year={2025},
    url={https://openreview.net/forum?id=sHAvMp5J4R}
}

@misc{wang2022supernaturalinstructionsgeneralizationdeclarativeinstructions,
    title={Super-NaturalInstructions: Generalization via Declarative Instructions on 1600+ NLP Tasks}, 
    author={Yizhong Wang and Swaroop Mishra and Pegah Alipoormolabashi and Yeganeh Kordi and Amirreza Mirzaei and Anjana Arunkumar and Arjun Ashok and Arut Selvan Dhanasekaran and Atharva Naik and David Stap and Eshaan Pathak and Giannis Karamanolakis and Haizhi Gary Lai and Ishan Purohit and Ishani Mondal and Jacob Anderson and Kirby Kuznia and Krima Doshi and Maitreya Patel and Kuntal Kumar Pal and Mehrad Moradshahi and Mihir Parmar and Mirali Purohit and Neeraj Varshney and Phani Rohitha Kaza and Pulkit Verma and Ravsehaj Singh Puri and Rushang Karia and Shailaja Keyur Sampat and Savan Doshi and Siddhartha Mishra and Sujan Reddy and Sumanta Patro and Tanay Dixit and Xudong Shen and Chitta Baral and Yejin Choi and Noah A. Smith and Hannaneh Hajishirzi and Daniel Khashabi},
    year={2022},
    eprint={2204.07705},
    archivePrefix={arXiv},
    primaryClass={cs.CL},
    url={https://arxiv.org/abs/2204.07705}, 
}

@inproceedings{yang-etal-2015-wikiqa,
    title = "{W}iki{QA}: A Challenge Dataset for Open-Domain Question Answering",
    author = "Yang, Yi  and
      Yih, Wen-tau  and
      Meek, Christopher",
    booktitle = "Proceedings of the 2015 Conference on Empirical Methods in Natural Language Processing",
    month = sep,
    year = "2015",
    address = "Lisbon, Portugal",
    publisher = "Association for Computational Linguistics",
    url = "https://aclanthology.org/D15-1237",
    doi = "10.18653/v1/D15-1237",
    pages = "2013--2018",
}

@misc{o1journey,
  author = {Yiwei Qin and Xuefeng Li and Haoyang Zou and Yixiu Liu and Shijie Xia and Zhen Huang and Yixin Ye and Weizhe Yuan and Zhengzhong Liu and Yuanzhi Li and Pengfei Liu},
  title = {O1 Replication Journey: A Strategic Progress Report – Part 1},
  year = {2024},
  publisher = {GitHub},
  journal = {GitHub repository},
  howpublished = {\url{https://github.com/GAIR-NLP/O1-Journey}},
}

@inproceedings{gandhi2023understanding,
    title={Understanding Social Reasoning in Language Models with Language Models},
    author={Kanishk Gandhi and Jan-Philipp Fr{\"a}nken and Tobias Gerstenberg and Noah Goodman},
    booktitle={Thirty-seventh Conference on Neural Information Processing Systems Datasets and Benchmarks Track},
    year={2023},
    url={https://openreview.net/forum?id=8bqjirgxQM}
}

@misc{eval-harness,
  author       = {Gao, Leo and Tow, Jonathan and Abbasi, Baber and Biderman, Stella and Black, Sid and DiPofi, Anthony and Foster, Charles and Golding, Laurence and Hsu, Jeffrey and Le Noac'h, Alain and Li, Haonan and McDonell, Kyle and Muennighoff, Niklas and Ociepa, Chris and Phang, Jason and Reynolds, Laria and Schoelkopf, Hailey and Skowron, Aviya and Sutawika, Lintang and Tang, Eric and Thite, Anish and Wang, Ben and Wang, Kevin and Zou, Andy},
  title        = {The Language Model Evaluation Harness},
  month        = 07,
  year         = 2024,
  publisher    = {Zenodo},
  version      = {v0.4.3},
  doi          = {10.5281/zenodo.12608602},
  url          = {https://zenodo.org/records/12608602}
}

@inproceedings{dpo,
    author = {Rafailov, Rafael and Sharma, Archit and Mitchell, Eric and Manning, Christopher D and Ermon, Stefano and Finn, Chelsea},
    booktitle = {Advances in Neural Information Processing Systems},
    editor = {A. Oh and T. Naumann and A. Globerson and K. Saenko and M. Hardt and S. Levine},
    pages = {53728--53741},
    publisher = {Curran Associates, Inc.},
    title = {Direct Preference Optimization: Your Language Model is Secretly a Reward Model},
    url = {https://proceedings.neurips.cc/paper_files/paper/2023/file/a85b405ed65c6477a4fe8302b5e06ce7-Paper-Conference.pdf},
    volume = {36},
    year = {2023}
}

@misc{meditron,
    author = {Bosselut, Antoine and Chen, Zeming and Romanou, Angelika and Bonnet, Antoine and Hernández-Cano, Alejandro and Alkhamissi, Badr and Matoba, Kyle and Salvi, Francesco and Pagliardini, Matteo and Fan, Simin and Köpf, Andreas and Mohtashami, Amirkeivan and Sallinen, Alexandre and Swamy, Vinitra and Sakhaeirad, Alireza and Krawczuk, Igor and Bayazit, Deniz and Marmet, Axel and Mi, Li and Jaggi, Martin},
    year = {2024},
    month = {03},
    pages = {},
    title = {MEDITRON: Open Medical Foundation Models Adapted for Clinical Practice},
    doi = {10.21203/rs.3.rs-4139743/v1}
}

@misc{ivison2024unpacking,
      title={Unpacking DPO and PPO: Disentangling Best Practices for Learning from Preference Feedback}, 
      author={Hamish Ivison and Yizhong Wang and Jiacheng Liu and Ellen Wu and Valentina Pyatkin and Nathan Lambert and Yejin Choi and Noah A. Smith and Hannaneh Hajishirzi},
      year={2024},
      eprint={2406.09279},
      archivePrefix={arXiv},
      primaryClass={cs.CL}
}

@article{Zhang2025MixtureOE,
  title={Mixture of Experts in Large Language Models},
  author={Danyang Zhang and Junhao Song and Ziqian Bi and Yingfang Yuan and Tianyang Wang and Joe Yeong and Junfeng Hao},
  journal={ArXiv},
  year={2025},
  volume={abs/2507.11181},
  url={https://api.semanticscholar.org/CorpusID:280277258}
}

@article{blank2020no,
  title={No evidence for differences among language regions in their temporal receptive windows},
  author={Blank, Idan A and Fedorenko, Evelina},
  journal={NeuroImage},
  volume={219},
  pages={116925},
  year={2020},
  publisher={Elsevier}
}

@article{allal2025smollm2,
  title={SmolLM2: When Smol Goes Big--Data-Centric Training of a Small Language Model},
  author={Allal, Loubna Ben and Lozhkov, Anton and Bakouch, Elie and Bl{\'a}zquez, Gabriel Mart{\'\i}n and Penedo, Guilherme and Tunstall, Lewis and Marafioti, Andr{\'e}s and Kydl{\'\i}{\v{c}}ek, Hynek and Lajar{\'\i}n, Agust{\'\i}n Piqueres and Srivastav, Vaibhav and others},
  journal={arXiv preprint arXiv:2502.02737},
  year={2025}
}

@InProceedings{cogbench,
  title = 	 {{C}og{B}ench: a large language model walks into a psychology lab},
  author =       {Coda-Forno, Julian and Binz, Marcel and Wang, Jane X and Schulz, Eric},
  booktitle = 	 {Proceedings of the 41st International Conference on Machine Learning},
  pages = 	 {9076--9108},
  year = 	 {2024},
  editor = 	 {Salakhutdinov, Ruslan and Kolter, Zico and Heller, Katherine and Weller, Adrian and Oliver, Nuria and Scarlett, Jonathan and Berkenkamp, Felix},
  volume = 	 {235},
  series = 	 {Proceedings of Machine Learning Research},
  month = 	 {21--27 Jul},
  publisher =    {PMLR},
  url = 	 {https://proceedings.mlr.press/v235/coda-forno24a.html}
}

@inproceedings{zhang-etal-2023-emergent,
    title = "Emergent Modularity in Pre-trained Transformers",
    author = "Zhang, Zhengyan  and
      Zeng, Zhiyuan  and
      Lin, Yankai  and
      Xiao, Chaojun  and
      Wang, Xiaozhi  and
      Han, Xu  and
      Liu, Zhiyuan  and
      Xie, Ruobing  and
      Sun, Maosong  and
      Zhou, Jie",
    editor = "Rogers, Anna  and
      Boyd-Graber, Jordan  and
      Okazaki, Naoaki",
    booktitle = "Findings of the Association for Computational Linguistics: ACL 2023",
    month = jul,
    year = "2023",
    address = "Toronto, Canada",
    publisher = "Association for Computational Linguistics",
    url = "https://aclanthology.org/2023.findings-acl.250/",
    doi = "10.18653/v1/2023.findings-acl.250",
    pages = "4066--4083"
}

@inproceedings{zhang-etal-2022-moefication,
    title = "{M}o{E}fication: Transformer Feed-forward Layers are Mixtures of Experts",
    author = "Zhang, Zhengyan  and
      Lin, Yankai  and
      Liu, Zhiyuan  and
      Li, Peng  and
      Sun, Maosong  and
      Zhou, Jie",
    editor = "Muresan, Smaranda  and
      Nakov, Preslav  and
      Villavicencio, Aline",
    booktitle = "Findings of the Association for Computational Linguistics: ACL 2022",
    month = may,
    year = "2022",
    address = "Dublin, Ireland",
    publisher = "Association for Computational Linguistics",
    url = "https://aclanthology.org/2022.findings-acl.71/",
    doi = "10.18653/v1/2022.findings-acl.71",
    pages = "877--890"
}

@article {Wang2025.aphasia,
	author = {Wang, Chengcheng and Fan, Zhiyu and Han, Zaizhu and Bi, Yanchao and Li, Jixing},
	title = {Emergent modularity in large language models: Insights from aphasia simulations},
	elocation-id = {2025.02.22.639416},
	year = {2025},
	doi = {10.1101/2025.02.22.639416},
	publisher = {Cold Spring Harbor Laboratory},
	URL = {https://www.biorxiv.org/content/early/2025/02/23/2025.02.22.639416},
	eprint = {https://www.biorxiv.org/content/early/2025/02/23/2025.02.22.639416.full.pdf},
	journal = {bioRxiv}
}

@article{fedus2022switch,
  title={Switch transformers: Scaling to trillion parameter models with simple and efficient sparsity},
  author={Fedus, William and Zoph, Barret and Shazeer, Noam},
  journal={Journal of Machine Learning Research},
  volume={23},
  number={120},
  pages={1--39},
  year={2022}
}

@article{jamaa2025evaluating,
  title={Evaluating Contrast Localizer for Identifying Causal Unitsin Social \& Mathematical Tasks in Language Models},
  author={Jamaa, Yassine and AlKhamissi, Badr and Ghosh, Satrajit and Schrimpf, Martin},
  journal={arXiv preprint arXiv:2508.08276},
  year={2025}
}

@inproceedings{DapelloMarques2020,
    title = {Simulating a {Primary} {Visual} {Cortex} at the {Front} of {CNNs} {Improves} {Robustness} to {Image} {Perturbations}},
    url = {https://www.biorxiv.org/content/10.1101/2020.06.16.154542v1},
    doi = {10.1101/2020.06.16.154542},
    urldate = {2020-06-18},
    booktitle = {Neural {Information} {Processing} {Systems} ({NeurIPS}, {Spotlight})},
    author = {Dapello, Joel and Marques, Tiago and Schrimpf, Martin and Geiger, Franziska and Cox, David D. and DiCarlo, James J.},
    month = jun,
    year = {2020},
}

@article {Kean2025.10.21.683445,
	author = {Kean, Hope and Fung, Alex and Ohams, Chiebuka and Chen, Jason and Rule, Josh and Tenenbaum, Joshua and Piantadosi, Steven and Fedorenko, Evelina},
	title = {A human brain network specialized for abstract formal reasoning},
	elocation-id = {2025.10.21.683445},
	year = {2025},
	doi = {10.1101/2025.10.21.683445},
	publisher = {Cold Spring Harbor Laboratory},
	URL = {https://www.biorxiv.org/content/early/2025/10/21/2025.10.21.683445},
	eprint = {https://www.biorxiv.org/content/early/2025/10/21/2025.10.21.683445.full.pdf},
	journal = {bioRxiv}
}
\bibliographystyle{iclr2026_conference}


\clearpage

\appendix

\section*{Appendix}

\section{Constructing the Experts Dataset}
\label{app:expert-dataset}

\paragraph{Datasets}
To construct the small curated datasets used in stages 1 and 2 of our training curriculum, we first identified existing datasets that align with the cognitive domain of each expert. Table~\ref{tab:stage-1-datasets} lists these datasets, the number of examples sampled from each, the corresponding high-level cognitive skill they were chosen to represent, and whether we used \modelname{O1} to generate responses or relied on the original reasoning chains provided with the dataset.

\begin{table}[htbp]
\centering
\caption{
\textbf{Datasets Used to Induce Specialization in Stage-1.}
Overview of datasets used to induce expert specialization during stages 1 and 2. Each dataset is aligned with a cognitive skill targeted by a specific expert. We indicate the number of examples sampled from each dataset and whether responses were generated using \modelname{O1} or taken directly from the dataset’s original reasoning chains.
}
\label{tab:stage-1-datasets}
\begin{tabular}{@{}lllll@{}}
\toprule
\textbf{Expert} & \textbf{Task} & \textbf{Dataset} & \textbf{\# Samples} & \textbf{Use O1} \\
\midrule
\multirow{6}{*}{Logic} & \multirow{3}{*}{Math} & O1-Journey \citep{o1journey} & 327 & No \\
 & & Math \citep{li2023camel} & 200 & No \\
 & & GSM8K \citep{cobbe2021gsm8k} & 100 & Yes \\
\cmidrule(lr){2-5}
 & \multirow{2}{*}{Logic} & Folio \citep{han2022folio} & 100 & Yes \\
 & & LogicQA \citep{liu2020logiqa} & 100 & Yes \\
\cmidrule(lr){2-5}
 & Physics & Physics \citep{li2023camel} & 200 & No \\
\midrule
\multicolumn{3}{@{}l}{\textbf{Total (Logic)}} & \textbf{1027} &  \\
\midrule
\multirow{12}{*}{Social} & \multirow{7}{*}{Pragmatics} 
 & Deceits \citep{hu-etal-2023-fine} & 20 & Yes \\
 & & Indirect Speech \citep{hu-etal-2023-fine} & 20 & Yes \\
 & & Irony \citep{hu-etal-2023-fine} & 25 & Yes \\
 & & Maxims \citep{hu-etal-2023-fine} & 19 & Yes \\
 & & Metaphor \citep{hu-etal-2023-fine} & 20 & Yes \\
 & & Humor \citep{hu-etal-2023-fine} & 25 & Yes \\
 & & Coherence \citep{hu-etal-2023-fine} & 40 & Yes \\
\cmidrule(lr){2-5}
 & Emotion Detection  & EmoCause \citep{Kim:2021:empathy} & 100 & Yes \\
\cmidrule(lr){2-5}
 & \multirow{3}{*}{Theory of Mind} & FanToM 1st Order \citep{kim2023fantom} & 100 & Yes \\
 & & FanToM 2nd Order \citep{kim2023fantom} & 100 & Yes \\
 & & BigToM \citep{gandhi2023understanding} & 128 & Yes \\
\cmidrule(lr){2-5}
 & Social Reasoning & Mixture \hyperlink{https://huggingface.co/datasets/ProlificAI/social-reasoning-rlhf}{ProlificAI/social-reasoning-rlhf} & 531 & Yes \\
\midrule
\multicolumn{3}{@{}l}{\textbf{Total (Social)}} & \textbf{1028} &  \\
\midrule
\multirow{8}{*}{World} & \multirow{4}{*}{World Knowledge} & Biology \citep{li2023camel} & 100 & No \\
 & & Chemistry \citep{li2023camel} & 100 & No \\
 & & PIQA \citep{Bisk2020} & 200 & Yes \\
 & & WikiQA \citep{yang-etal-2015-wikiqa} & 100 & Yes \\
\cmidrule(lr){2-5}
 & Spatial Reasoning & SpatialEval \citep{wang2024spatial} & 100 & Yes \\
\cmidrule(lr){2-5}
 & Temporal Reasoning &  TextTemporal \citep{li2025temporal} & 100 & Yes \\
\cmidrule(lr){2-5}
 & World Building & World Building \hyperlink{https://huggingface.co/datasets/archit11/worldbuilding}{archit11/worldbuilding} & 200 & No \\
\cmidrule(lr){2-5}
 & Cause Effect & CoPA \citep{wang2022supernaturalinstructionsgeneralizationdeclarativeinstructions} & 100 & Yes \\
\midrule
\multicolumn{3}{@{}l}{\textbf{Total (World)}} & \textbf{1000} &  \\
\bottomrule
\end{tabular}
\end{table}

\paragraph{Generating Reasoning Responses}
Since most of the existing datasets we identified (listed in Table~\ref{tab:stage-1-datasets}) do not include reasoning steps for the final answer, we used \modelname{O1} to generate them. Specifically, we prompted the model with the input followed by “Let's think step by step.” to elicit a longer response that includes intermediate reasoning before reaching the final answer. This was only done for datasets that did not already contain suitable reasoning chains.

\begin{figure}
    \centering
    \includegraphics[width=1\linewidth]{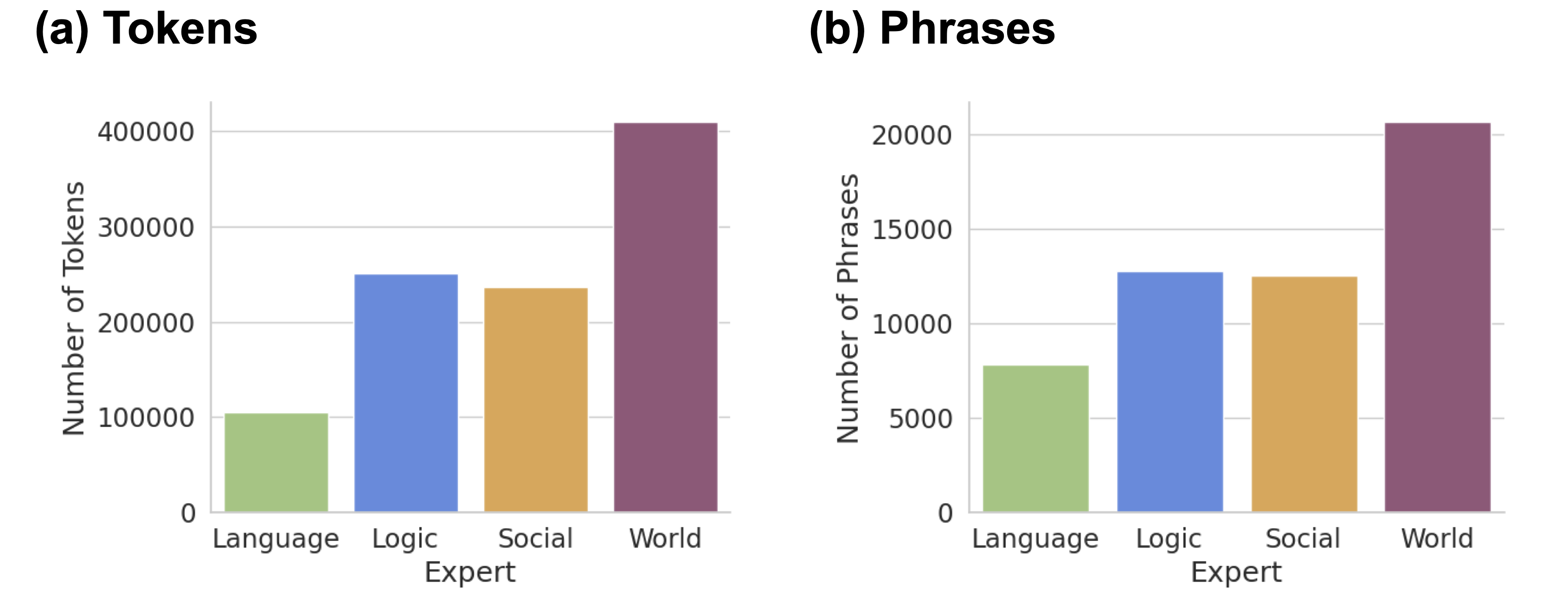}
    \caption{
        \textbf{Distribution of Expert Assignments across Tokens and Phrases.}
        \textbf{(a)} The distribution of expert assignments across tokens using the \modelname{Llama-3.2-1B} tokenizer. \textbf{(b)} The distribution of expert assignments labeled using \modelname{GPT-4o} for each phrase in the provided response.
    }
    \label{fig:pseudo-label-dist}
\end{figure}

\begin{figure}
    \centering
    \includegraphics[width=1\linewidth]{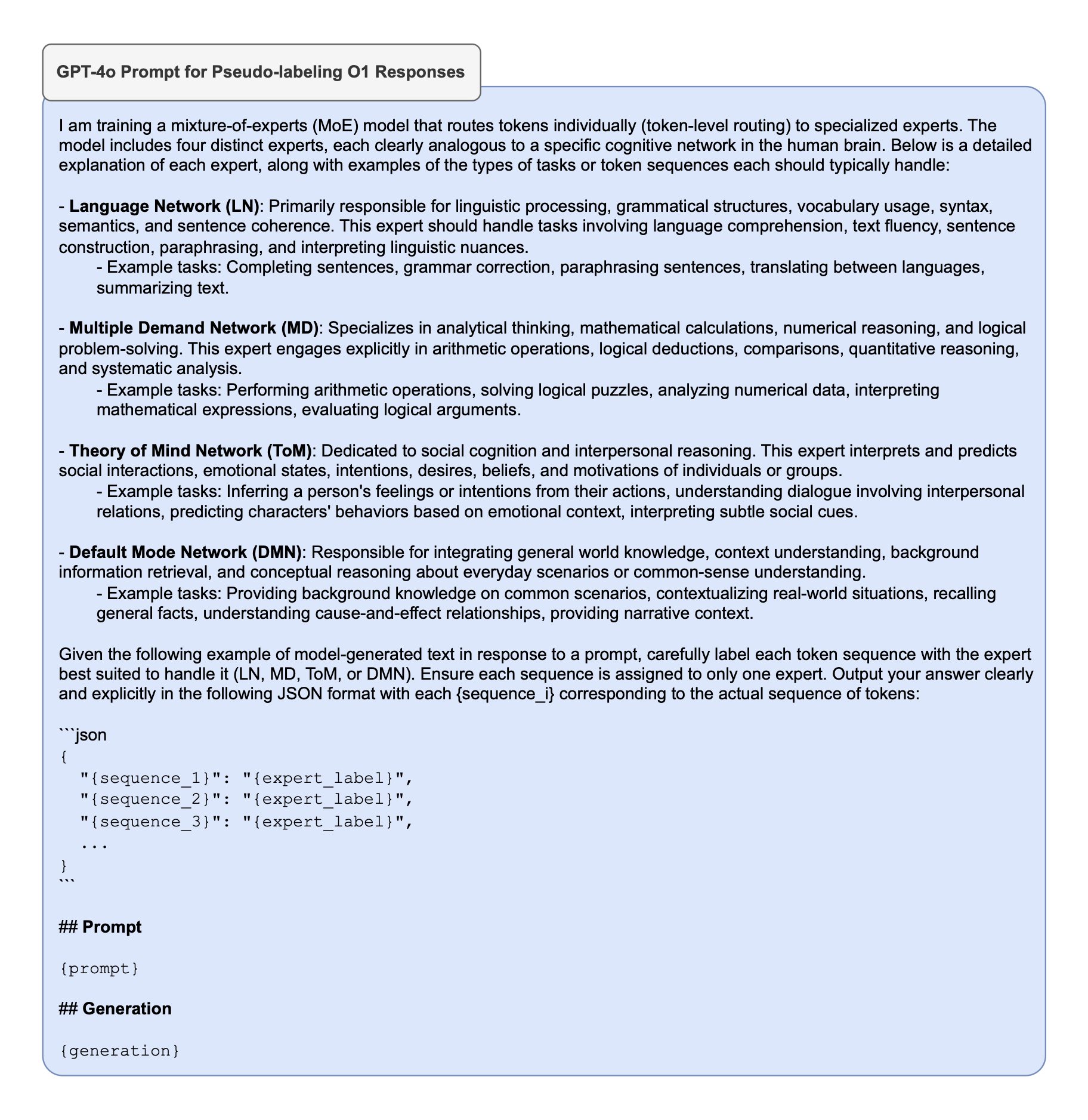}
    \caption{
        \textbf{Prompt Used for Pseudo-Labeling \modelname{O1} Responses}
        The prompt used to instruct \modelname{GPT-4o} to label the \modelname{O1} model generations given a specific input prompt.
    }
    \label{fig:pseudo-labeling-prompt}
\end{figure}

\begin{figure}
    \centering
    \includegraphics[width=1\linewidth]{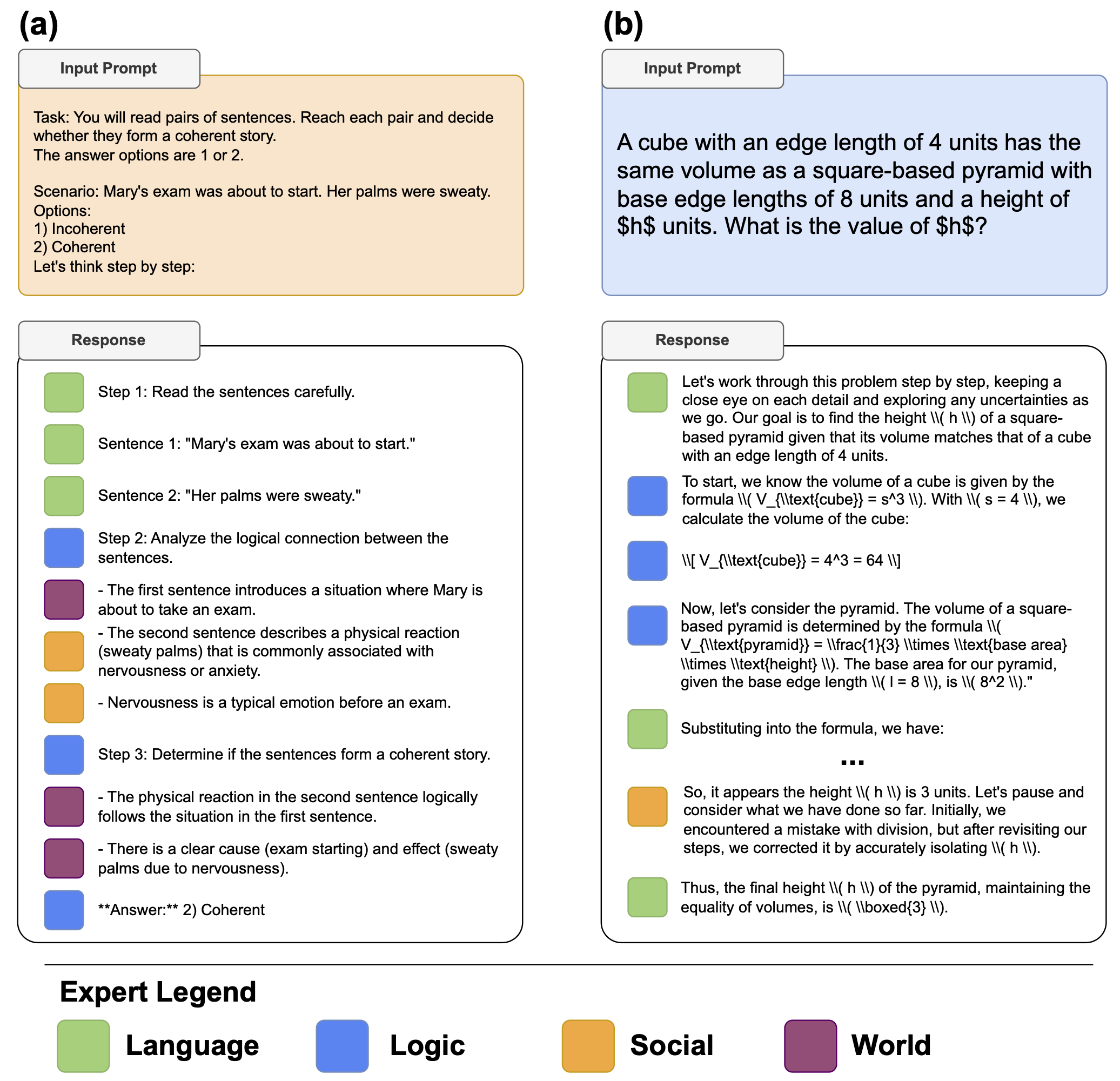}
    \caption{
        \textbf{Examples of Pseudo-Labeled Responses using \modelname{GPT-4o}}
        \textbf{(a)} shows a response generated by \modelname{O1} for a prompt from the coherence subset of the \datasetname{Pragmatics} dataset \citep{hu-etal-2023-fine}.
        \textbf{(b)} shows a response taken directly from the \datasetname{O1-Journey} dataset \citep{o1journey}.
        Each subfigure includes the original prompt, the full model-generated response, and the corresponding pseudo-labels assigned to each phrase.
    }
    \label{fig:pseudo-labeled-examples}
\end{figure}

\paragraph{Pseudo-Labeling Responses}
Finally, once we obtained responses with intermediate reasoning steps for all sampled examples, we used \modelname{GPT-4o} to pseudo-label each phrase in the response. During stage-1 training, each token in a phrase was then assigned to the expert identified by the pseudo-label. This labeling process was guided by the prompt shown in Figure~\ref{fig:pseudo-labeling-prompt}. Figure~\ref{fig:pseudo-labeled-examples} provides examples of labeled responses, while Figure~\ref{fig:pseudo-label-dist} shows the distribution of expert assignments across tokens and phrases.

\begin{figure}
    \centering
    \includegraphics[width=1\linewidth]{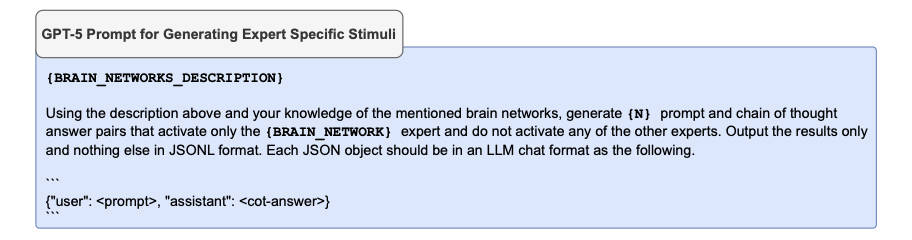}
    \caption{
        \textbf{Expert-Specific Prompt Template Used with GPT-5}
        Prompt provided to \modelname{GPT-5} for generating expert-specific question–answer pairs. The stimuli for each expert was prompted separately using the same brain-network descriptions as in Figure~\ref{fig:pseudo-labeling-prompt}.
    }
    \label{fig:gpt5-token-routing}
\end{figure}

\section{Inter-Annotator Agreement Analysis of \microsft}
\label{app:agreement}

We assess the reliability of the \microsft pseudo-labels using a set of standard agreement metrics between three human annotators and the \modelname{GPT-4o} labels on a subset of the \microsft dataset. For the three human annotators, we report Krippendorff's $\alpha$ (a chance-corrected multi-annotator reliability metric), Fleiss' $\kappa$ (multi-rater extension of Cohen's $\kappa$), and pairwise Cohen's $\kappa$ with percent agreement. We also measure the agreement between the humans and the LLM labels both against the human majority vote and by treating the LLM as a fourth annotator.

\subsection{Human--Human Agreement}
Table~\ref{tab:human-human} summarizes agreement across the three human annotators. Both Krippendorff's $\alpha$ and Fleiss' $\kappa$ indicate moderate reliability ($0.517$). The three-way exact agreement rate (all annotators assign the same label) is $49.4\%$.

Pairwise metrics (Table~\ref{tab:human-human-pairwise}) reveal that annotators H1 and H2 exhibit stronger mutual agreement ($\kappa = 0.681$) relative to H1--H3 and H2--H3, which show lower yet nontrivial agreement levels ($\kappa \approx 0.43$--$0.45$).

\begin{table}[h]
\centering
\caption{Agreement among human annotators.}
\label{tab:human-human}
\begin{tabular}{l c}
\toprule
\textbf{Metric} & \textbf{Value} \\
\midrule
Krippendorff's $\alpha$ (nominal, 3 annotators) & 0.517 \\
Fleiss' $\kappa$ (3 annotators) & 0.517 \\
3-way exact agreement & 49.4\% \\
\bottomrule
\end{tabular}
\end{table}

\begin{table}[t]
\centering
\caption{Pairwise agreement between human annotators.}
\label{tab:human-human-pairwise}
\begin{tabular}{l c c}
\toprule
\textbf{Annotator Pair} & \textbf{Percent Agreement} & \textbf{Cohen's $\kappa$} \\
\midrule
H1--H2 & 76.3\% & 0.681 \\
H1--H3 & 58.9\% & 0.431 \\
H2--H3 & 58.5\% & 0.446 \\
\bottomrule
\end{tabular}
\end{table}

\subsection{Human--LLM Agreement}
We evaluate human--LLM agreement in two ways:  
(1) comparing the LLM to the majority-vote label of the three humans, and  
(2) treating the LLM as a fourth annotator.

\paragraph{LLM vs. Human Majority Vote.}  
Out of the full set, 240 items had a unique human majority label (13 items exhibited three-way ties). On this subset, the LLM achieves the performance shown in Table~\ref{tab:llm-majority}. The Cohen's $\kappa$ between the LLM and the majority vote is $0.533$, indicating substantial agreement comparable to human--human levels.

\begin{table}[h!]
\centering
\caption{LLM agreement with human majority-vote labels.}
\label{tab:llm-majority}
\begin{tabular}{l c}
\toprule
\textbf{Metric} & \textbf{Value} \\
\midrule
Accuracy & 0.658 \\
Macro F1 & 0.666 \\
Cohen's $\kappa$ (LLM vs. majority) & 0.533 \\
\bottomrule
\end{tabular}
\end{table}

\paragraph{LLM as a Fourth Annotator.}  
We also compute multi-annotator reliability including the LLM (Table~\ref{tab:llm-fourth}). Krippendorff’s $\alpha$ decreases slightly to $0.497$, reflecting the LLM’s moderate alignment with the human annotators.

Pairwise comparisons between each human annotator and the LLM (Table~\ref{tab:llm-pairwise}) show agreement levels similar to those between some human pairs (particularly H1--H3 and H2--H3).

\begin{table}[h!]
\centering
\caption{Agreement among 3 humans + LLM.}
\label{tab:llm-fourth}
\begin{tabular}{l c}
\toprule
\textbf{Metric} & \textbf{Value} \\
\midrule
Krippendorff's $\alpha$ (nominal, 4 annotators) & 0.497 \\
\bottomrule
\end{tabular}
\end{table}

\begin{table}[h!]
\centering
\caption{Pairwise agreement between humans and the LLM.}
\label{tab:llm-pairwise}
\begin{tabular}{l c c}
\toprule
\textbf{Annotator Pair} & \textbf{Percent Agreement} & \textbf{Cohen's $\kappa$} \\
\midrule
H1--LLM & 62.8\% & 0.489 \\
H2--LLM & 62.5\% & 0.492 \\
H3--LLM & 59.7\% & 0.448 \\
\bottomrule
\end{tabular}
\end{table}

Overall, the dataset exhibits moderate inter-annotator consistency across all metrics, with variation typical of multi-class subjective labeling tasks. The LLM aligns with human labels at levels comparable to human–human agreement, and the LLM agrees more with H1 and H2, who have a higher inter-annotator agreement. 


\section{Token Routing Patterns}
\label{app:token-routing-patterns}

\begin{figure}
    \centering
    \includegraphics[width=1\linewidth]{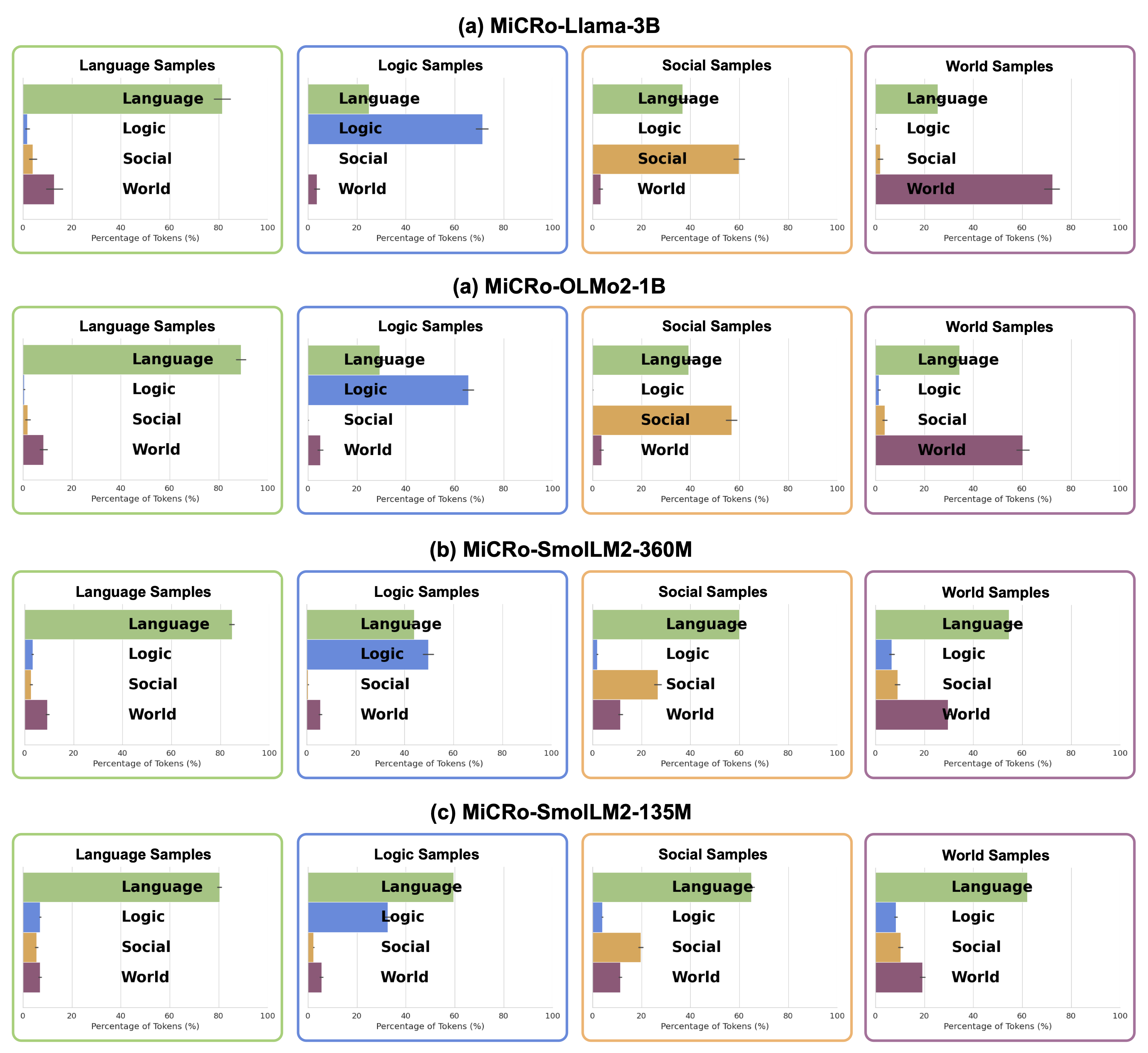}
    \caption{
        \textbf{Token Routing Patterns in Additional \ourmodel-Models.}
        Percentage of tokens routed to each expert, aggregated across all layers, for additional \ourmodel models. Distributions are computed over GPT-5–generated question–answer pairs designed to engage specific domains. Results show consistent brain-inspired specialization, with tokens preferentially assigned to the relevant experts depending on the task domain. Figure~\ref{fig:micro-layer-wise-token-routing} shows the corresponding layer-wise token routing of these plots, while Figure~\ref{fig:appendix-token-routing-patterns} shows the token routing on benchmark testing data.
    }
    \label{fig:micro-token-routing}
\end{figure}

\paragraph{GPT-5 Prompt for Generating Expert Specific Stimuli}
Figure~\ref{fig:gpt5-token-routing} presents the prompt used to instruct \modelname{GPT-5} to generate the question–answer pairs shown in the non-benchmark token-routing pattern plots. The descriptions of the brain networks are identical to those in Figure~\ref{fig:pseudo-labeling-prompt}, which were previously used to pseudo-label \modelname{O1}-generated responses for constructing the \datasetname{\microsft} dataset. We queried \modelname{GPT-5} separately for each expert. We show the routing patterns for additional models in Figure~\ref{fig:micro-token-routing} and for \modelname{MoE} models in Figure~\ref{fig:moe-routing-patterns}.

\begin{figure}
    \centering
    \includegraphics[width=1\linewidth]{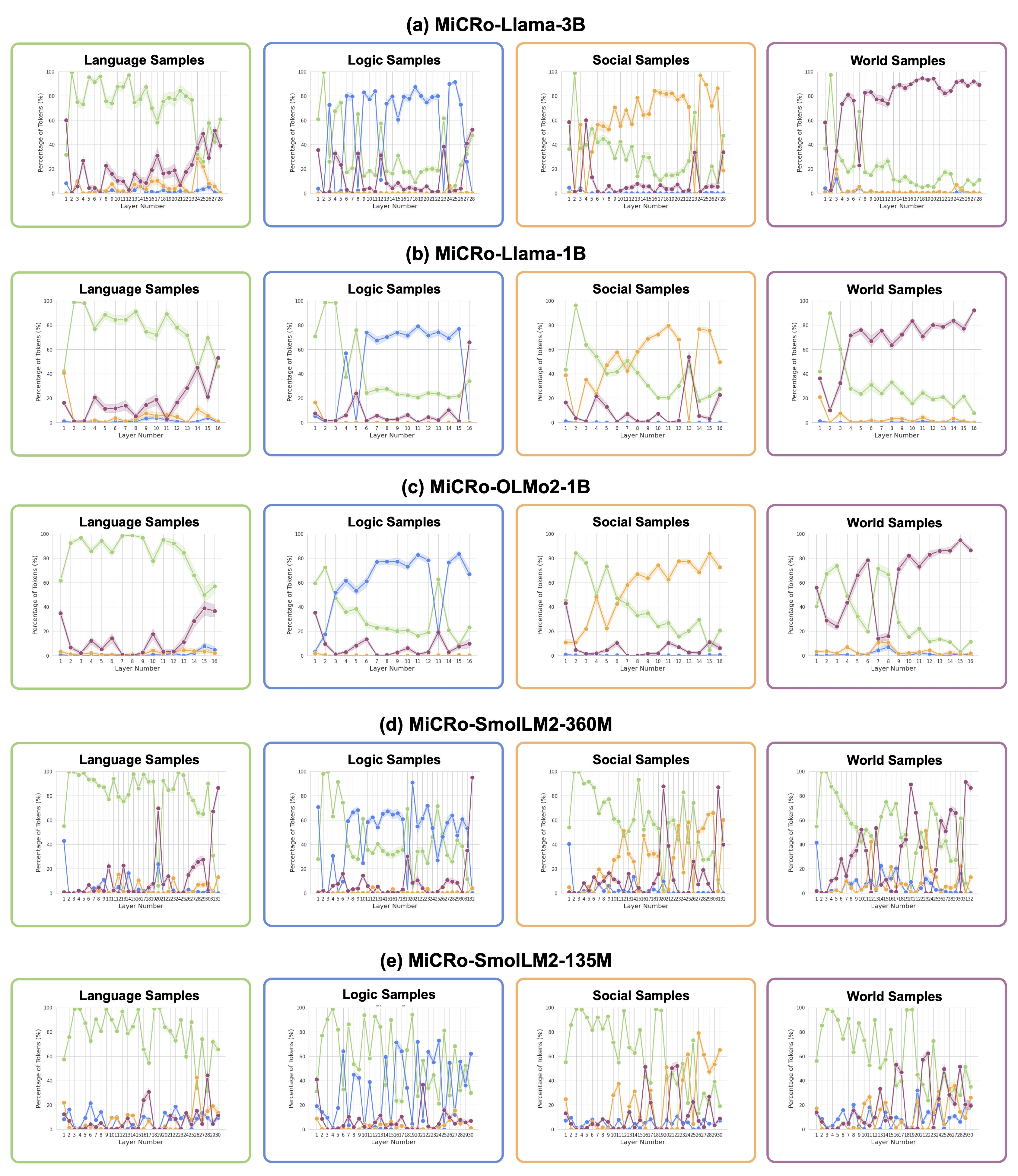}
    \caption{
        \textbf{Layer-wise Token Routing in \ourmodel Models}.
        Token routing distributions across layers for five \ourmodel models, measured on GPT-5–generated question–answer pairs targeting specific domains. In all models, the language expert is consistently engaged in early layers, while domain-specific experts (logic, social, world) are increasingly activated in deeper layers. This hierarchical organization parallels findings from cognitive neuroscience, where linguistic processing precedes engagement of higher-level networks.
    }
    \label{fig:micro-layer-wise-token-routing}
\end{figure}

\begin{figure}
    \centering
    \includegraphics[width=1\linewidth]{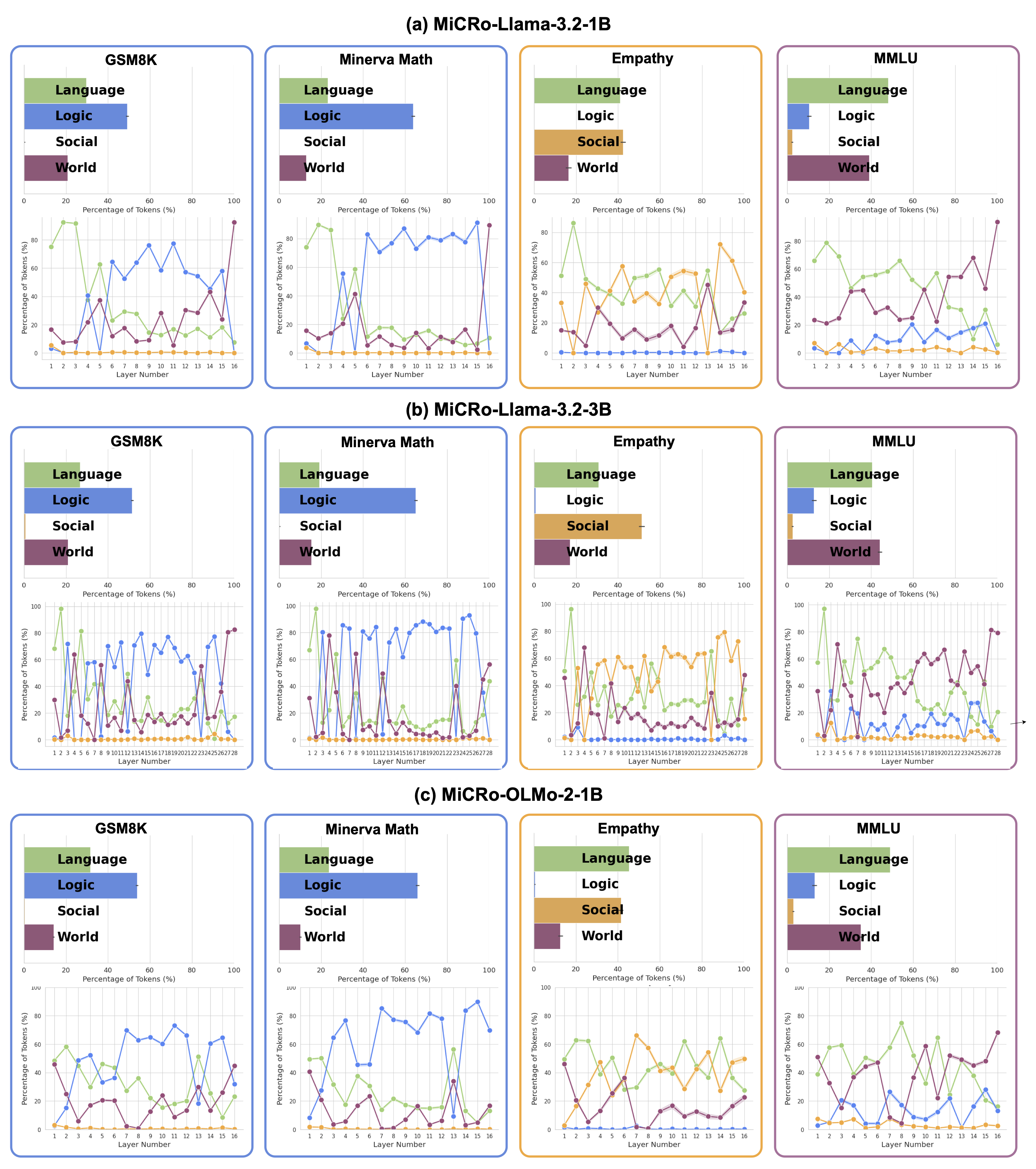}
    \caption{
        \textbf{Benchmark Token Routing Patterns.}
    Token routing patterns for \textbf{(a)} \modelname{MiCRo-Llama-3.2-1B}, \textbf{(b)} \modelname{MiCRo-Llama-3.2-3B}, and \textbf{(c)} \modelname{MiCRo-OLMo-2-1B}, evaluated on up to 1,000 samples drawn from the \datasetname{GSM8K}, \datasetname{Minerva-Math}, \datasetname{Empathy}, and \datasetname{MMLU} test sets. For each model, the top panel reports the overall percentage of tokens routed to each expert across the whole model (variance across samples), while the bottom panel shows layer-wise routing. The latter reveals an emergent hierarchy: earlier layers emphasize language grounding, whereas deeper layers increasingly delegate to domain-relevant experts.
    }
    \label{fig:appendix-token-routing-patterns}
\end{figure}

\paragraph{Layerwise Routing Patterns}
Figure~\ref{fig:micro-layer-wise-token-routing} illustrates layer-wise token routing patterns for five \ourmodel models. Surprisingly, consistent trend emerges: tokens are initially processed by the language expert before being delegated to higher-level experts depending on the task domain. This organization parallels findings in cognitive neuroscience, where the language network is engaged early for virtually all linguistic input and then interfaces with other specialized networks (such as multiple-demand or social cognition systems) depending on task demands \citep{Fedorenko2024}. In Figure~\ref{fig:appendix-token-routing-patterns}, we show benchmark-specific token routing patterns across layers as well. To probe social specialization directly, we also include evaluation on the \datasetname{Empathy} benchmark \citep{buechel-etal-2018-modeling-empathy}, which primarily engages the social expert, further confirming the expected routing behavior is generalizable across datasets.

\section{Correlation with Human Judgments}
\label{app:corr-human-ratings}

We use a dataset of 1,000 six-word sentences from \citet{tuckute2024driving}, each annotated with human ratings across several behavioral dimensions, collected independently of our routing framework. To test correlations with human judgments, we selected features expected to align with specific experts: \datasetname{Grammaticality} and \datasetname{Plausibility} with the language expert, \datasetname{Mental States} with the social expert, and \datasetname{Physical Objects} and \datasetname{Places} with the world expert. The dataset does not include features relevant to the logic expert.

\begin{figure}
    \centering
    \includegraphics[width=1\linewidth]{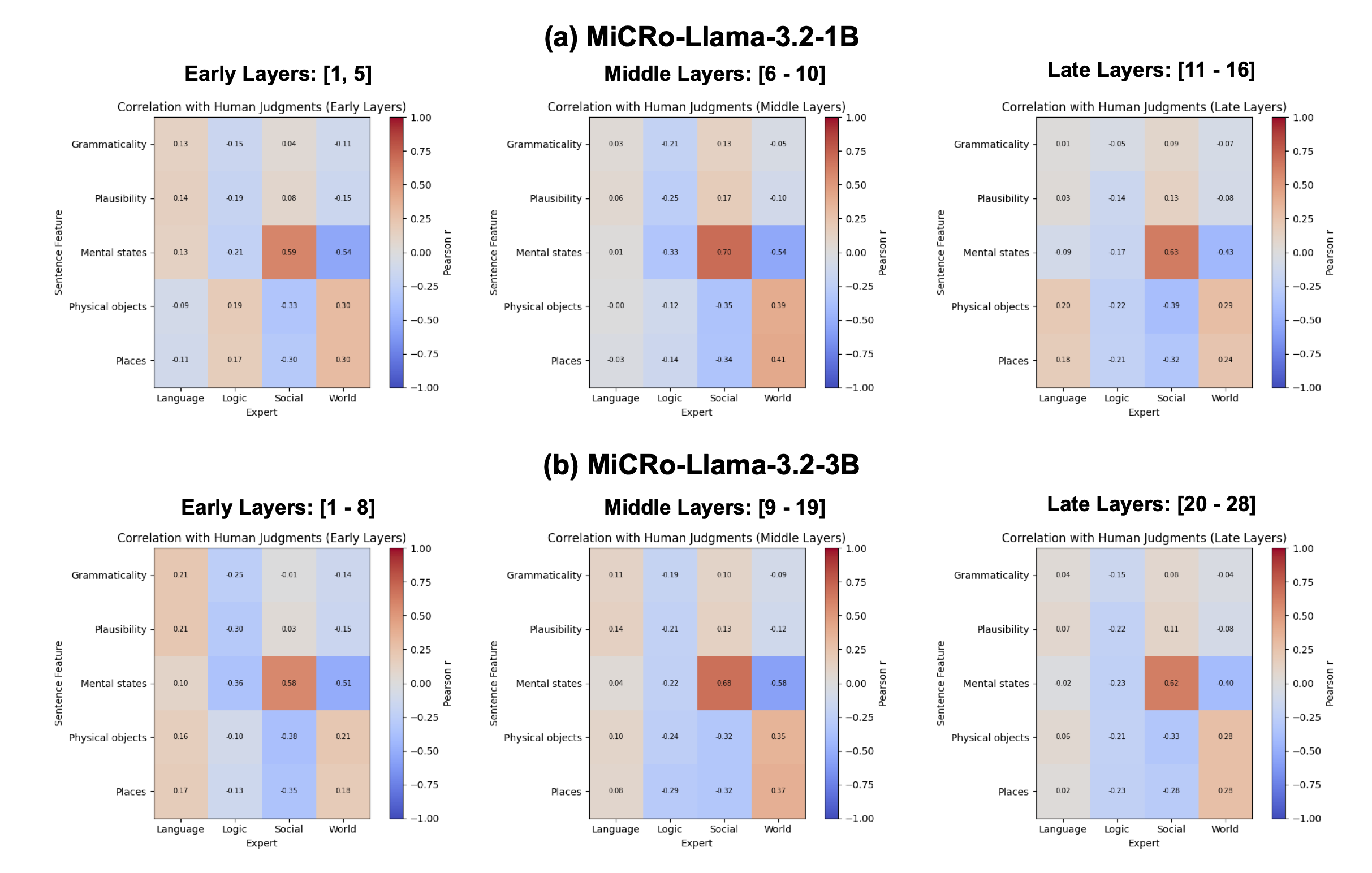}
    \caption{
        \textbf{Correlations Between Expert Routing Probabilities and Human Ratings.} 
        Correlations are shown for \modelname{MiCRo-Llama-3.2-1B} and \modelname{MiCRo-Llama-3.2-3B}, averaged across early, middle, and late layer segments. Mental state ratings correlate most strongly with the social expert, grammaticality and plausibility correlate to some degree with the language expert (primarily in early layers), and physical objects and places with the world expert. The logic expert shows no positive correlations with these features.
    }
    \label{fig:corr-human-ratings}
\end{figure}

To analyze these relationships, we divide each model into three layer segments (early, middle, late) and averaged router probabilities within each segment. Figure~\ref{fig:corr-human-ratings} reports correlations between the average routing probability of each expert and human ratings for \modelname{MiCRo-Llama-3.2-1B} and \modelname{MiCRo-Llama-3.2-3B}. For both models and layer segments, mental state ratings correlate most strongly with the social expert. Language expert probabilities correlate with \datasetname{Grammaticality} and \datasetname{Plausibility}, but primarily in early layers. \datasetname{Physical Objects} and \datasetname{Places} correlate with the world expert, while the logic expert shows no positive correlations (and in most cases negative correlations) with these features. These findings suggest that our router exhibits a meaningful degree of correspondence with human behavioral judgments.

\section{Additional Expert Ablation Results}
\label{app:expert-ablations}

\begin{figure}
    \centering
    \includegraphics[width=1\linewidth]{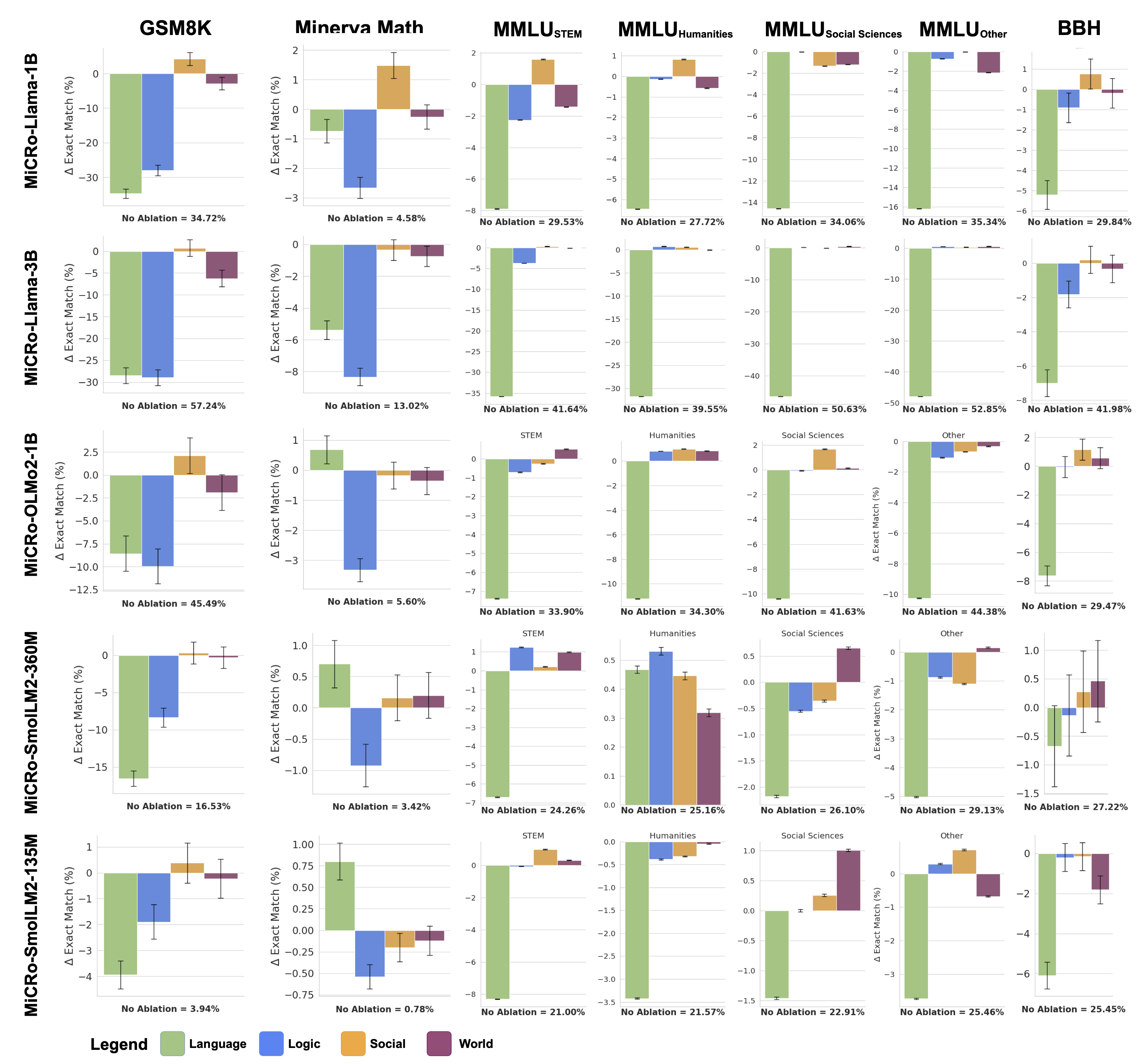}
    \caption{
        \textbf{Expert Ablation Results Across Benchmarks.}
        Impact of ablating individual experts on benchmark performance for five \ourmodel models. Results are shown for \datasetname{GSM8K}, \datasetname{Minerva Math}, \datasetname{BBH}, and \datasetname{MMLU}, with the latter divided into its four subcategories. Removing the language expert causes substantial drops across most tasks, while domain-specific experts (e.g., logic for math benchmarks) are critical for their respective domains. In some cases, ablating an expert improves performance, suggesting interference with more relevant experts.
    }
    \label{fig:appendix-ablations}
\end{figure}

Figure~\ref{fig:appendix-ablations} reports the effect of ablating individual experts, including the language expert, on benchmark performance for five \ourmodel models. We find that the language expert is essential for most tasks, while domain-specific experts---such as the logic expert for \datasetname{GSM8K} and \datasetname{Minerva Math}---are also necessary to maintain performance. Interestingly, in some cases, ablating an expert improves performance, suggesting that certain experts may interfere with more relevant ones, leading to performance degradation when all are active.

\section{Benchmarks}
\label{app:benchmarks}

\begin{table}[htbp]
\centering
\caption{
    \textbf{Number of Shots and Samples Per Benchmark Used in Evaluation.}
    Number of shots and samples used when evaluating the test-set of each benchmark. Last two row shows whether we used CoT or evaluated using log-probabilities and the metric used to obtain the final accuracy.
}
\label{tab:benchmark_shots}
\resizebox{1\textwidth}{!}{%
\begin{tabular}{@{}lccccccccc@{}} 
\toprule
\textbf{Benchmark} & \textbf{GSM8K}  & \textbf{Minerva Math} & \textbf{MMLU} & \textbf{MMLU-Pro} & \textbf{BBH} & \textbf{HellaSwag}  & \textbf{PIQA}  & \textbf{ARC$_\text{Easy}$}   & \textbf{ARC$_\text{Challenge}$}  \\
\midrule
\textbf{N-Shots} & 0-Shot  & 4-shots & 4-shots & 5-shots & 3-Shots & 0-Shot & 0-Shot & 0-Shot & 0-Shot \\
\textbf{Num Samples} & 1,319 & 5,000 & 14,042 & 12,032 & 6,511 & 10,042 & 1,838 & 2,376 & 1,172 \\
\textbf{CoT Prompting} & Yes & Yes & Yes & Yes & Yes & No & No & No & No \\
\textbf{Metric} & Exact Match & Exact Match & Exact Match & Exact Match & Exact Match & Acc Norm & Acc Norm & Acc & Acc Norm \\
\bottomrule 
\end{tabular}%
}
\end{table}

\paragraph{Benchmarks Description}
We evaluate our models on eight benchmarks using various fewshot settings, four of which are prompted to generate a reasoning chain before producing the final answer. These reasoning steps are intended to more meaningfully engage the expert modules throughout the generation process, which is why we focused on them in the main paper. The other benchmarks are multiple choice questions where the most likely candidate---as measured by the log-probabilities of the model---is taken as the prediction. Table~\ref{tab:benchmark_shots} lists the number of in-context examples used and the number of samples tested for each benchmark. For the remaining benchmarks, we used the default fewshot examples from the \texttt{lm-evaluation-harness} \citep{eval-harness} repository. Specifically, we used the \texttt{bbh\_cot\_fewshot} task for \datasetname{BBH}, the \texttt{mmlu\_flan\_cot\_fewshot} task for \datasetname{MMLU}, the \texttt{mmlu\_pro} for \datasetname{MMLU-Pro}, the \texttt{minerva\_math} task for \datasetname{Minerva Math}, and the \texttt{gsm8k\_cot\_zeroshot} for the \datasetname{GSM8K} task. We used the default tasks for the multiple-choice benchmarks.

\paragraph{Extended Benchmark Results}
Table~\ref{tab:additional-results} reports results for additional base models as well as on benchmarks beyond those presented in the main paper. Consistent with the main results, \modelname{MiCRo} remains comparable to Dense and MoB baselines across most tasks while being interpretable. These supplementary experiments provide further evidence that the observed trends hold across a broader range of model scales and evaluation settings.

\begin{table}
\caption{
    \textbf{Additional Benchmark Results for MiCRo and Baselines}
    Accuracy (\%) ± standard error across reasoning and knowledge benchmarks. Results are reported for different model classes (Dense, MoB, and MiCRo) under each base model.
}
\label{tab:additional-results}
\resizebox{1\textwidth}{!}{%
\begin{tabular}{lllllllllll}
\toprule
 &  & \textbf{GSM8K} & \textbf{Minerva} & \textbf{MMLU} & \textbf{MMLU$_{\text{Pro}}$} & \textbf{BBH} & \textbf{ARC$_{\text{Easy}}$} & \textbf{ARC$_{\text{Challenge}}$} & \textbf{HellaSwag} & \textbf{PIQA} \\
\textbf{Base Model} & \textbf{Model} &  &  &  &  &  &  &  &  \\
\midrule
\multirow[c]{3}{*}{SmollM2-135M} & Dense & 2.7 $\pm$ 0.4 & 0.5 $\pm$ 0.1 & 21.5 $\pm$ 0.3 & 7.8 $\pm$ 0.2 & 24.1 $\pm$ 0.5 & 62.8 $\pm$ 1.0 & 29.6 $\pm$ 1.3 & 43.6 $\pm$ 0.5 & 67.7 $\pm$ 1.1 \\
 & MoB & 3.0 $\pm$ 0.5 & 0.6 $\pm$ 0.1 & 21.9 $\pm$ 0.3 & 7.4 $\pm$ 0.2 & 23.5 $\pm$ 0.5 & 63.0 $\pm$ 1.0 & 29.9 $\pm$ 1.3 & 43.5 $\pm$ 0.5 & 67.8 $\pm$ 1.1 \\
 & MiCRo & 3.9 $\pm$ 0.5 & 0.8 $\pm$ 0.1 & 22.5 $\pm$ 0.4 & 7.9 $\pm$ 0.2 & 25.4 $\pm$ 0.5 & 56.0 $\pm$ 1.0 & 27.6 $\pm$ 1.3 & 41.8 $\pm$ 0.5 & 67.5 $\pm$ 1.1 \\
\midrule
\multirow[c]{3}{*}{SmollM2-360M} & Dense & 15.0 $\pm$ 1.0 & 3.7 $\pm$ 0.3 & 26.4 $\pm$ 0.4 & 9.9 $\pm$ 0.3 & 27.3 $\pm$ 0.5 & 69.7 $\pm$ 0.9 & 37.5 $\pm$ 1.4 & 56.6 $\pm$ 0.5 & 71.4 $\pm$ 1.1 \\
 & MoB & 17.4 $\pm$ 1.0 & 3.9 $\pm$ 0.3 & 26.8 $\pm$ 0.4 & 9.8 $\pm$ 0.3 & 27.7 $\pm$ 0.5 &  70.0 $\pm$ 0.9 & 37.1 $\pm$ 1.4 & 56.9 $\pm$ 0.5 & 72.0 $\pm$ 1.0 \\
 & MiCRo & 16.5 $\pm$ 1.0 & 3.4 $\pm$ 0.3 & 26.0 $\pm$ 0.4 & 10.1 $\pm$ 0.3 & 27.2 $\pm$ 0.5 &  69.9 $\pm$ 0.9 & 38.2 $\pm$ 1.4 & 56.7 $\pm$ 0.5 & 71.7 $\pm$ 1.1 \\
\midrule
\multirow[c]{3}{*}{Llama-3.2-1B} & Dense & 36.8 $\pm$ 1.3 & 4.8 $\pm$ 0.3 & 29.7 $\pm$ 0.4 & 11.2 $\pm$ 0.3 & 30.4 $\pm$ 0.5 & 64.3 $\pm$ 1.0 & 33.7 $\pm$ 1.4 & 58.4 $\pm$ 0.5 & 73.8 $\pm$ 1.0 \\
 & MoB & 30.5 $\pm$ 1.3 & 3.7 $\pm$ 0.3 & 27.1 $\pm$ 0.4 & 11.0 $\pm$ 0.3 & 27.4 $\pm$ 0.5 & 61.7 $\pm$ 1.0 & 32.4 $\pm$ 1.4 & 56.2 $\pm$ 0.5 & 71.3 $\pm$ 1.1 \\
 & MiCRo & 34.7 $\pm$ 1.3 & 4.6 $\pm$ 0.3 & 31.2 $\pm$ 0.4 & 10.7 $\pm$ 0.3 & 29.8 $\pm$ 0.5 & 59.6 $\pm$ 1.0 & 32.8 $\pm$ 1.4 & 54.7 $\pm$ 0.5 & 73.1 $\pm$ 1.0 \\
\midrule
\multirow[c]{3}{*}{Llama-3.2-3B} & Dense & 58.0 $\pm$ 1.4 & 14.4 $\pm$ 0.5 & 48.6 $\pm$ 0.4 & 19.6 $\pm$ 0.4 & 44.1 $\pm$ 0.6 & 73.6 $\pm$ 0.9 & 42.9 $\pm$ 1.4 & 68.9 $\pm$ 0.5 & 77.0 $\pm$ 1.0 \\
 & MoB & 51.6 $\pm$ 1.4 & 12.3 $\pm$ 0.5 & 45.2 $\pm$ 0.4 & 19.1 $\pm$ 0.4 & 42.2 $\pm$ 0.6 & 71.6 $\pm$ 0.9 & 41.3 $\pm$ 1.4 & 67.3 $\pm$ 0.5 & 77.0 $\pm$ 1.0 \\
 & MiCRo & 57.2 $\pm$ 1.4 & 13.0 $\pm$ 0.5 & 45.4 $\pm$ 0.4 & 19.0 $\pm$ 0.4 & 42.0 $\pm$ 0.6 & 73.0 $\pm$ 0.9 & 43.3 $\pm$ 1.4 & 67.4 $\pm$ 0.5 & 76.6 $\pm$ 1.0 \\
\bottomrule
\end{tabular}%
}
\end{table}


\section{Robustness Across Post-training Methods} 
\label{app:robustness-post-training}

We further assess the robustness of our method to different post-training methods by applying two variations. First, we further post-train our \ourmodel models using Direct Preference Optimization (DPO) \citep{dpo} on a subset of the \datasetname{Tülu-2.5} preference dataset \citep{ivison2024unpacking} (Table \ref{tab:dpo-model-performance}). Second, we replace the large-scale general-purpose \datasetname{Tülu-3} dataset used in stage-3 with a more domain-specific (medical) instruction-tuning set used in \citet{meditron} (Table \ref{tab:medical-performance}). Our results show that our method is robust to different post-training pipelines, whether applying DPO or using an alternative instruction-tuning dataset, as shown in Tables \ref{tab:dpo-model-performance} and \ref{tab:medical-performance} respectively.

\begin{table}[h!]
\centering
\caption{
    \textbf{Performance After DPO Finetuning}
    Comparison of \modelname{MiCRo} models and baselines after further finetuning with DPO on a preference dataset. Results show average performance across the 4 benchmarks, indicating that specialization remains beneficial after DPO.
}
\label{tab:dpo-model-performance}
\begin{tabular}{@{}llcccccc|c@{}}
\toprule
\textbf{Base Model} & \textbf{Model} & \textbf{GSM8K} & \textbf{Minerva Math} & \textbf{MMLU} & \textbf{BBH}   & \textbf{Average} \\
\midrule
\multirow{2}{*}{Llama-3.2-1B} 
& Dense & 38.1 & 3.9 & 29.4 &  \textbf{30.3}   & 25.4 \\
& \ourmodel & \textbf{39.3} & \textbf{5.8} & \textbf{31.8} &  \textbf{30.3}   & \textbf{26.8} \\
\midrule
\multirow{2}{*}{OLMo-2-1B} 
& Dense & 45.8  & 5.6 & 39.3 & 29.8  & 30.1\\
& \ourmodel & \textbf{48.1} & \textbf{5.8} & \textbf{39.8}  & \textbf{30.4} & \textbf{31.0} \\
\bottomrule
\end{tabular}
\end{table}

\begin{table}[h!]
\centering
\caption{
    \textbf{Performance on Medical Benchmarks After Domain-Specific Instruction Tuning.}  
    Models are finetuned during Stage-3 using a medical instruction-tuning dataset instead of \datasetname{Tülu-3}, and evaluated on four medical benchmarks. Results show that specialization achieve competitive performance across both base models, and outperforming in the out-of-distribution (OOD) setting.  We choose the option with the highest log‑probability among the multiple‑choice options.
}
\label{tab:medical-performance}
\setlength{\tabcolsep}{4pt}
\begin{tabular}{@{}llcc|cc|c@{}}
\toprule
& & \multicolumn{2}{c|}{\textbf{Out-of-Distribution}} & \multicolumn{2}{c|}{\textbf{In-Distribution}} & \\
\cmidrule(lr){3-4} \cmidrule(lr){5-6}
\textbf{Base Model} & \textbf{Model} & \textbf{MMLU Medicine} & \textbf{MedQA} & \textbf{MedMCQA} & \textbf{PubMedQA} & \textbf{Average} \\
\midrule
\multirow{2}{*}{Llama-3.2-1B}  & Dense     & 26.0 & 34.3 & 33.9 & \textbf{73.4} & 41.9 \\
             & \ourmodel    & \textbf{28.3} & \textbf{35.2} & 33.9 & 71.4 & \textbf{42.2} \\
\midrule
\multirow{2}{*}{OLMo-2-1B}     & Dense     & 35.8 & 34.2 & \textbf{35.3} & \textbf{74.0} & 44.8 \\
             & \ourmodel    & 35.8 & \textbf{36.3} & 34.5 & 73.8 & \textbf{45.1} \\
\bottomrule
\end{tabular}
\end{table}

\section{Mixture-of-Experts Results}
\label{app:moe-section}

In the main paper, we report results using the mixture-of-blocks (\modelname{MoB}) architecture, where each expert is a full transformer block with its own attention mechanism. Here, we contrast these results with the more standard mixture-of-experts (\modelname{MoE}) architecture, where experts consist only of FFN blocks and attention is shared across experts within each layer. We first present routing patterns for \ourmodel-MoE models, highlighting cases where our training curriculum fails to induce the intended specialization—an issue we primarily observe in models larger than 1.5B parameters. We then report the performance of the models that did exhibit specialization on reasoning benchmarks.

\subsection{MoE Token Routing Patterns}
\label{app:moe-specialization}

\begin{figure}
    \centering
    \includegraphics[width=1\linewidth]{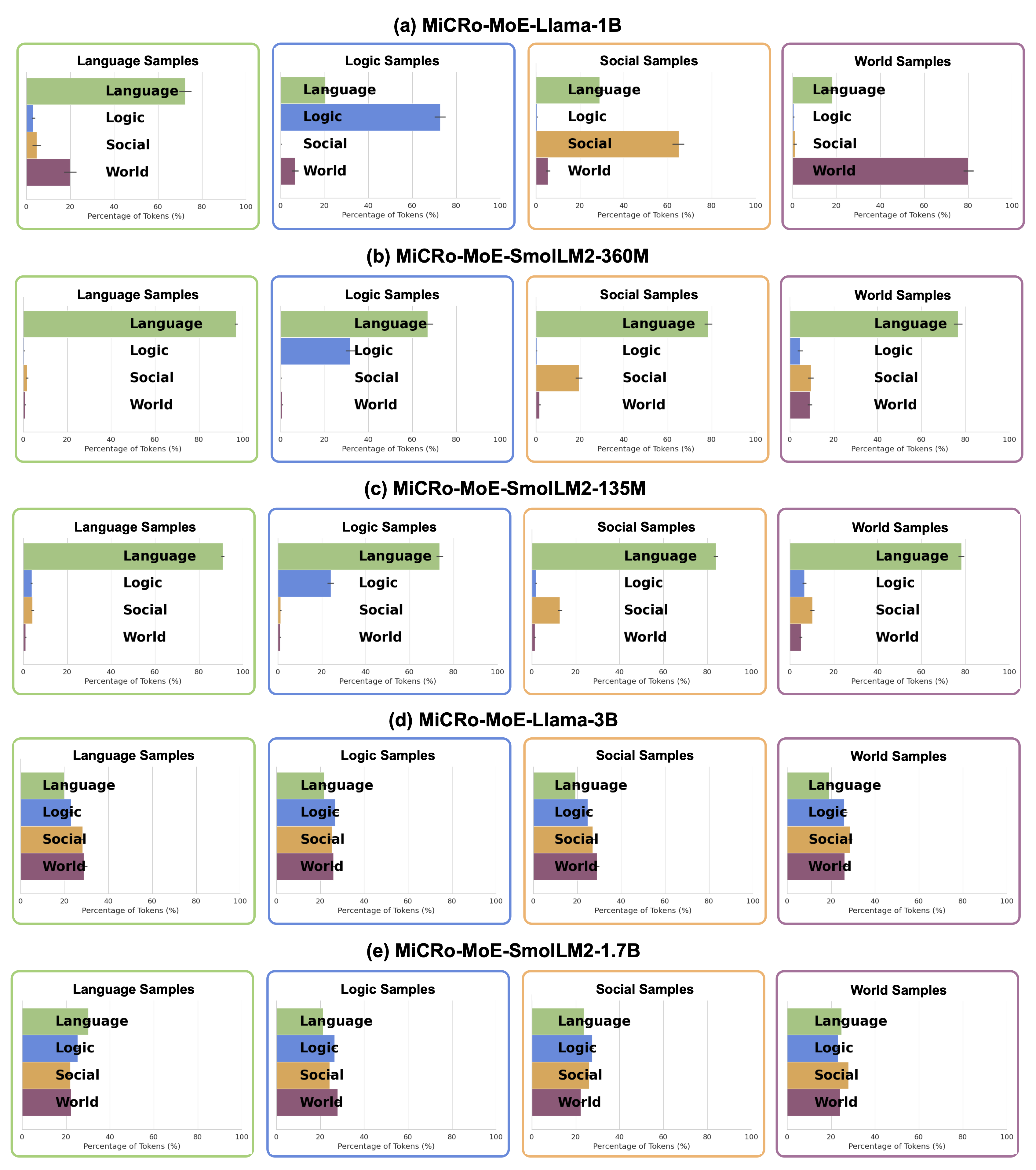}
    \caption{
        \textbf{Routing Patterns in \ourmodel-MoE Models.}
        Routing behavior for five \ourmodel-MoE models on GPT-5–generated question–answer pairs targeting specific experts. The \modelname{MiCRo-MoE-Llama-1B} model shows the intended specialization, while larger variants (e.g., 3B, SmolLM2-1.7B) fail to specialize. Smaller \modelname{SmolLM2} models (135M and 360M) display partial but less consistent specialization, often defaulting to the language expert. These results suggest that the MoE architecture does not reliably induce brain-like specialization under our training curriculum.
    }
    \label{fig:moe-routing-patterns}
\end{figure}

Figure~\ref{fig:moe-routing-patterns} shows routing patterns for five \ourmodel-MoE models on question–answer pairs generated with \modelname{GPT-5} to target specific experts. The \modelname{MiCRo-MoE-Llama-1B} model exhibits the intended specialization, whereas the 3B variant does not. Within the \modelname{SmolLM2} family, the 135M and 360M models display partial specialization, though less cleanly than \modelname{Llama-1B}, often defaulting to the language expert regardless of the input domain. The 1.7B model fails to specialize, similar to \modelname{MiCRo-MoE-Llama-3B}, indicating that the MoE architecture does not reliably induce the desired specialization under our training curriculum.

\subsection{MoE Benchmark Results}
\label{app:moe-results}

Table~\ref{tab:moe-results} presents results for models trained with a Mixture-of-Experts (MoE) design, complementary to the Mixture-of-Blocks (MoB) results reported in the main paper.
The key distinction between MoE and MoB lies in what is replicated to form the experts. In standard MoE, only the feed-forward network (FFN) within each layer is cloned into multiple experts, with the self-attention module shared across all experts. In contrast, MoB duplicates the entire transformer block---including both the attention and FFN components---so that each expert has its own attention mechanism as well as its own FFN.
We find that MoB scales more effectively: under our training curriculum, specialization emerges reliably in larger models ($>$,1B parameters) for MoB, but not for MoE. For this reason, we focus on MoB in the main text and do not include the MoE variants of the other base models, as they did not exhibit the expected functional specialization.

\begin{table}
\caption{
    \textbf{Results with Mixture-of-Experts (MoE) Architectures.}
    Accuracy (\%) ± standard error is reported for Dense, MoE, and \modelname{MiCRo-MoE} models across multiple benchmarks. For each base model, the best score per benchmark is highlighted in bold.
}
\label{tab:moe-results}
\resizebox{1\textwidth}{!}{%
\begin{tabular}{llllllllll}
\toprule
 &  & \textbf{GSM8K} & \textbf{Minerva} & \textbf{MMLU} & \textbf{BBH} & \textbf{ARC$_\text{Easy}$}   & \textbf{ARC$_\text{Challenge}$} & \textbf{HellaSwag} & \textbf{PIQA} \\
\textbf{Base Model} & \textbf{Model} &  &  &  &  &  &  &  &  \\
\midrule
\multirow[c]{3}{*}{SmollM2-135M} & Dense & 2.7 $\pm$ 0.4 & 0.5 $\pm$ 0.1 & 21.5 $\pm$ 0.3 & 24.1 $\pm$ 0.5 & 62.8 $\pm$ 1.0 & 29.6 $\pm$ 1.3 & 43.6 $\pm$ 0.5 & 67.7 $\pm$ 1.1 \\
 & MoE & 2.8 $\pm$ 0.5 & 0.6 $\pm$ 0.1 & 22.3 $\pm$ 0.3 & 24.6 $\pm$ 0.5 & 62.9 $\pm$ 1.0 & 29.4 $\pm$ 1.3 & 43.6 $\pm$ 0.5 & 67.6 $\pm$ 1.1 \\
 & MiCRo-MoE & 4.1 $\pm$ 0.5 & 0.4 $\pm$ 0.1 & 22.2 $\pm$ 0.3 & 24.5 $\pm$ 0.5 & 62.0 $\pm$ 1.0 & 29.0 $\pm$ 1.3 & 43.4 $\pm$ 0.5 & 67.5 $\pm$ 1.1 \\
\midrule
\multirow[c]{3}{*}{SmollM2-360M} & Dense & 15.0 $\pm$ 1.0 & 3.7 $\pm$ 0.3 & 26.4 $\pm$ 0.4 & 27.3 $\pm$ 0.5 & 69.7 $\pm$ 0.9 & 37.5 $\pm$ 1.4 & 56.6 $\pm$ 0.5 & 71.4 $\pm$ 1.1 \\
 & MoE & 16.1 $\pm$ 1.0 & 3.6 $\pm$ 0.3 & 26.6 $\pm$ 0.4 & 27.4 $\pm$ 0.5 & 70.2 $\pm$ 0.9 & 37.2 $\pm$ 1.4 & 56.8 $\pm$ 0.5 & 71.8 $\pm$ 1.1 \\
 & MiCRo-MoE & 16.1 $\pm$ 1.0 & 4.0 $\pm$ 0.3 & 26.1 $\pm$ 0.4 & 27.0 $\pm$ 0.5 & 69.7 $\pm$ 0.9 & 37.5 $\pm$ 1.4 & 56.7 $\pm$ 0.5 & 71.4 $\pm$ 1.1 \\
\midrule
\multirow[c]{3}{*}{Llama-3.2-1B} & Dense & 36.8 $\pm$ 1.3 & 4.8 $\pm$ 0.3 & 29.7 $\pm$ 0.4 & 30.4 $\pm$ 0.5 & 64.3 $\pm$ 1.0 & 33.7 $\pm$ 1.4 & 58.4 $\pm$ 0.5 & 73.8 $\pm$ 1.0 \\
 & MoE & 29.1 $\pm$ 1.3 & 4.7 $\pm$ 0.3 & 25.7 $\pm$ 0.4 & 28.6 $\pm$ 0.5 & 64.1 $\pm$ 1.0 & 35.2 $\pm$ 1.4 & 57.9 $\pm$ 0.5 & 72.5 $\pm$ 1.0 \\
 & MiCRo-MoE & 35.4 $\pm$ 1.3 & 5.0 $\pm$ 0.3 & 30.4 $\pm$ 0.4 & 30.1 $\pm$ 0.5 & 65.0 $\pm$ 1.0 & 35.5 $\pm$ 1.4 & 57.3 $\pm$ 0.5 & 73.8 $\pm$ 1.0 \\
\bottomrule
\end{tabular}%
}
\end{table}


\section{Additional Behavioral Alignment Results}
\label{app:cogbench}

\begin{figure}
    \centering
    \includegraphics[width=1\linewidth]{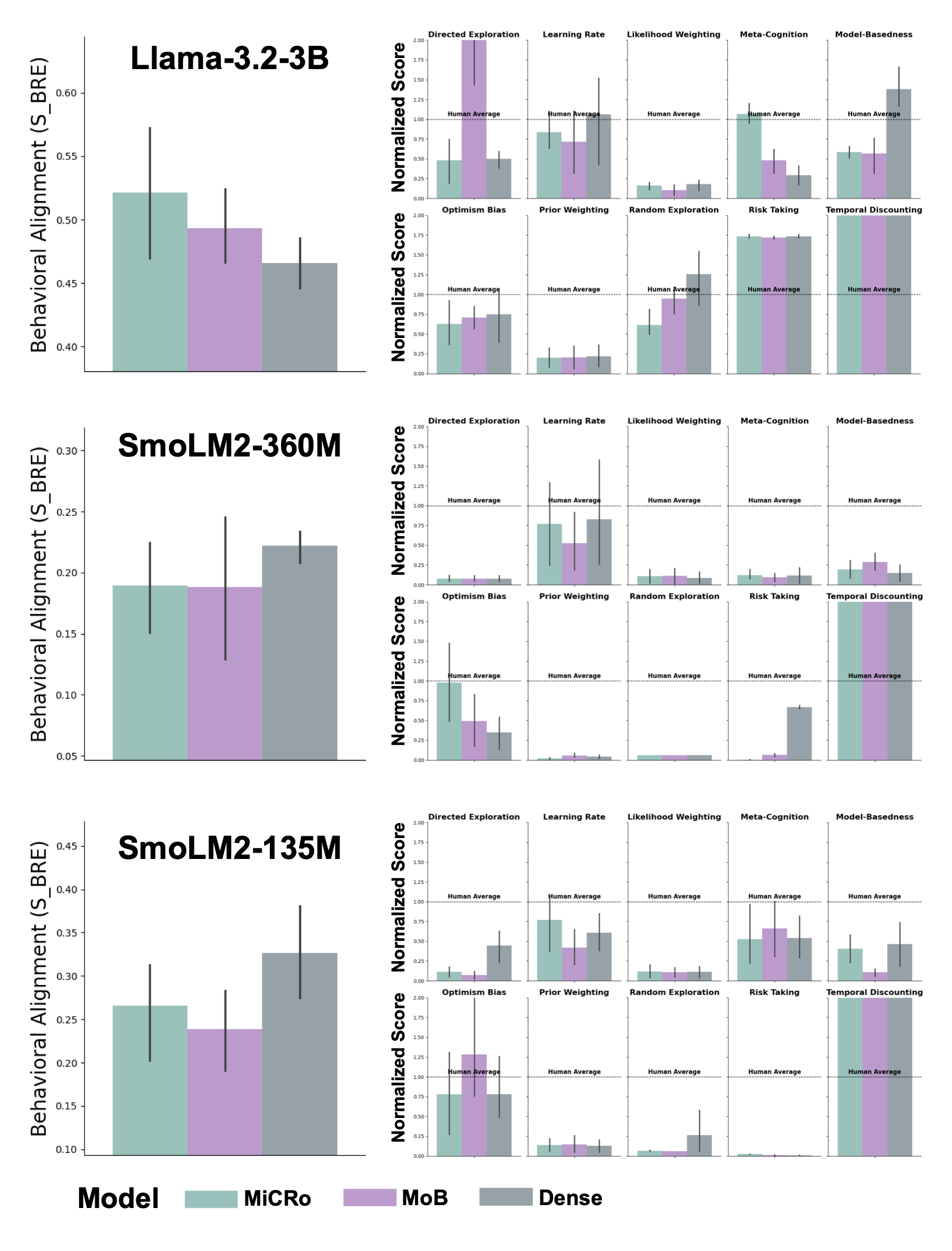}
    \caption{
        \textbf{Behavioral Alignment on Additional Base Models.}
        Results for \modelname{Llama-3.2-3B}, \modelname{SmolLM2-360M}, and \modelname{SmolLM2-135M} on \datasetname{CogBench}. Left: average similarity to human behavior across all metrics. Right: fine-grained results for each behavioral metric. \ourmodel is compared with \modelname{MoB} and \modelname{Dense} baselines, showing stronger alignment in larger models and comparable performance in smaller ones.
    }
    \label{fig:cogbench-appendix}
\end{figure}

Figure~\ref{fig:cogbench-appendix} shows alignment to human behavior for additional base models, comparing \ourmodel with corresponding \modelname{MoB} and \modelname{Dense} baselines. We find that \ourmodel achieves higher average behavioral alignment on \datasetname{CogBench} metrics in larger models, while maintaining comparable performance in smaller models. Please refer to \S\ref{sec:cogbench} for more details on how we evaluate the models.

\section{Specialization Remains Consistent Throughout Training}
\label{app:routing-over-checkpoints}

\begin{figure}[t]
    \centering
    \includegraphics[width=1\linewidth]{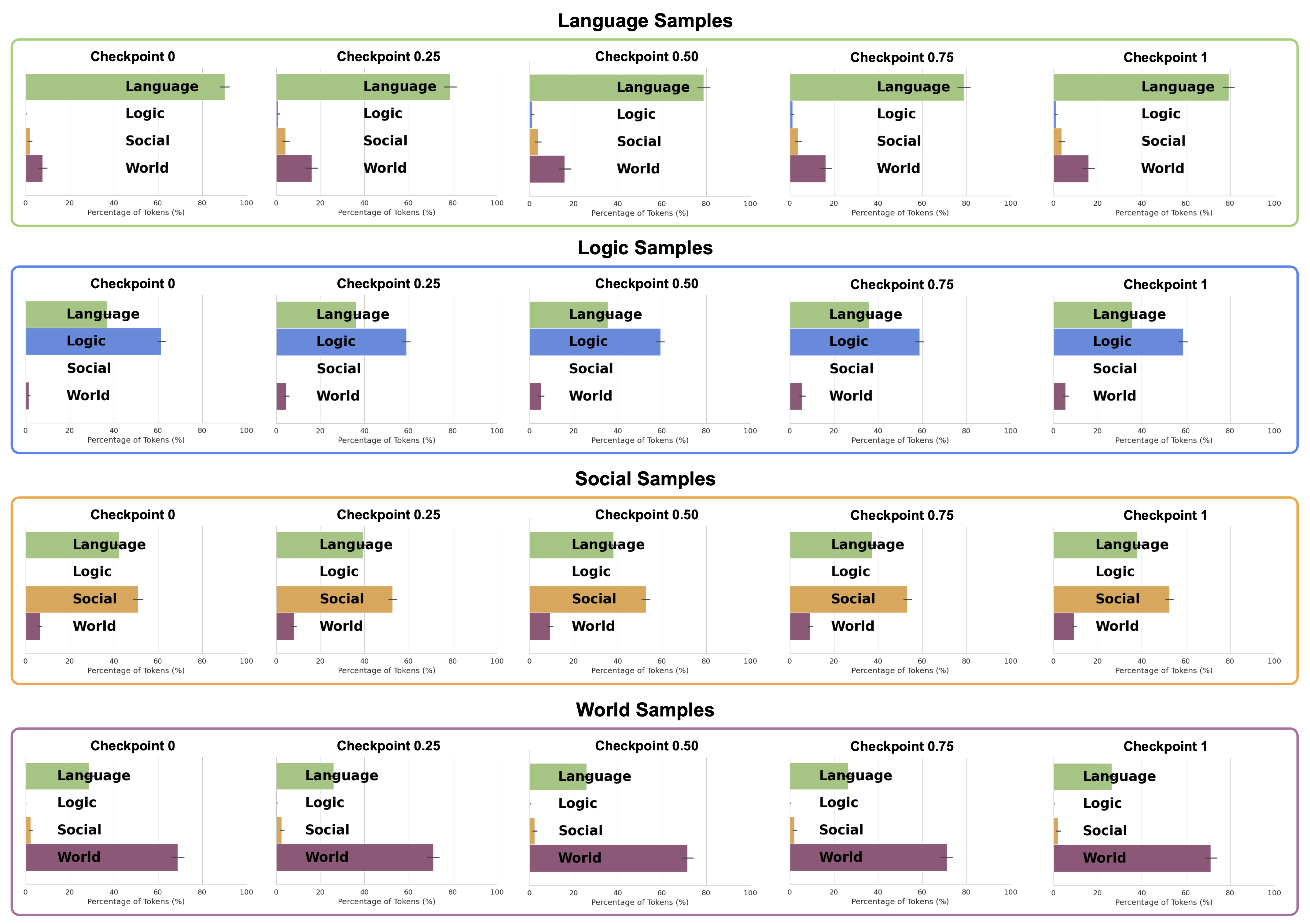}
    \caption{
        \textbf{Specialization Remains Consistent Throughout Training.}
        Token routing across checkpoints during Stage 3 training of \modelname{MiCRo-Llama-1B} on the samples generated to probe each corresponding expert. Checkpoint 0 corresponds to the final weights from Stage 2. The plot shows that expert assignments remain stable throughout training, with minimal variation, indicating that the model retains its learned specialization despite the absence of explicit constraints.
    }
    \label{fig:specialization-ckpts}
\end{figure}

Figure~\ref{fig:specialization-ckpts} illustrates token routing assignments across checkpoints during Stage 3 training of \modelname{MiCRo-Llama-1B}, with checkpoint-0 representing the final weights from Stage 2. The results show that the model consistently preserves the specialization established in stages 1 and 2, despite no explicit constraints being enforced during this phase, except for the initial weak inductive bias. This suggests that brain-like specialization may offer a robust initialization, enabling the model to maintain functionally distinct expert behaviors throughout continued end-to-end training.

\section{Qualitative Examples of Steering Behavior}
\label{app:steering-examples}

\begin{figure}
    \centering
    \includegraphics[width=1\linewidth]{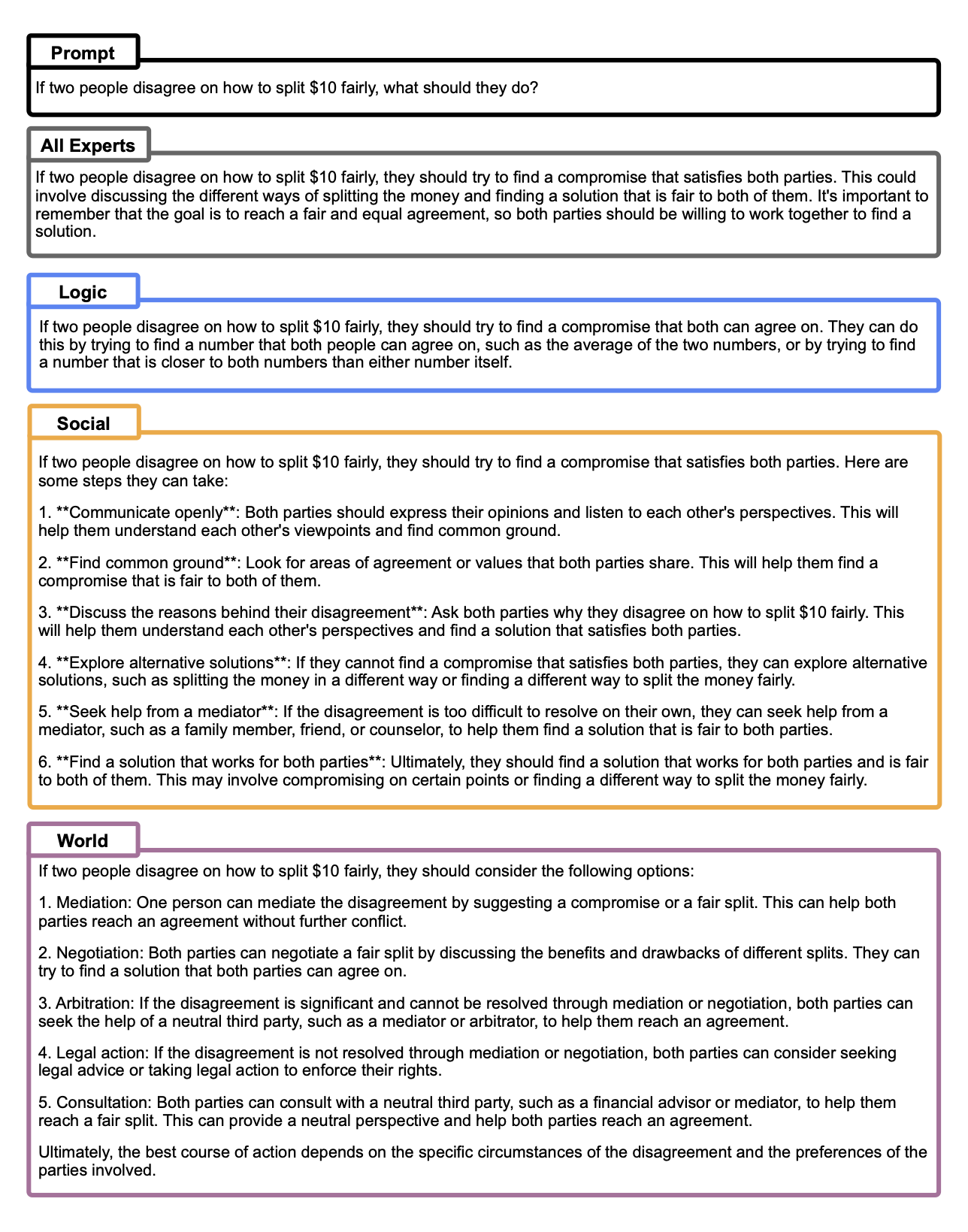}
    \caption{
       \textbf{Example for Steering Model Behavior by Expert Ablation.}
       Responses of \modelname{MiCRo-Llama-3B} to the given prompt when only the target expert and the language expert are retained. The differences illustrate the causal role of each expert and demonstrate how ablations can steer the model’s behavior.
    }
    \label{fig:steering-example-1}
\end{figure}

\begin{figure}
    \centering
    \includegraphics[width=1\linewidth]{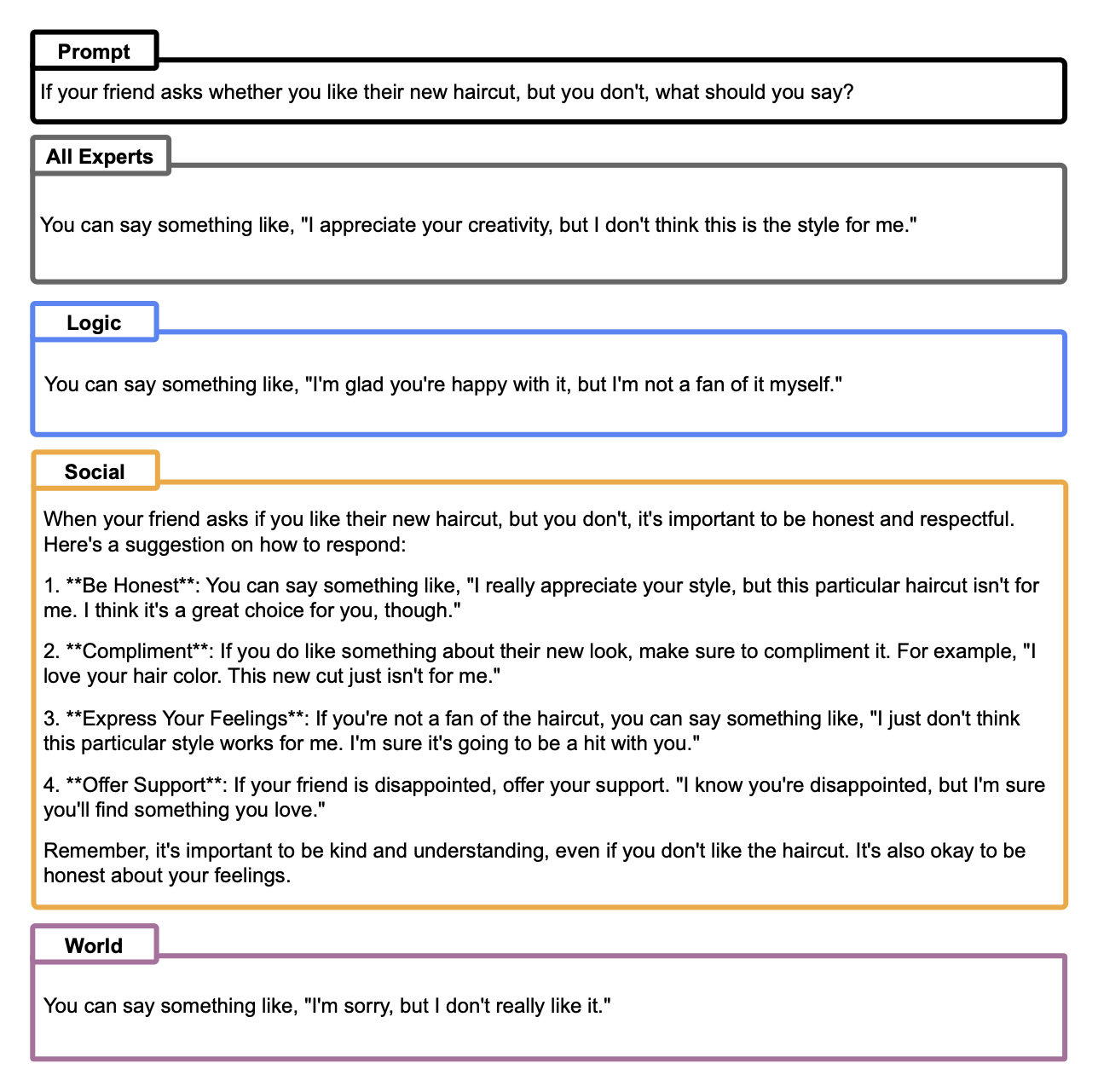}
    \caption{
        \textbf{Example for Steering Model Behavior by Expert Ablation.}
       Responses of \modelname{MiCRo-Llama-3B} to the given prompt when only the target expert and the language expert are retained. The differences illustrate the causal role of each expert and demonstrate how ablations can steer the model’s behavior.
    }
    \label{fig:steering-example-2}
\end{figure}

\begin{figure}
    \centering
    \includegraphics[width=1\linewidth]{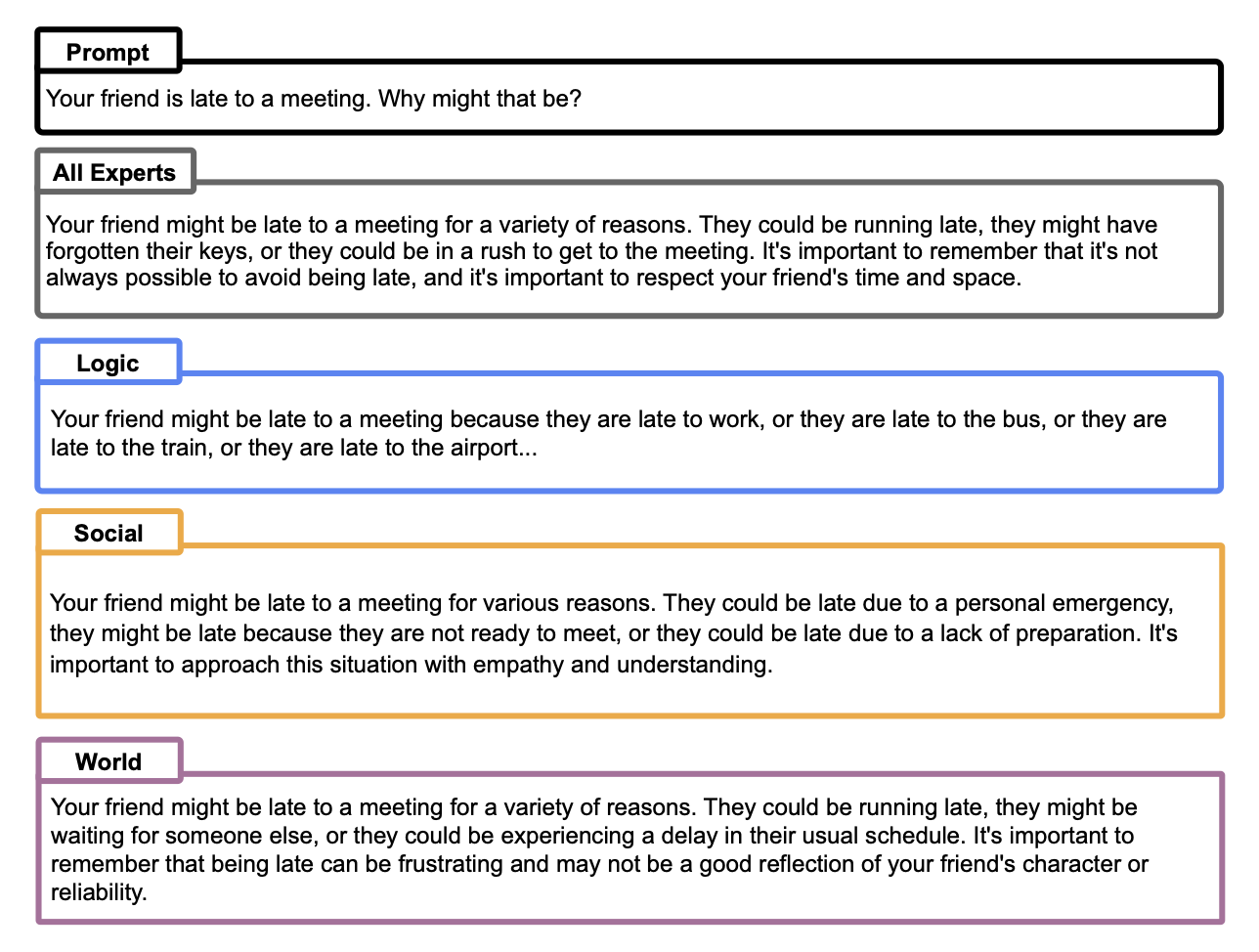}
    \caption{
        \textbf{Example for Steering Model Behavior by Expert Ablation.}
       Responses of \modelname{MiCRo-Llama-1B} to the given prompt when only the target expert and the language expert are retained. The differences illustrate the causal role of each expert and demonstrate how ablations can steer the model’s behavior.
    }
    \label{fig:steering-example-3}
\end{figure}

\begin{figure}
    \centering
    \includegraphics[width=0.90\linewidth]{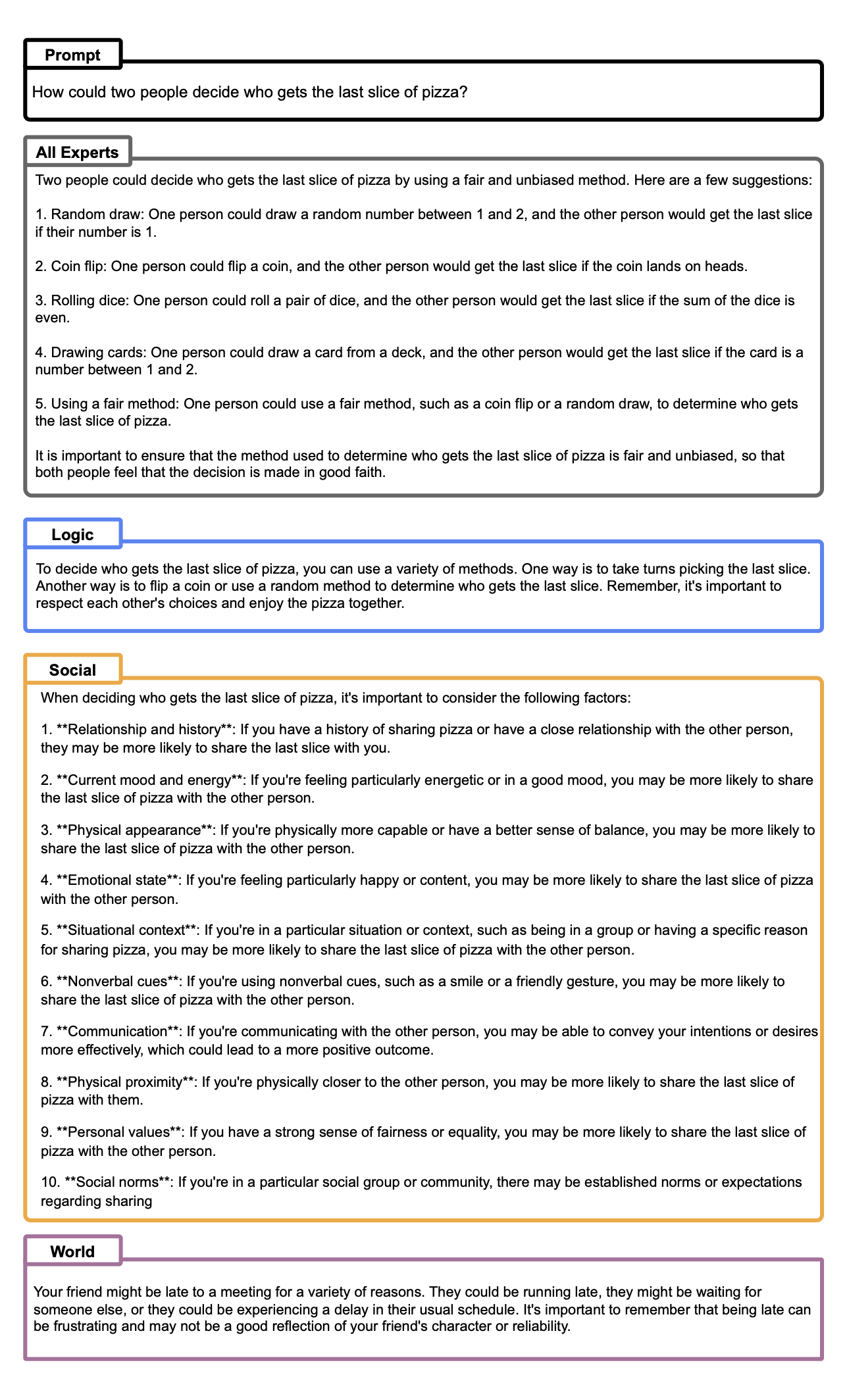}
    \caption{
        \textbf{Example for Steering Model Behavior by Expert Ablation.}
       Responses of \modelname{MiCRo-Llama-3B} to the given prompt when only the target expert and the language expert are retained. The differences illustrate the causal role of each expert and demonstrate how ablations can steer the model’s behavior.
    }
    \label{fig:steering-example-4}
\end{figure}

Figures~\ref{fig:steering-example-1}-\ref{fig:steering-example-4} show examples of how one can use \ourmodel to steer the model's behavior by selectively ablating or activating certain experts. In the examples provided, we only retain the target expert along with the language expert using the \ourmodel models. When the social expert is ablated, the model shifts toward a more analytical tone, producing a response that is logically coherent but lacking in empathy.

\section{Test-Time Scaling by Increasing the Number of Active Experts}
Table~\ref{tab:testtime-scaling} reports the effect of increasing the number of active experts at test time from 1 to 2 for \ourmodel{}-OLMo-1B. The results show that test-time compute can be scaled by enabling additional experts, and that the model generalizes well to this setting and improves performance on all benchmarks, even though the large-scale SFT stage was trained exclusively with top-1 routing. This indicates that \ourmodel{} retains robustness when the routing capacity is expanded at inference time under the $k$=2 setting. However, with larger values of $k$ the performance degrades slightly.

\begin{table}[t]
\centering
\small
\caption{
    Performance of \ourmodel{}-OLMo-1B when increasing the number of active experts from Top-1 to Top-2 at test time. Enabling an additional expert leads to consistent improvements across most benchmarks, demonstrating that the model generalizes well to increased routing capacity even though SFT was performed with Top-1 routing. The benchmarks \datasetname{MATH}, \datasetname{ARC-E}, and \datasetname{ARC-C} refer to \datasetname{Minerva-Math}, \datasetname{ARC-Easy}, and \datasetname{ARC-Challenge} respectively.
}
\begin{tabular}{l|ccccccccc}
\toprule
\textbf{K} & \textbf{GSM8K} & \textbf{BBH} & \textbf{MMLU} & \textbf{MATH} & \textbf{HellaSwag} & \textbf{PIQA} & \textbf{ARC-E} & \textbf{ARC-C} & \textbf{Avg} \\
\midrule

1 & 45.5\% & 29.5\% & 37.9\% & 5.6\% & 65.4\% & 75.2\% & 70.2\% & 41.3\% & 46.3\% \\
2 & \textbf{47.7\%} & \textbf{30.6\%} & \textbf{38.2\%} & \textbf{6.8\%} & \textbf{66.4\%} & \textbf{75.4\%} & \textbf{71.9\%} & \textbf{41.6\%} & \textbf{47.3\%} \\
\bottomrule
\end{tabular}
\label{tab:testtime-scaling}
\end{table}

\section{Scaling \ourmodel to \modelname{Llama-3.1-8B} Base Model}
\begin{figure}[t]
    \centering
    \includegraphics[width=1\linewidth]{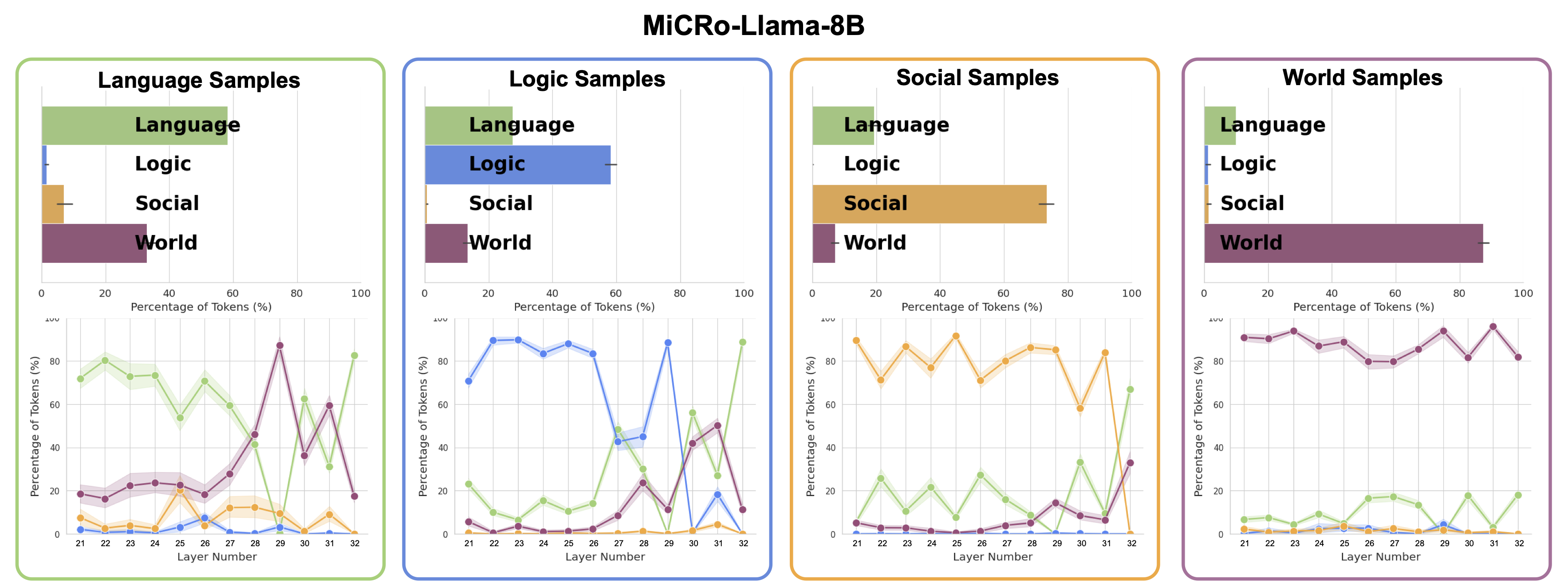}
    \caption{
        \textbf{Token Routing Patterns for \modelname{MiCRo-Llama-8B}}.
        (Top) The percentage of tokens routed to each expert aggregated across the last 12 layers of \modelname{MiCRo-Llama-8B}. The samples are GPT-5 generated question-answer pairs targeting specific domains. (Bottom) The corresponding layer-wise token routing patterns.
    }
    \label{fig:llama-8b-routing-patterns}
\end{figure}

We post-trained three Llama-3.1-8B variants: (1) \modelname{MiCRo-Llama-8B}, (2) its modular baseline \modelname{Llama-8B-MoB}, and (3) \modelname{Llama-8B-Dense}.
However, due to compute constraints, we instantiated experts only in the last 12 layers of \modelname{MiCRo-Llama-8B} and \modelname{Llama-8B-MoB}, keeping the earlier layers dense. This choice is motivated by our prior findings (Figures \ref{fig:micro-layer-wise-token-routing}–\ref{fig:appendix-token-routing-patterns}), which show that early layers predominantly route to language experts, while non-language specializations emerge in later layers. The results confirm that \modelname{MiCRo-Llama-8B} exhibits the expected routing patterns in its last 12 layers, as shown in Figure \ref{fig:llama-8b-routing-patterns}. Table \ref{tab:llama8b-results} shows the results on 9 benchmarks of the three \modelname{Llama-3.1-8B} model variants, along with the results when we remove the most detrimental expert across all layers for a given task for the \ourmodel and \modelname{MoB} models. Using a paired Wilcoxon signed-rank test across the nine benchmarks, we find no statistically significant difference between MiCRo and either of the two baselines. The \modelname{Dense} vs. \ourmodel comparison yields a test statistic of 31.0 (p = 0.18), and the \modelname{MoB} vs. \ourmodel comparison yields 33.5 (p = 0.10), both well above the conventional 0.05 significance threshold. However, when we ablate the most detrimental expert per task for both MiCRo and MoB we see a jump in performance, as also illustrated in Figure \ref{fig:performance}. In general, the Llama-8B experiments confirm our initial results that \ourmodel remains competitive on a suite of benchmarks while remaining interpretable and relevant to cognitive neuroscience.

\begin{table}[t!]
\centering
\small
\caption{
    \textbf{Benchmark Results for Llama-3.1-8B Model Variants}.
    Performance comparison of \modelname{Llama-3.1-8B} variants across multiple benchmarks. Ablation refers to selectively removing the least relevant expert per benchmark.
}
\resizebox{1\textwidth}{!}{%
\begin{tabular}{llcccccccc|c}
\toprule
\textbf{Model Name} 
& \textbf{GSM8K} & \textbf{BBH} & \textbf{MMLU} 
& \textbf{MMLU Pro} & \textbf{MATH} 
& \textbf{HellaSwag} & \textbf{PIQA} & \textbf{ARC-E} 
& \textbf{ARC-C} & \textbf{Avg} \\
\midrule

Llama-8B-Dense              & 71.3 & 54.2 & 51.6 & 24.6 & 21.5 & 72.8 & 78.3 & 74.3 & 44.1 & 54.7 \\
Llama-8B-MoB                & 69.3 & 54.1 & 52.0 & 23.5 & 21.3 & 72.2 & 78.9 & 74.6 & 46.9 & 54.8 \\
Llama-8B-MoB (Ablation)     & 70.0 & 55.9 & 54.1 & 36.3 & 21.6 & 72.4 & 79.3 & 75.0 & 46.9 & 56.8 \\
MiCRo-Llama-8B              & 69.1 & 53.6 & 50.7 & 23.1 & 18.1 & 72.7 & 78.7 & 75.7 & 45.9 & 54.2 \\
MiCRo-Llama-8B (Ablation)   & 69.1 & 55.0 & 52.7 & 36.3 & 18.2 & 73.1 & 78.7 & 76.3 & 46.7 & 56.2 \\
\bottomrule
\end{tabular}%
}
\label{tab:llama8b-results}
\end{table}

\begin{figure}[t]
    \centering
    \includegraphics[width=1\linewidth]{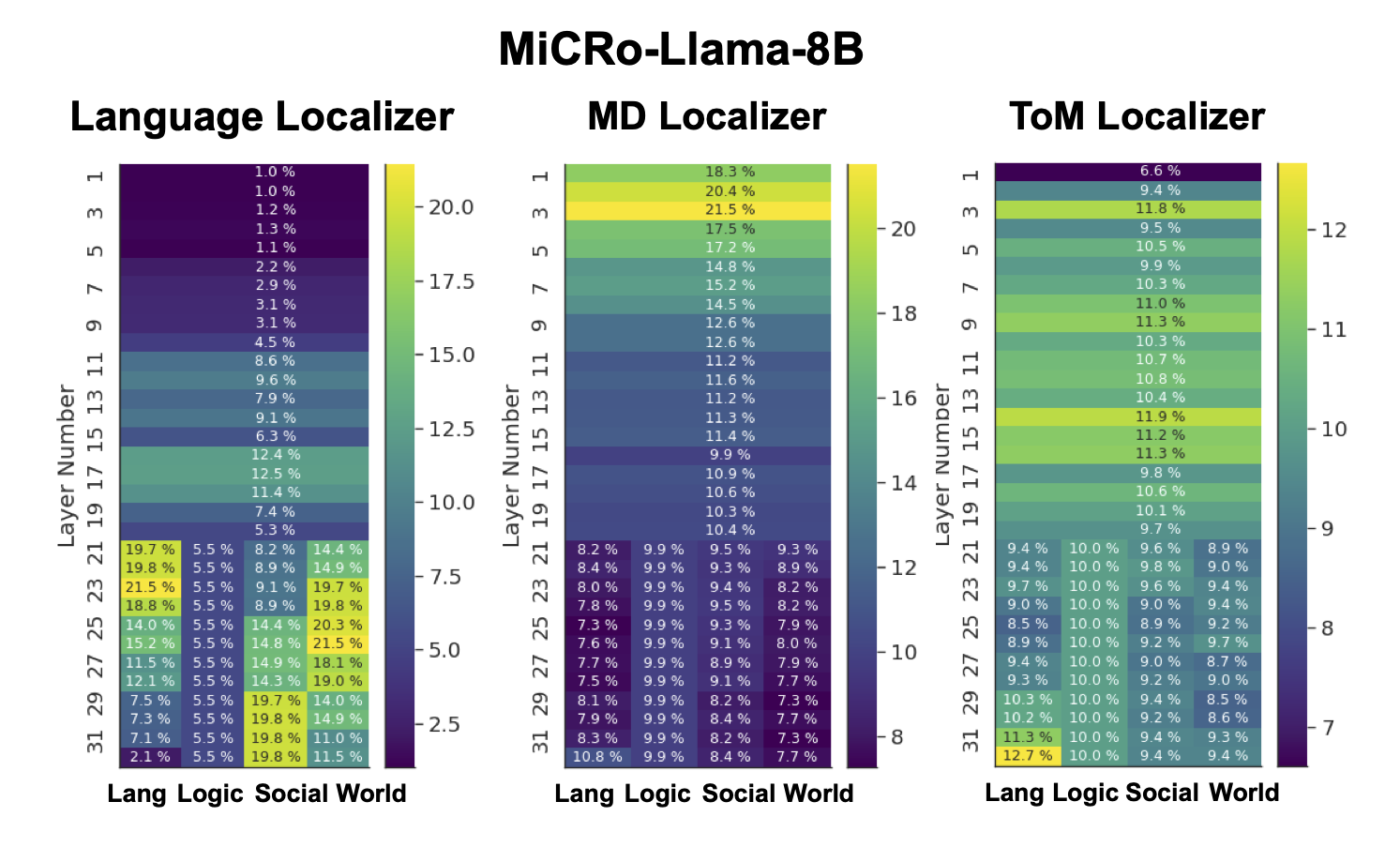}
    \caption{
       \textbf{Neuroscience Localizers Fail to Recover Experts For Hybrid \ourmodel-Llama-8B}
       Following the procedure in Figure \ref{fig:localization}, we apply three neuroscience-inspired functional localizers to \modelname{MiCRo-Llama-8B}, which places experts only in the final 12 layers due to compute constraints. In contrast to the full \ourmodel variants with experts in every layer, the localizers are unable to reliably recover the expected expert specializations in this hybrid configuration.
    }
    \label{fig:llama-8b-localization}
\end{figure}

\section{Large Language Model Usage}

We used large language models (LLMs) solely for editing and grammatical refinement of the manuscript. All substantive ideas, analyses, and conclusions presented in this work are our own.

\end{document}